\documentclass{article}

\usepackage[final]{neurips_2024}

\usepackage[utf8]{inputenc} %
\usepackage[T1]{fontenc}    %
\usepackage{hyperref}       %
\usepackage{url}            %
\usepackage{booktabs}       %
\usepackage{amsfonts}       %
\usepackage{nicefrac}       %
\usepackage{microtype}      %
\usepackage{xcolor}         %

\usepackage{soul}
\usepackage{enumitem,kantlipsum}

\usepackage{amsmath}
\usepackage{amssymb}
\usepackage[title]{appendix}
\usepackage{tabularx}
\usepackage{fontawesome5}
\usepackage{graphicx}

\usepackage{multirow}
\usepackage{multicol}
\usepackage{makecell}
\usepackage{longtable}

\usepackage{pifont}%
\newcommand{\cmark}{\ding{51}}%
\newcommand{\xmark}{\ding{55}}%

\newcommand{\eg}[0]{\textit{e.g.,}}
\newcommand{\etal}[0]{\textit{et al.}}

\newcommand{\eqn}[0]{Eqn.}
\newcommand{\fig}[0]{Fig.}
\newcommand{\tab}[0]{Tab.}

\newcommand{\shortsec}[0]{Sec.}

\newcolumntype{Y}{>{\centering\arraybackslash}X}
\newcolumntype{P}[1]{>{\centering\arraybackslash}p{#1}}

\newcommand{\deformthings}[0]{\text{DeformingThings4D-Animals }}

\definecolor{src}{RGB}{200, 115, 115}
\definecolor{tgt}{RGB}{130, 130, 185}
\definecolor{gt}{RGB}{150, 150, 150}
\definecolor{color_1}{RGB}{255,0,128}
\definecolor{color_2}{RGB}{0,128,128}
\definecolor{color_3}{RGB}{229,161,245}

\newcommand\src[1] {\emph{\textcolor{src}{#1}}}
\newcommand\tgt[1] {\emph{\textcolor{tgt}{#1}}}
\newcommand\gt[1] {\emph{\textcolor{gt}{#1}}}

\title{Neural Pose Representation Learning for Generating and Transferring Non-Rigid Object Poses}

\author{%
  Seungwoo Yoo
    $\quad$
  Juil Koo
    $\quad$
  Kyeongmin Yeo
    $\quad$
  Minhyuk Sung \\
  KAIST \\
  \texttt{\{dreamy1534,63days,aaaaa,mhsung\}@kaist.ac.kr}
}

\begin{document}

\maketitle

\begin{figure*}[!h]
    \vspace{-2\baselineskip}
    \begin{minipage}{0.49\textwidth}
        \centering
        \includegraphics[width=\linewidth]{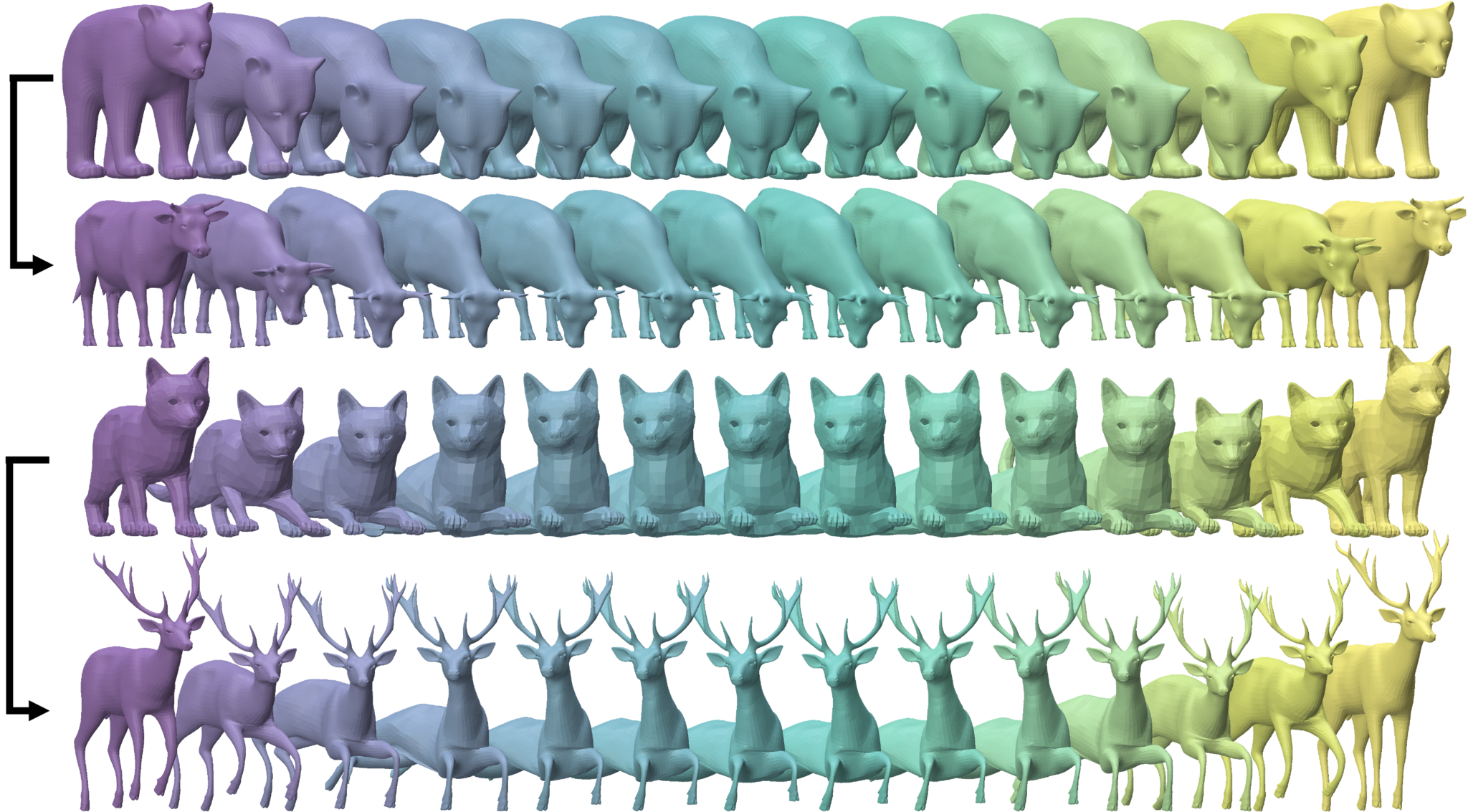}
    \end{minipage}\hfill%
    \begin{minipage}{0.49\textwidth}
        \centering
        \includegraphics[width=\linewidth]{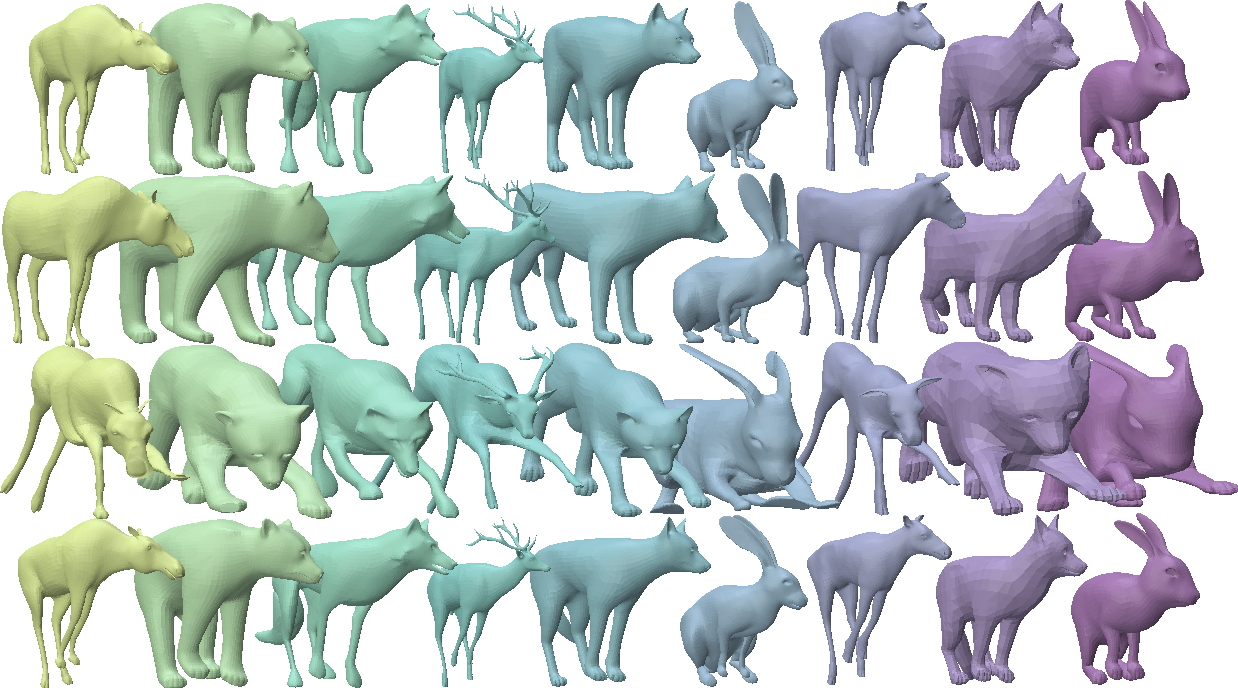}
    \end{minipage}
    \caption{
    Results of motion sequence transfer (left) and shape variation generation (right) using the proposed neural pose representation. On the left, poses from source shapes (first and third rows) are transferred to target shapes (second and fourth rows), preserving intricate details like horns and antlers. On the right, new poses sampled from a cascaded diffusion model, trained with shape variations of the bunny (last column), are transferred to other animal shapes.
    }
    \label{fig:teaser}
\end{figure*}

\begin{abstract}
\vspace{-0.5\baselineskip}
We propose a novel method for learning representations of poses for 3D deformable objects, which specializes in 1) disentangling pose information from the object's identity, 2) facilitating the learning of pose variations, and 3) transferring pose information to other object identities. Based on these properties, our method enables the generation of 3D deformable objects with diversity in both identities and poses, using variations of a single object. It does not require explicit shape parameterization such as skeletons or joints, point-level or shape-level correspondence supervision, or variations of the target object for pose transfer.
We first design the pose extractor to represent the pose as a keypoint-based hybrid representation and the pose applier to learn an implicit deformation field. To better distill pose information from the object's geometry, we propose the implicit pose applier to output an intrinsic mesh property, the face Jacobian. Once the extracted pose information is transferred to the target object, the pose applier is fine-tuned in a self-supervised manner to better describe the target object's shapes with pose variations. The extracted poses are also used to train a cascaded diffusion model to enable the generation of novel poses.
Our experiments with the DeformThings4D and Human datasets demonstrate state-of-the-art performance in pose transfer and the ability to generate diverse deformed shapes with various objects and poses.
\vspace{-1\baselineskip}
\end{abstract}

\section{Introduction}
\label{sec:intro}
\vspace{-\baselineskip}
The recent great success of generative models~\cite{Ho:2019DDPM, Song:2019Gradient, Song:2021SDE, Song:2021DDIM} has been made possible not only due to advances in techniques but also due to the enormous scale of data that has become available, such as LAION~\cite{Schuhmann2022:LAION5B} for 2D image generation. For 3D data, the scale has been rapidly increasing, as exemplified by ObjaverseXL~\cite{Deitke2024:ObjaverseXL}. However, it is still far from sufficient to cover all possible 3D shapes, particularly deformable, non-rigid 3D shapes such as humans, animals, and characters. The challenge with deformable 3D shapes is especially pronounced due to the diversity in \emph{both} the \emph{identities} and \emph{poses} of the objects. Additionally, for a new 3D character created by a designer, information about possible variations of the creature does not even exist.

To remedy the requirement of a large-scale dataset for 3D deformable shape generation, we aim to answer the following question: \ul{Given variations of a \emph{single} deformable object with its different poses, how can we effectively learn the pose variations while factoring out the object's identity and also make the pose information applicable to other objects?} For instance, when we have a variety of poses of a bear (\fig~\ref{fig:teaser} left, first row), our objective is to learn the space of poses without entangling them with the geometric characteristics of the bears. Also, we aim to enable a sample from this space to be applied to a new object, such as a bull, to generate a new shape (\fig~\ref{fig:teaser} left, second row).
We believe that such a technique, effectively separating pose from the object's identity and enabling the transfer of poses to other identities, can significantly reduce the need for collecting large-scale datasets covering the diversity of both object identities and poses. This approach can even enable creating variations of a new creature without having seen any possible poses of that specific object.

Transferring poses from one object to another has been extensively studied in computer graphics and vision, with most methods requiring target shape supervision~\cite{Sumner:2004Transfer,BenChen:2009Spatial,Yang:2018Biharmonic,Baran:2009Semantic,Gao:2018VCGAN} or predefined pose parameterization~\cite{Gleicher:1998Retargeting,Choi:1999Online,Lee:1999Hierarchical,Tak:2005RetargetingFilter,AlBorno:2018PhysicsBased,Yamane:2010NonHumanoid,Rhodin:2015Generalizing,Delhaisse:2017Transfer,Villegas:2018NeuralKinematic,Jang2018:VariationalUNet,Lim:2019PMNet,Villegas:2021ContactAware,Chen:2023WeaklySupervised}. Without such additional supervision, our key idea for extracting identity-agnostic pose information and learning their variations is to introduce a novel pose representation along with associated encoding and decoding techniques. For this, we consider the following three desiderata:
\begin{enumerate}[leftmargin=*,noitemsep,topsep=0em]
\item \textbf{Pose Disentanglement}: The representation should effectively represent the pose only without resembling the source object's identity when applied to the other object. 
\item \textbf{Compactness}: The representation should be compact enough to effectively learn its variation using a generative model, such as a diffusion model.
\item \textbf{Transferability}: The encoded pose information should be applicable to new target objects.
\end{enumerate}

As a representation that meets the aforementioned criteria, we propose an autoencoding framework and a latent diffusion model with three core components. Firstly, we design a \emph{pose extractor} and a \emph{pose applier} to encode an \textbf{implicit deformation field} with a \textbf{keypoint-based hybrid representation}, comprising 100 keypoints in the space, each associated with a latent feature. Learning the deformation field enables \emph{disentangling} the pose information from the object's identity, while the keypoint-based representation \emph{compactly} encodes it and makes it \emph{transferable} to other objects. However, simply learning the deformation as a new position of the vertex is not sufficient to properly adapt the source object's pose information to others. Hence, secondly, we propose predicting an \textbf{intrinsic property} of the deformed mesh, \textbf{Jacobian fields}~\cite{Yu:2004Poisson,Lipman:2004Differential,Sorkine:2004LaplacianSurfaceEditing,Aigerman:2022NJF}, which can successfully apply the pose while preserving the identity of the target shape. To better preserve the target's identity while applying the pose variation from the source, thirdly, we propose a \textbf{per-identity refinement step} that fine-tunes the decoder in a \emph{self-supervised} way to adapt to the variations of target shapes, with poses transferred from the source object. Thanks to the compact hybrid representation of pose, a pose generative model can also be effectively learned using \textbf{cascaded diffusion models}~\cite{Ho:2021Cascaded,Koo:2023Salad}, enabling the generation of varying poses of an object with an arbitrary identity different from the source object.

In our experiments, we compare our framework against state-of-the-art techniques for pose transfer on animals (\shortsec~\ref{subsec:quadruped_transfer}) and humans (\shortsec~\ref{subsec:human_transfer}). Both qualitative and quantitative analyses underscore the key design factors of our framework, demonstrating its efficiency in capturing identity-agnostic poses from exemplars and its superior performance compared to existing methods. Additionally, we extend the proposed representation to the task of unconditional generation of shape variations. Our representation serves as a compact encoding of poses that can be generated using diffusion models (\shortsec~\ref{subsec:uncond_gen}) and subsequently transferred to other shapes. This approach facilitates the generation of various shapes, particularly in categories where exemplar collection is challenging.

\vspace{-0.5\baselineskip}
\section{Related Work}
\label{sec:related_work}
\vspace{-1\baselineskip}

Due to space constraints, we focus on reviewing the literature on non-rigid shape pose transfer, including methods that operate without parameterizations, those that rely on predefined parameterizations, and recent learning-based techniques that derive parameterizations from data.

\vspace{-\baselineskip}
\paragraph{Parameterization-Free Pose Transfer.}
Early works~\cite{Sumner:2004Transfer,Xu:2007Gradient,BenChen:2009Spatial,Yang:2018Biharmonic} focused on leveraging \emph{point-wise} correspondences between source and target shapes. A seminal work by~\citet{Sumner:2004Transfer} transfers per-triangle affine transforms applied to the target shape by solving an optimization problem. A follow-up work by Ben-Chen~\etal~\cite{BenChen:2009Spatial} transfers deformation gradients by approximating source deformations using harmonic bases. On the other hand, a technique proposed by Baran~\etal~\cite{Baran:2009Semantic} instead employs \emph{pose-wise} correspondences by learning shape spaces from given pairs of poses shared across the source and target identities. The poses are transferred by blending existing exemplars.
While these techniques require point-wise or pose-wise correspondence supervision, our method does not require such supervision during training or inference.

\vspace{-1.5\baselineskip}
\paragraph{Skeleton- or Joint-Based Pose Transfer.} Another line of work utilizes handcrafted skeletons, which facilitate pose transfer via motion retargeting~\cite{Gleicher:1998Retargeting}. This approach has been extended by incorporating physical constraints~\cite{Choi:1999Online,Lee:1999Hierarchical,Tak:2005RetargetingFilter,AlBorno:2018PhysicsBased} or generalizing the framework to arbitrary objects~\cite{Yamane:2010NonHumanoid,Rhodin:2015Generalizing}. Several learning-based methods~\cite{Delhaisse:2017Transfer,Villegas:2018NeuralKinematic,Jang2018:VariationalUNet,Lim:2019PMNet,Villegas:2021ContactAware} have also been proposed to predict joint transformations involved in forward kinematics from examples.
Recently, Chen~\etal~\cite{Chen:2023WeaklySupervised} proposed a framework that does not require skeletons during test time by predicting keypoints at joints. The method is trained to predict both relative transformations between corresponding keypoints in two distinct kinematic trees and skinning weights. However, the tasks of rigging and skinning are labor-intensive, and different characters and creatures often require distinct rigs with varying topologies.
Liao~\etal~\cite{Liao:2022SPT} notably presented a representation that comprises character-agnostic deformation parts and a semi-supervised network predicting skinning weights that link each vertex to these deformation parts, although its performance hinges on accurate skinning weight prediction.
In this work, we design a more versatile framework that is applicable to various shapes and provides better performance.

\vspace{-1.5\baselineskip}
\paragraph{Pose Transfer via Learned Parameterization.}
To bypass the need for correspondence or parameterization supervision, learning-based approaches~\cite{Gao:2018VCGAN,Yifan:2020NeuralCage,Wang:2020Neural,Zhou:2020Unsupervised,Chen:2021IEPGAN,Aigerman:2022NJF,Liao:2022SPT,Wang:2023ZPT,Song:2021Pose,Song:2023XDualNet} explore alternative parameterizations \emph{learned} from exemplars. Yifan~\etal~\cite{Yifan:2020NeuralCage} propose to predict source and target cages and their offsets simultaneously, although their method still requires manual landmark annotations.
Gao~\etal~\cite{Gao:2018VCGAN} introduce a VAE-GAN framework that takes \emph{unpaired} source and target shape sets, each containing its own set of pose variations. The network is trained without direct pose-wise correspondences between samples from these sets, instead enforcing cycle consistency between latent representations. Although this work relaxes the requirement for correspondence supervision, it still requires pose variations for \emph{both} the source and target identities and individual training for each new source-target pair.
Numerous works~\cite{Wang:2020Neural, Chen:2021IEPGAN, Zhou:2020Unsupervised, Aigerman:2022NJF} lift the requirement for gathering variations of target shapes by disentangling identities from poses, enforcing cycle consistency~\cite{Zhou:2020Unsupervised}, or adapting conditional normalization layers~\cite{Wang:2020Neural} from image style transfer~\cite{Park:2019SPADE}. Notably, Aigerman~\etal~\cite{Aigerman:2022NJF} train a network that regresses Jacobian fields from SMPL~\cite{Loper:2015SMPL} pose parameters. The vertex coordinates are computed by solving Poisson's equation~\cite{Yu:2004Poisson}, effectively preserving the shapes' local details. Wang~\etal~\cite{Wang:2023ZPT} also train a neural implicit function and retrieve a shape latent from a template mesh of an unseen identity via autodecoding~\cite{Park:2019DeepSDF}. This method models local deformations through a coordinate-based network that learns continuous deformation fields. 
However, such methods struggle to generalize to unseen identities due to their reliance on \emph{global} latent embeddings encoding shapes.
We propose a representation that not only disentangles poses from identities but also allows for implicit queries using the surface points, thereby improving generalization to new identities.

\vspace{-0.5\baselineskip}
\section{Method}
\label{sec:method}
\vspace{-0.5\baselineskip}

\begin{figure}
\includegraphics[width=\textwidth]{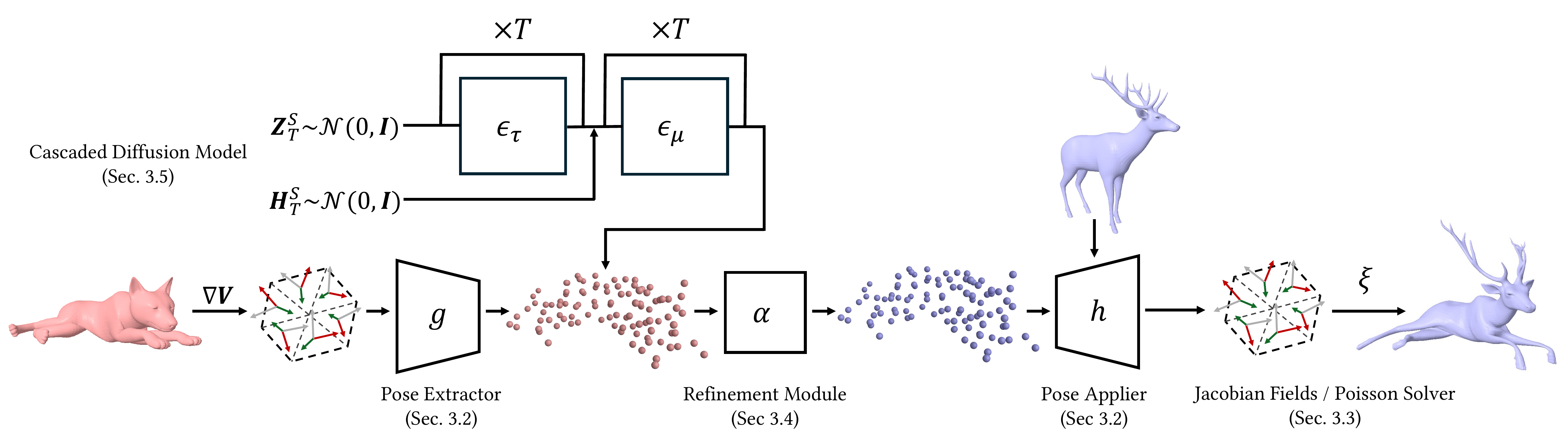}
\vspace{-1.5\baselineskip}
\caption{\textbf{Method overview.}
Our framework extracts keypoint-based hybrid pose representations from Jacobian fields. These fields are mapped by the pose extractor $g$ and mapped back by the pose applier $h$. The pose applier, conditioned on the extracted pose, acts as an implicit deformation field for various shapes, including those unseen during training. A refinement module $\alpha$, positioned between $g$ and $h$, is trained in a self-supervised manner, leveraging the target's template shape. The compactness of our latent representations facilitates the training of a diffusion model, enabling diverse pose variations through generative modeling in the latent space.
}
\label{fig:pipeline}
\vspace{-1.5\baselineskip}
\end{figure}

\subsection{Problem Definition}
\label{subsec:prob_def}
\vspace{-0.5\baselineskip}
Consider a \emph{source template mesh} $\overline{\mathcal{M}}^S = (\overline{\mathbf{V}}^S, \mathbf{F}^S)$, given as a 2-manifold triangular mesh. The mesh comprises vertices $\overline{\mathbf{V}}^S$ and faces $\mathbf{F}^S$. Suppose there exist $N$ variations of the source template mesh, $\{ \mathcal{M}_1^S, \dots, \mathcal{M}_N^S \}$, where each $\mathcal{M}_i^S = (\mathbf{V}_i^S, \mathbf{F}^S)$ is constructed with a different \emph{pose}, altering the vertex positions while sharing the same mesh connectivity $\mathbf{F}^S$.

Assume a \emph{target template mesh} $\overline{\mathcal{M}}^{T} = (\overline{\mathbf{V}}^T, \mathbf{F}^{T})$ is given without any information about its variations or existing pose parameterization (\eg, skeletons or joints). Our goal is to define functions $g$ and $h$ that can \textit{transfer} the pose variations from the source meshes to the target template mesh.
For each variation of the source shape $\mathcal{M}^S_i$, its corresponding mesh $\mathcal{M}_i^T$ for the target is obtained as:
\begin{align}
    \mathcal{M}_{i}^{T} &= ( h ( g( \mathcal{M}_i^S ), \overline{\mathcal{M}}^T ), \mathbf{F}^T ),\quad \text{for } i=1, 2, \cdots, N.
\end{align}

Specifically, we design $g$ as a \emph{pose extractor} that embeds a source object mesh $\mathcal{M}_i^S$ into a \emph{pose latent representation} $\mathcal{Z}_i^S = g(\mathcal{M}_i^S)$. This representation disentangles the pose information from the object's identity in $\mathcal{M}_i^S$ and facilitates transferring the pose to the target template mesh $\overline{\mathcal{M}}^T$.
Given this pose representation, the \emph{pose applier} $h$ then applies the pose to $\overline{\mathcal{M}}^T$, yielding the corresponding variation of the target object $\mathcal{M}_i^T = h (\mathcal{Z}^S_i, \overline{\mathcal{M}}^T)$.
Note that our method is not limited to transferring the pose of a \emph{given} variation of the source object to the target mesh but can also apply a pose \emph{generated} by a diffusion model to the target mesh. In \shortsec~\ref{subsec:jacobian_ldm}, we explain how a diffusion model can be trained with the latent pose representation extracted from source object variations.
In the following, we first describe the key design factors of the functions $g$ and $h$ to tackle the problem.

\vspace{-0.5\baselineskip}
\subsection{Keypoint-Based Hybrid Pose Representation}
\label{subsec:identity_agnostic}
\vspace{-0.5\baselineskip}

To encode a source shape $\mathcal{M}^S$ into a latent representation $\mathcal{Z}^S$, we consider its vertices $\mathbf{V}^S$ as its geometric representation and use them as input to the pose extractor $g$, which is designed as a sequence of Point Transformer~\citep{Zhao:2021PointTransformer,Tang:2022NSDP} layers. These layers integrate vector attention mechanisms~\citep{Zhao:2020SAN} with progressive downsampling of input point clouds.
The output of the pose extractor $g$ is a set of unordered tuples $\mathcal{Z}^S = \left\{(\mathbf{z}_k^S, \mathbf{h}_k^S)\right\}_{k=1}^{K}$ where $\mathbf{z}^S_k \in \mathbb{R}^{3}$ represents a 3D coordinate of a keypoint subsampled from $\mathbf{V}^S$ via farthest point sampling (FPS) and $\mathbf{h}^S_k$ is a learned feature associated with $\mathbf{z}^S_k$. This set $\mathcal{Z}^S$ forms a sparse point cloud of keypoints in 3D space, augmented with latent features. We set $K=100$ in our experiments.

This keypoint-based hybrid representation, visualized in \fig~\ref{fig:pipeline}, is designed to exclusively transfer pose information from the source to the target while preventing leakage of the source shape's identity characteristics. Since the keypoints $\{\mathbf{z}^S_k\}$ are sampled from $\mathbf{V}^S$ using FPS, they effectively capture the overall pose structure of $\mathbf{V}^S$ while also supporting geometric queries with the vertices of a new mesh. This property is essential during the decoding phase, where the pose applier $h$ predicts the pose-applied mesh from the input template as an \emph{implicit deformation field}.

The pose applier $h$ is implemented with a neural network that takes the 3D coordinates of a vertex from the input template mesh as a query, along with the hybrid pose latent representation $\mathcal{Z}$, and outputs the new position of the vertex in the pose-applied deformed mesh. Note that $h$ indicates a function that collectively maps all vertices in the input template mesh to their new positions using the network. Like the pose extractor $g$, the implicit deformation network is also parameterized as Point Transformer layers~\citep{Zhao:2021PointTransformer}. It integrates the pose information encoded in $\mathcal{Z}^S$ by combining vector attention mechanisms with nearest neighbor queries to aggregate features of the keypoints ${\mathbf{z}^S_k}$ around each query point. The aggregated features are then decoded by an MLP to predict the vertex coordinates of the deformed shape. (This is a base network, and we also introduce a \emph{better} way to design the implicit deformation network in \shortsec~\ref{subsec:jacobian_fields}.)

Given only the variations of the source object $\{ \mathcal{M}_1^S, \dots, \mathcal{M}_N^S \}$, we jointly learn the functions $g$ and $h$ by reconstructing the variations of the source object as a deformation of its template:
\begin{align}
    \mathcal{L}_{V} = \Vert \mathbf{V}_i^S - h ( g( \mathcal{M}_i^S ), \overline{\mathcal{M}}^S ) \Vert^2.
\end{align}
While $g$ and $h$ are trained using the known variations of the source object $\mathcal{M}^S$, the latent representation $\mathcal{Z}^S = g(\mathcal{M}^S)$, when queried and decoded with the target template mesh, effectively transfers the pose extracted by $g$ from $\mathcal{M}^S$. However, we also observe that $g$ and $h$, when trained using the loss $\mathcal{L}_V$, often result in geometry with noticeable imperfections and noise on the surfaces. To address this, we explore an alternative representation of a mesh that better captures and preserves geometric details, which will be discussed in the following section.

\subsection{Representing Shapes as Jacobian Fields}
\label{subsec:jacobian_fields}
\vspace{-0.5\baselineskip}
In this work, we advocate employing the differential properties of surfaces as dual representations of a mesh.
Of particular interest are \emph{Jacobian fields}, a gradient-domain representation noted for its efficacy in preserving local geometric details during deformations~\citep{Aigerman:2022NJF}, while ensuring that the resulting surfaces maintain smoothness~\citep{Gao:2023TextDeformer}.

Given a mesh $\mathcal{M} = (\mathbf{V}, \mathbf{F} )$, a Jacobian field $\mathbf{J}$ represents the spatial derivative of a scalar-valued function $\mathcal{\phi}: \mathcal{M} \to \mathbb{R}$ defined over the surface. We discretize $\mathcal{\phi}$ as $\mathcal{\phi}_{\mathbf{V}} \in \mathbb{R}^{\vert \mathbf{V} \vert}$, sampling its value at each vertex $\mathbf{v}$ of the vertex set $\mathbf{V}$. The spatial derivative of $\mathcal{\phi}$ at each triangle $\mathbf{f} \in \mathbf{F}$ is computed as $\boldsymbol{\nabla}_{\mathbf{f}} \mathcal{\phi}_{\mathbf{V}}$ using the per-triangle gradient operator $\boldsymbol{\nabla}_{\mathbf{f}}$.
Given that each dimension of vertex coordinates $\mathbf{V}$ is such a function, we compute its spatial gradient at each triangle $\mathbf{f}$ as $\mathbf{J}_{\mathbf{f}} = \boldsymbol{\nabla}_{\mathbf{f}} \mathbf{V}$. Iterating this process for all triangles yields the Jacobian field $\mathbf{J} = \left\{\mathbf{J}_{\mathbf{f}} \vert \mathbf{f} \in \mathbf{F} \right\}$.

To recover $\mathbf{V}$ from a given Jacobian field $\mathbf{J}$, we solve a least-squares problem, referred to as Poisson's equation:
\begin{align}
    \label{eqn:poisson_unconstrained}
    \mathbf{V}^{*} &= \underset{\mathbf{V}}{\text{argmin}} \Vert \mathbf{L} \mathbf{V} - \boldsymbol{\nabla}^T \mathcal{A} \mathbf{J} \Vert^2,
\end{align}
where $\mathbf{L} \in \mathbb{R}^{\vert \mathbf{V} \vert \times \vert \mathbf{V} \vert}$ is the cotangent Laplacian of $\mathcal{M}$, $\boldsymbol{\nabla}$ is the stack of gradient operators defined at each $\mathbf{f} \in \mathbf{F}$, and $\mathcal{A} \in \mathbb{R}^{3 \vert \mathbf{F} \vert \times 3 \vert \mathbf{F} \vert}$ is the diagonal mass matrix, respectively.
Since the rank of $\mathbf{L}$ is at most $|\mathbf{V}| - 1$, we can obtain the solution by fixing a single point, which is equivalent to eliminating one row of the system in \eqn~\ref{eqn:poisson_unconstrained}.
Since $\mathbf{L}$ in \eqn~\ref{eqn:poisson_unconstrained} remains constant for a given shape $\mathcal{M}$, we can prefactorize the matrix (\eg using Cholesky decomposition) and quickly solve the system for different Jacobian fields $\mathbf{J}$'s. Furthermore, the upstream gradients can be propagated through the solver since it involves only matrix multiplications~\citep{Aigerman:2022NJF}.

Employing Jacobian fields as shape representations, we now modify the implicit deformation network described in \shortsec~\ref{subsec:identity_agnostic} to take \emph{face center coordinates} as input instead of vertex coordinates and to output a new \emph{face Jacobian} for a query face instead of a new vertex position. This results in decomposing $h$ into two functions $h = (\xi \circ h^{\prime})$, where $h^{\prime}$ is a function that collectively maps all the faces in the input mesh to the new Jacobian, and $\xi$ is a differentiable Poisson solver layer. Both $g$ and $h^{\prime}$ are then trained by optimizing the following loss:
\begin{align}
    \mathcal{L}_{J} &= \Vert \mathbf{V}_i^S - \xi ( h^{\prime} ( g( \mathcal{M}_i^S ), \overline{\mathcal{M}}^S ) ) \Vert^2.
\end{align}

\subsection{Per-Identity Refinement using Geometric Losses}
\label{subsec:refinement}
\vspace{-0.5\baselineskip}
While the latent pose representation $\mathcal{Z}$ learned by $g$ and $h$ exhibits promising generalization capabilities in transferring poses, the quality of the transferred shapes can be further improved by incorporating a trainable, identity-specific refinement module into our system. This module is trained in a \emph{self-supervised} manner with the set of pose-applied target meshes.
Similarly to techniques for personalized image generation~\cite{Hu2022:LoRA,Ye2023:IPAdapter}, we introduce a shallow network $\alpha$ between $g$ and $h$, optimizing its parameters while keeping the rest of the pipeline frozen.

The optimization of $\alpha$ is driven by geometric losses, aiming to minimize the geometric discrepancies in terms of the object's identity between the target template mesh $\overline{\mathcal{M}}^{T}$ and the pose-transferred meshes $\mathcal{M}^T_i$.
In particular, we first extract poses $\left\{\mathcal{Z}^S_1, \dots, \mathcal{Z}^S_N\right\}$ corresponding to the known shapes $\left\{\mathcal{M}_1^S, \dots, \mathcal{M}_N^S\right\}$ of the source object. A transformer-based network $\alpha$, which maps a latent representation $\mathcal{Z}^S$ to $\mathcal{Z}^{S \prime}$, is plugged in between the pose extractor $g$ and the pose applier $h$ :
\begin{align}
    \mathbf{V}_i^T &= h ( \alpha ( \mathcal{Z}_i^S ), \overline{\mathcal{M}}^T ).
\end{align}

The parameters of $\alpha$ are updated by optimizing the following loss function:
\begin{align}
    \mathcal{L}_{\text{ref}} = \lambda_{\text{lap}} \mathcal{L}_{\text{lap}} (\mathbf{V}_i^{T}, \overline{\mathbf{V}}^{T}) + \lambda_{\text{edge}} \mathcal{L}_{\text{edge}} ( \mathbf{V}_i^{T}, \overline{\mathbf{V}}^{T} ) + \lambda_{\text{reg}} ( \sum_{k} \Vert \mathbf{z}_k^S - \mathbf{z}_k^{S \prime} \Vert^2 + \Vert \mathbf{h}_k^S- \mathbf{h}_k^{S \prime} \Vert^2 ),
\end{align}
where $\mathcal{L}_{\text{lap}} (\cdot)$ is the mesh Laplacian loss~\cite{Liu:2021DeepMetaHandles}, $\mathcal{L}_{\text{edge}} (\cdot)$ is the edge length preservation loss~\citep{Liao:2022SPT}, and $\lambda_{\text{lap}}$, $\lambda_{\text{edge}}$, and $\lambda_{\text{reg}}$ are the weights of the loss terms. The definitions of the losses are provided in the appendix. Note that this refinement step leverages only the originally provided template shape $\overline{\mathcal{M}}^T$ and does not require its given variations or any other additional supervision.

\vspace{-0.5\baselineskip}
\subsection{Learning Latent Diffusion via Cascaded Training}
\label{subsec:jacobian_ldm}
\vspace{-0.5\baselineskip}
The use of the keypoint-based hybrid representation discussed in \shortsec~\ref{subsec:identity_agnostic} offers a compact latent space suitable for generative modeling using diffusion models~\citep{Song:2019Gradient,Ho:2019DDPM,Song:2021SDE,Song:2021DDIM}. Unlike the Jacobian fields with dimensionality $\vert \mathbf{F} \vert \times 9$, the keypoints and their feature vectors that comprise the pose representation $\mathcal{Z}^S$ lie in significantly lower dimensional space, facilitating generative modeling with latent diffusion models~\cite{Rombach:2022LDM}.

We employ a cascaded diffusion framework~\citep{Ho:2021Cascaded,Koo:2023Salad} to separately capture the layouts of keypoints and the associated feature vectors. Given a set $\{\mathcal{Z}^S_1, \dots, \mathcal{Z}^S_N \}$ of $N$ latent embeddings extracted from the known source shape variations $\left\{ \mathcal{M}_1^{S}, \dots, \mathcal{M}_N^S \right\}$, we first learn the distribution over $\mathbf{Z}^S = \left\{ \mathbf{z}^S_k \right\}_{k=1}^{K}$. To handle unordered sets with small cardinality, we employ a transformer-based network to facilitate interactions between each element within the noise prediction network $\boldsymbol{\epsilon}_{\tau} (\mathbf{Z}^S_t, t )$ where $t$ is a diffusion timestep and $\mathbf{Z}^S_t$ is a noisy 3D point cloud obtained by perturbing a clean keypoint set $\mathbf{Z}^S_0 \left(= \mathbf{Z}^S\right)$ via forward diffusion process~\cite{Ho:2019DDPM}. We train the network by optimizing the denoising loss:
\begin{align}
    \mathcal{L}_{\mathbf{Z}^S} &= \mathbb{E}_{\mathbf{Z}^S, \boldsymbol{\epsilon}~\sim~\mathcal{N}(\mathbf{0}, \mathbf{I}), t~\sim~\mathcal{U}(0, 1)} \left[\Vert \boldsymbol{\epsilon} - \boldsymbol{\epsilon}_{\tau} (\mathbf{Z}^S_t, t) \Vert^2 \right].
\end{align}

Likewise, the distribution of the set of latent features $\mathbf{H}^S = \left\{ \mathbf{h}^S_k \right\}_{k=1}^{K}$ is modeled as a conditional diffusion model $\boldsymbol{\epsilon}_{\mu}$, which takes $\mathbf{Z}^S$ as an additional input to capture the correlation between $\mathbf{Z}^S$ and $\mathbf{H}^S$. The network is trained using the same denoising loss:
\begin{align}
    \mathcal{L}_{\mathbf{H}^S} &= \mathbb{E}_{\mathbf{H}^S, \boldsymbol{\epsilon}~\sim~\mathcal{N}(\mathbf{0}, \mathbf{I}), t~\sim~\mathcal{U}(0, 1)} \left[\Vert \boldsymbol{\epsilon} - \boldsymbol{\epsilon}_{\mu} (\mathbf{H}^S_t, \mathbf{Z}^S, t) \Vert^2 \right].
\end{align}

Once trained, the models can sample new pose representations through the reverse diffusion steps~\cite{Ho:2019DDPM}:
\begin{align}
    \mathbf{Z}^S_{t-1} &= \frac{1}{\sqrt{\alpha_t}} \left( \mathbf{Z}_{t}^S - \frac{\beta_t}{\sqrt{1 - \bar{\alpha}_t}} \boldsymbol{\epsilon}_{\tau} \left(\mathbf{Z}_t^S, t\right) \right), \\
    \mathbf{H}^S_{t-1} &= \frac{1}{\sqrt{\alpha_t}} \left( \mathbf{H}_{t}^S - \frac{\beta_t}{\sqrt{1 - \bar{\alpha}_t}} \boldsymbol{\epsilon}_{\mu} \left(\mathbf{H}_t^S, \mathbf{Z}_0^S, t\right) \right),
\end{align}
where $\alpha_t$, $\bar{\alpha}_t$, and $\beta_t$ are the diffusion process coefficients.

\vspace{-0.5\baselineskip}
\section{Experiments}
\label{sec:experiments}
\vspace{-0.5\baselineskip}

\subsection{Experiment Setup}
\label{subsec:experiment_setup}
\vspace{-0.5\baselineskip}
\paragraph{Datasets.}
In our experiments, we consider animal and human shapes that are widely used in various applications. For the animal shapes, we utilize animation sequences from the \deformthings dataset~\cite{Li:2021Deform4D}. Specifically, we extract 300 meshes from the animation sequences of each of 9 different animal identities, spanning diverse species such as bears, rabbits, dogs, and moose. For humanoids, we use SMPL~\cite{Loper:2015SMPL, Pavlakos:2019SMPL-X}, which facilitates easy generation of synthetic data for both training and testing. We sample 300 random pose parameters from VPoser~\cite{Pavlakos:2019SMPL-X} to generate variations of an unclothed human figure using the \emph{default} body shape parameters, which are used to train our networks. For testing, we keep the pose parameters constant and sample 40 different body shapes from the parametric space covered by the unit Gaussian. This produces 40 different identities, each in 300 poses. The generated meshes serve as the ground truth for pose transfer. To assess the generalization capability to unusual identities that deviate significantly from the default body shape, we increase the standard deviation to 2.5 when sampling SMPL body parameters for 30 of the 40 identities. Additionally, we collect 9 stylized character meshes from the Mixamo dataset~\cite{Mixamo:2020} to test the generalizability of different methods.
For diffusion model training, the extracted keypoints and their associated features from the given set of source meshes are used as the training data for our cascaded diffusion model, which is trained separately for each identity.

\vspace{-0.5\baselineskip}
\paragraph{Baselines.} To assess the performance of pose transfer, we compare our method against \citet{Aigerman:2022NJF} (NJF), \citet{Liao:2022SPT} (SPT), \citet{Wang:2023ZPT} (ZPT), and various modifications of our framework.
For NJF~\cite{Aigerman:2022NJF}, we use the official code from the \textit{Morphing Humans} experiment, employing a PointNet~\cite{Qi:2016PointNet} encoder to map input shapes to global latents, and we train the model on our datasets.
For SPT~\cite{Liao:2022SPT}, we use the official code and pretrained model on humanoid shapes. Since a pretrained model for animal shapes is not provided, the comparison with SPT on the \deformthings dataset is omitted. For ZPT~\cite{Wang:2023ZPT}, the official implementation is not provided, so we implemented the model based on the description in the paper.
In our ablation study, we explore different variations of our method to assess their impact on performance, including: (1) using vertex coordinates as shape representations (as described in \shortsec~\ref{subsec:identity_agnostic}), and (2) omitting the per-identity refinement module (\shortsec~\ref{subsec:refinement}). All models (except for SPT~\cite{Liao:2022SPT}, for which we employ a pretrained model) are trained for \emph{each} shape identity.

\vspace{-0.5\baselineskip}
\paragraph{Evaluation Metrics.}
For pose transfer, when the corresponding shapes of the same pose are given for both source and target shapes, in the SMPL case, we measure accuracy using Point-wise Mesh Euclidean Distance (PMD)~\cite{Zhou:2020Unsupervised, Wang:2020Neural}, following our baselines\cite{Liao:2022SPT, Wang:2023ZPT}. Note that this measurement cannot be applied in the \deformthings case since pose-wise correspondences are not provided. For both pose transfer and shape generation (via pose generation), we measure the visual plausibility of the output meshes using FID~\cite{Heusel:2017FID}, KID~\cite{bińkowski:2018KID}, and ResNet classification accuracy with images rendered from four viewpoints (front, back, left, and right) without texture. For the latter, we train a ResNet-18~\cite{He:2015ResNet} network using 10,800 images rendered from the four viewpoints of all ground truth shape variations of each animal.

\vspace{-0.5\baselineskip}
\subsection{Pose Transfer on \deformthings}
\label{subsec:quadruped_transfer}
\vspace{-0.5\baselineskip}
We begin our experiments by transferring pose variations across different animals, a challenging task that necessitates strong generalization capabilities due to the diverse shapes of the animals involved.

\begin{table}[h!]
{
\vspace{-0.5\baselineskip}
    \footnotesize
    \setlength{\tabcolsep}{1.0pt}
    \centering
    {
    \begin{tabularx}{\textwidth}{YYYY|YYYY}
        \toprule
        & \multicolumn{3}{c|}{\deformthings~\cite{Li:2021Deform4D}} & \multicolumn{4}{c}{SMPL~\cite{Loper:2015SMPL}} \\
        & FID $\downarrow$ & KID $\downarrow$ & ResNet & PMD $\downarrow$ & FID $\downarrow$ & KID $\downarrow$ & ResNet \\
        & ($\times 10^{-2}$) & ($\times 10^{-2}$) & Acc. $\uparrow$ (\%) & ($\times 10^{-3}$) & ($\times 10^{-2}$) & ($\times 10^{-2}$) & Acc. $\uparrow$ (\%) \\
        \midrule
        NJF~\cite{Aigerman:2022NJF}  & 11.33 & 5.71 & 64.43 & 2.55 & 1.57 & 0.82 & 70.93 \\
        SPT~\cite{Liao:2022SPT} & - & - & - & 0.28 & 0.83 & 0.43 & 75.38 \\
        ZPT~\cite{Wang:2023ZPT} & 19.88 & 11.09 & 48.15 & 1.28 & 0.77 & 0.45 & 69.88 \\
        Ours & \textbf{1.11} & \textbf{0.42} & \textbf{78.72} & \textbf{0.13} & \textbf{0.30} & \textbf{0.19} & \textbf{79.09} \\
        \bottomrule
    \end{tabularx}
    }
}
\caption{Quantitative results on the experiments using the \deformthings dataset~\cite{Li:2021Deform4D} and the human shape dataset populated using SMPL~\cite{Loper:2015SMPL}.}
\label{tbl:deform4d_smpl_quantitative}
\vspace{-1.5\baselineskip}
\end{table}

\begin{figure*}[!h]
{
    \vspace{-0.5\baselineskip}
    \centering
    \setlength{\tabcolsep}{0em}
    \def\arraystretch{0.0}

    \begin{minipage}{0.49\textwidth}
        \centering
        {\footnotesize
        \begin{tabularx}{\linewidth}{YYYYY}
            \makecell{$\overline{\mathcal{M}}^T$} & \makecell{$\mathcal{M}^S$} & \makecell{NJF~\cite{Aigerman:2022NJF}} & \makecell{ZPT~\cite{Wang:2023ZPT}} & \makecell{Ours} \\
            \multicolumn{5}{c}{\includegraphics[width=\linewidth]{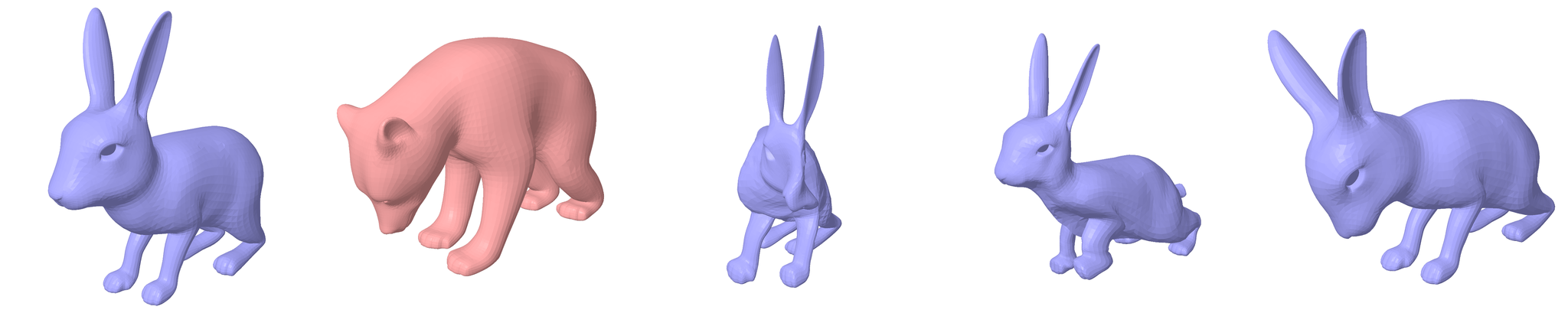}} \\
            \multicolumn{5}{c}{\includegraphics[width=\linewidth]{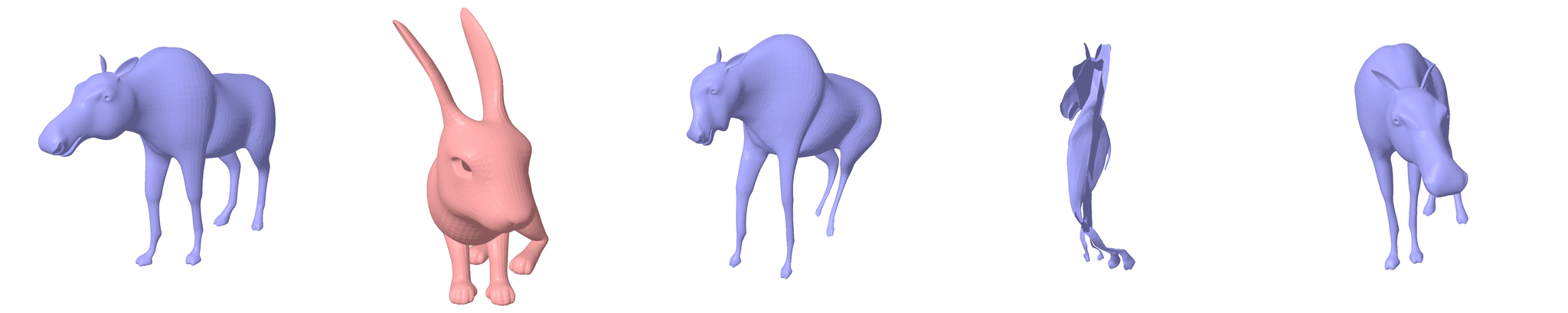}} \\
            \multicolumn{5}{c}{\includegraphics[width=\linewidth]{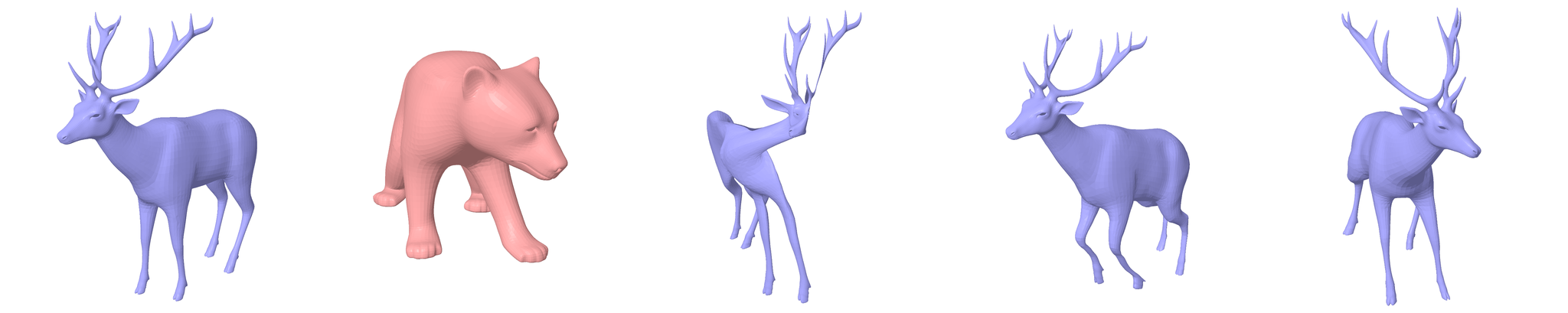}} \\
        \end{tabularx}
        }
    \end{minipage}\hfill\vline\hfill%
    \begin{minipage}{0.49\textwidth}
        \centering
        {\footnotesize
        \begin{tabularx}{\linewidth}{YYYYY}
            \makecell{$\overline{\mathcal{M}}^T$} & \makecell{$\mathcal{M}^S$} & \makecell{NJF~\cite{Aigerman:2022NJF}} & \makecell{ZPT~\cite{Wang:2023ZPT}} & \makecell{Ours} \\
            \multicolumn{5}{c}{\includegraphics[width=\linewidth]{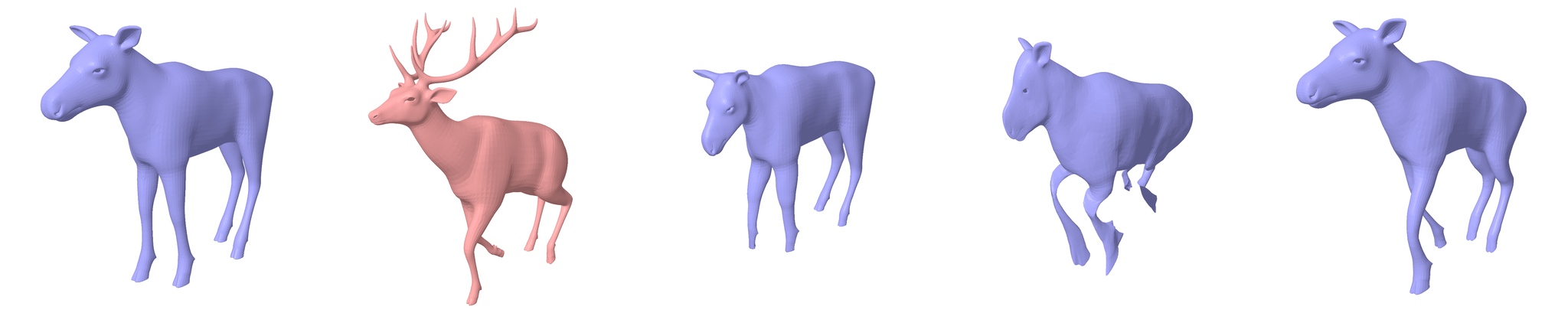}}
            \\
            \multicolumn{5}{c}{\includegraphics[width=\linewidth]{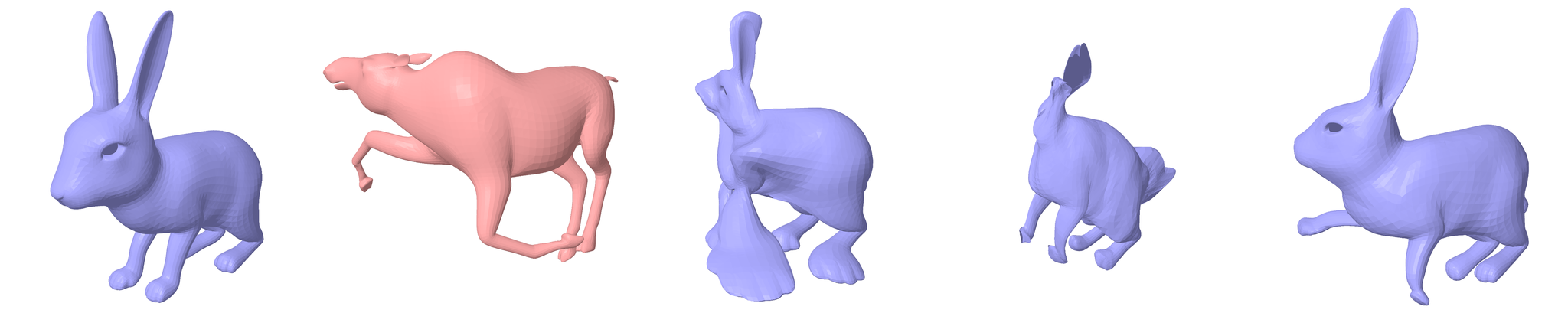}}
            \\
            \multicolumn{5}{c}{\includegraphics[width=\linewidth]{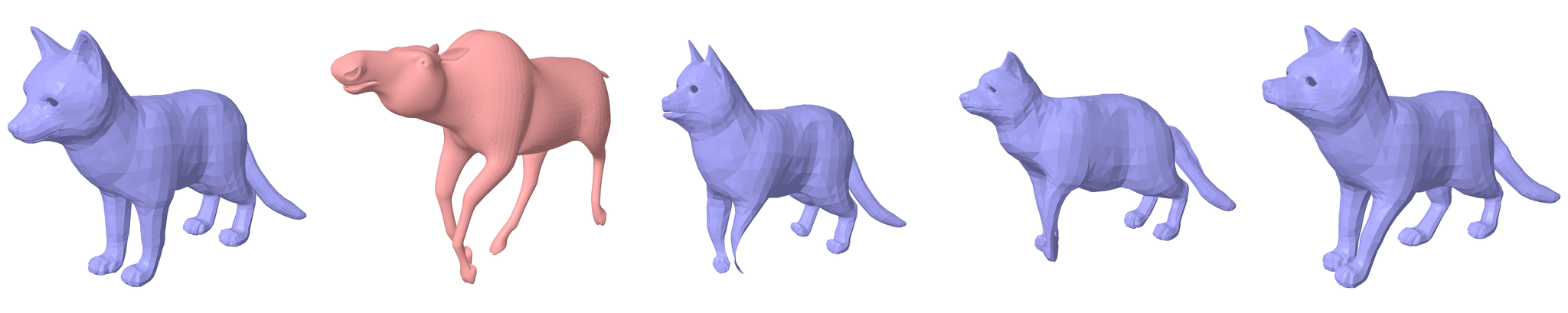}} \\
        \end{tabularx}
        }
    \end{minipage}
}
\caption{Qualitative results of transferring poses of the source meshes $\mathcal{M}^S$'s (\src{red}) in the DeformingThings4D animals~\cite{Li:2021Deform4D} to target templates $\overline{\mathcal{M}}^T$'s (\tgt{blue}). Best viewed when zoomed in.}
\label{fig:deform4d_transfer}
\end{figure*}

We summarize the quantitative metrics in \tab~\ref{tbl:deform4d_smpl_quantitative} (left). Our method outperforms NJF~\cite{Aigerman:2022NJF} and ZPT~\cite{Wang:2023ZPT}, both of which use global latent codes to encode shapes, while ours uses a keypoint-based hybrid representation. Note that SPT~\cite{Liao:2022SPT} is not compared in this experiment since the pretrained model is not provided for animal shapes. Qualitative results are also shown in \fig~\ref{fig:deform4d_transfer}, demonstrating the transfer of poses from a source mesh $\mathcal{M}^S$ (second and seventh column, \src{red}) to a target template mesh $\overline{\mathcal{M}}^T$ (first and sixth column, \tgt{blue}). Both NJF~\cite{Aigerman:2022NJF} and ZPT~\cite{Wang:2023ZPT} introduce significant distortions to the results and often fail to properly align the pose extracted from the source to the target. Our method, on the other hand, effectively transfers poses to the targets while preserving local geometric details. More results can be found in the appendix.

\vspace{-0.5\baselineskip}
\subsection{Pose Transfer on SMPL and Mixamo}
\label{subsec:human_transfer}
\vspace{-0.5\baselineskip}
We further test our method and baselines using humanoid shapes ranging from SMPL to stylized characters from the Mixamo~\cite{Mixamo:2020} dataset. While we employ the parametric body shape and pose model of SMPL~\cite{Loper:2015SMPL, Pavlakos:2019SMPL-X}, it is important to note that this is only for evaluation purposes; our method does not assume any parametric representations, such as skeletons, for either training or inference.

\begin{figure*}[!h]
    \vspace{-0.5\baselineskip}
    \centering
    \setlength{\tabcolsep}{0em}
    \def\arraystretch{0.0}
    {\footnotesize
    \begin{tabular}{P{0.092\textwidth}P{0.092\textwidth}P{0.163\textwidth}P{0.163\textwidth}P{0.163\textwidth}P{0.163\textwidth}P{0.164\textwidth}}
        $\overline{\mathcal{M}}^T$ & $\mathcal{M}^S$ & NJF~\cite{Aigerman:2022NJF} & SPT~\cite{Liao:2022SPT} & ZPT~\cite{Wang:2023ZPT} & Ours & $\mathcal{M}^T_{\text{GT}}$\\
        \multicolumn{7}{c}{\includegraphics[width=\textwidth]{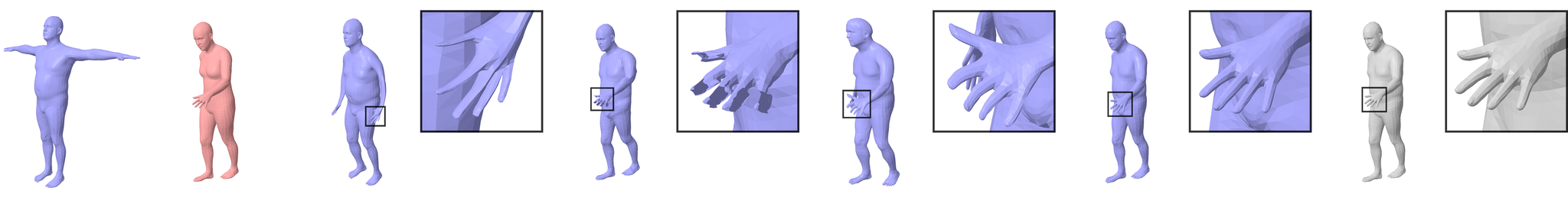}} \\
        \multicolumn{7}{c}{\includegraphics[width=\textwidth]{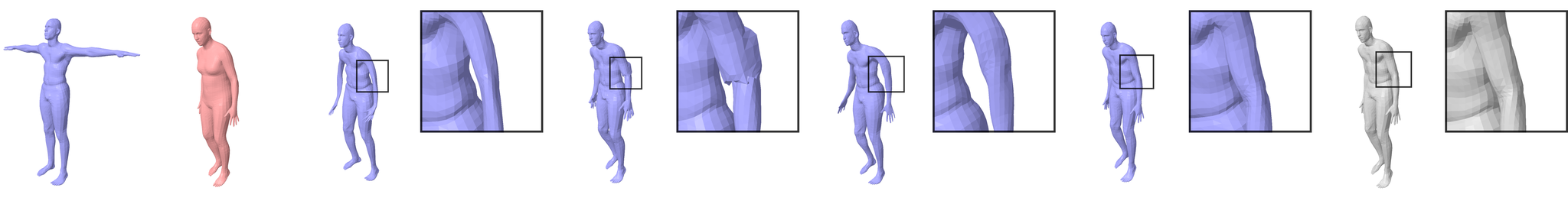}} \\
        \multicolumn{7}{c}{\includegraphics[width=\textwidth]{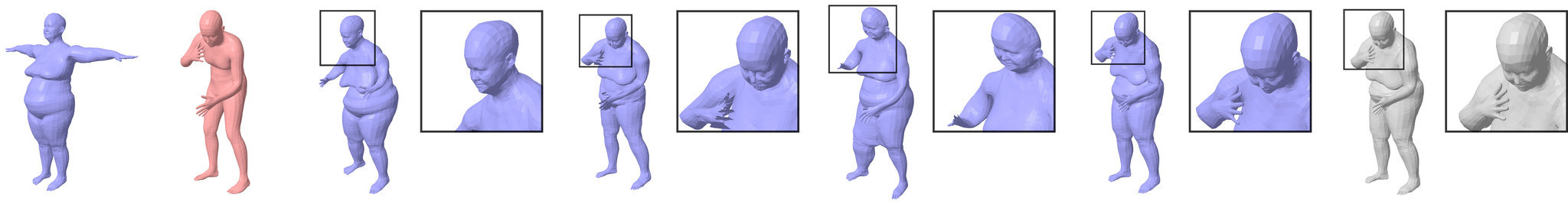}} \\
    \end{tabular}
    }
    \vspace{-0.5\baselineskip}
\caption{Qualitative results of transferring poses of the default human meshes $\mathcal{M}^S$'s (\src{red}) to different target template meshes $\overline{\mathcal{M}}^T$'s (\tgt{blue}). The ground truth targets $\mathcal{M}_{\text{GT}}^T$'s (\gt{grey}) are displayed for reference. Best viewed when zoomed in.}
\label{fig:smpl_transfer_error}
\vspace{-0.5\baselineskip}
\end{figure*}

\tab~\ref{tbl:deform4d_smpl_quantitative} (right) summarizes the evaluation metrics measured across the 40 target shapes. Notably, our method achieves lower PMD than SPT~\cite{Liao:2022SPT}, which is trained on a large-scale dataset consisting of diverse characters and poses, while ours is trained using only 300 pose variations of the default human body. This is further illustrated in the qualitative results in \fig~\ref{fig:smpl_transfer_error}, where we demonstrate pose transfer from source meshes (\src{red}) to target template meshes (\tgt{blue}) not seen during training. As shown, the shapes transferred by our method accurately match the overall poses. Our method benefits from combining a keypoint-based hybrid representation with Jacobian fields, outperforming the baselines in preserving local details, especially in areas with intricate geometric features such as the hands. See the zoomed-in views in~\fig~\ref{fig:smpl_transfer_error}. More results can be found in the appendix.

\begin{figure*}[!h]
    \vspace{-0.5\baselineskip}
    \centering
    \setlength{\tabcolsep}{0em}
    \def\arraystretch{0.0}
    {\footnotesize
    \begin{tabular}{P{0.1\textwidth}P{0.180\textwidth}P{0.180\textwidth}P{0.180\textwidth}P{0.180\textwidth}P{0.180\textwidth}}
        $\overline{\mathcal{M}}^T$ & $\mathcal{M}^S$ & NJF~\cite{Aigerman:2022NJF} & SPT~\cite{Liao:2022SPT} & ZPT~\cite{Wang:2023ZPT} & Ours \\
        \multicolumn{6}{c}{\includegraphics[width=\textwidth]{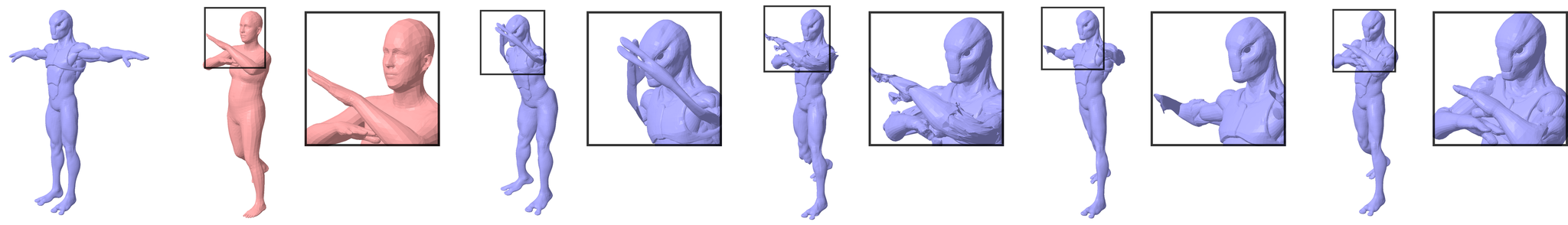}} \\
        \multicolumn{6}{c}{\includegraphics[width=\textwidth]{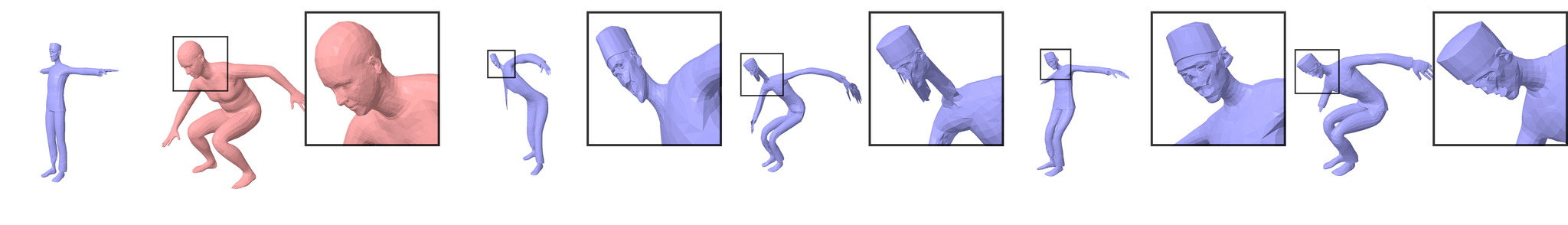}} \\
        \multicolumn{6}{c}{\includegraphics[width=\textwidth]{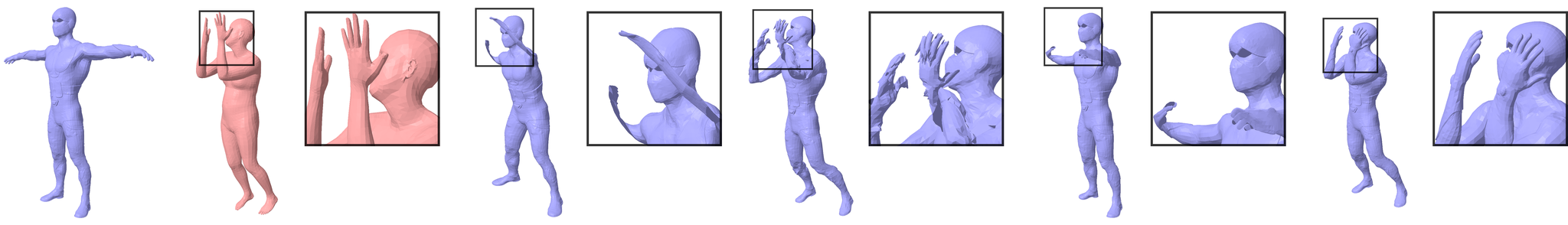}} \\
    \end{tabular}
    }
    \vspace{-0.5\baselineskip}
\caption{Qualitative results of transferring poses of the default human meshes $\mathcal{M}^S$'s (\src{red}) to target template meshes $\overline{\mathcal{M}}^T$'s (\tgt{blue}) of Mixamo characters~\cite{Mixamo:2020}. Best viewed when zoomed in.}
\label{fig:mixamo_transfer}
\end{figure*}

Furthermore, we apply our model to a more challenging setup involving stylized characters. In \fig~\ref{fig:mixamo_transfer}, we present qualitative results using shapes from the Mixamo~\cite{Mixamo:2020} dataset. Despite being trained on a single, unclothed SMPL body shape, our method generalizes well to stylized humanoid characters with detailed geometry (first row) and even to a character missing one arm (second row).

\vspace{-0.5\baselineskip}
\subsection{Ablation Study}
\label{subsec:ablation_study}
\vspace{-0.5\baselineskip}
Our framework design is further validated by comparisons against different variations of our framework, as listed in \shortsec~\ref{subsec:experiment_setup}, in the pose transfer experiment on animal shapes discussed in \shortsec~\ref{subsec:quadruped_transfer}. \tab~\ref{tbl:ablation_diffusion} (left) summarizes the image plausibility metrics measured using the results from our internal baselines.
Qualitative results are presented in \fig~\ref{fig:deform4d_ablation}.
Our method, which extracts pose representations from Jacobian fields and leverages the per-identity refinement module, achieves the best performance among all the variations.

\begin{table}[!h]
    {
    \footnotesize
    \setlength{\tabcolsep}{1.0pt}
    \begin{tabularx}{\textwidth}{YY|YYY|YYY}
        \toprule
        & & \multicolumn{3}{c|}{Poses from $\mathcal{M}_i^S$} & \multicolumn{3}{c}{Generated Poses (\shortsec~\ref{subsec:jacobian_ldm})} \\
        \multirow{2}{*}{\makecell{Jacobian \\ (\shortsec~\ref{subsec:jacobian_fields})}} & \multirow{2}{*}{\makecell{Refinement \\ (\shortsec~\ref{subsec:refinement})}} & FID $\downarrow$ & KID $\downarrow$ & ResNet & FID $\downarrow$ & KID $\downarrow$ & ResNet \\
        & & ($\times 10^{-2}$) & ($\times 10^{-2}$) & Acc. $\uparrow$ (\%) & ($\times 10^{-2}$) & ($\times 10^{-2}$) & Acc. $\uparrow$ (\%) \\
       \midrule
       \xmark & \xmark & 3.52 & 2.13 & 55.67 & 9.52 & 4.69 & 44.34 \\
       \cmark & \xmark & 1.17 & 0.47 & 75.13 & 4.40 & 2.39 & 75.82 \\
       \cmark & \cmark & \textbf{1.11} & \textbf{0.42} & \textbf{78.72} & \textbf{4.22} & \textbf{2.24} & \textbf{78.81} \\
        \bottomrule
    \end{tabularx}
    }
    \caption{Ablation study using the poses from the source shapes in \deformthings~\cite{Li:2021Deform4D} dataset (left) and the poses generated from our cascaded diffusion model.}
    \label{tbl:ablation_diffusion}
    \vspace{-1.5\baselineskip}
\end{table}

\begin{figure}[!h]
    \centering
    \setlength{\tabcolsep}{0em}
    \def\arraystretch{0.0}
    {\footnotesize
    \begin{tabular}{P{0.2\textwidth}P{0.125\textwidth}P{0.225\textwidth}P{0.225\textwidth}P{0.225\textwidth}}
        $\overline{\mathcal{M}}^T$ & $\mathcal{M}^S$ & Vertex Only & Jacobian Field Only & Ours \\
        \multicolumn{5}{c}{\includegraphics[width=\textwidth]{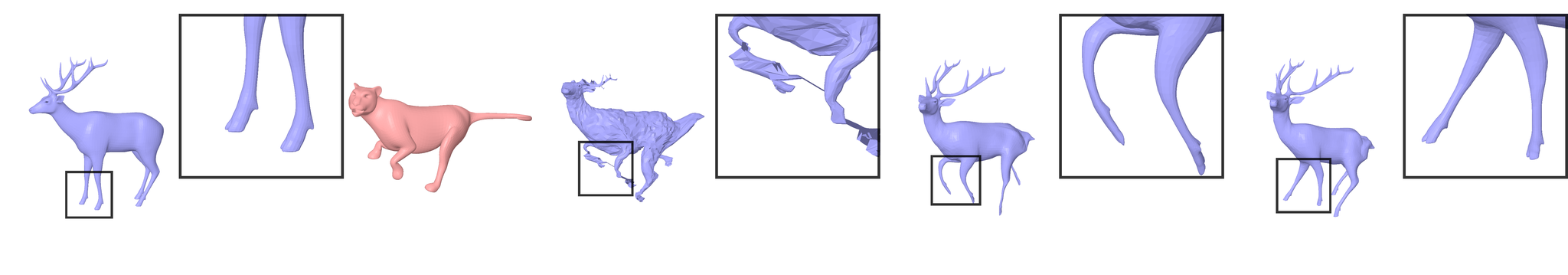}} \\
    \end{tabular}
    }
    \vspace{-1.0\baselineskip}
\caption{Qualitative results from the ablation study where a pose of the source shape $\mathcal{M}^S$ (\src{red}) in the \deformthings~\cite{Li:2021Deform4D} is transferred to the target template shape $\overline{\mathcal{M}}^T$ (\tgt{blue}).}
\label{fig:deform4d_ablation}
\vspace{-\baselineskip}
\end{figure}

\vspace{-0.5\baselineskip}
\subsection{Sensitivity to Number of Keypoints}
\label{subsec:keypoint_analysis}
\vspace{-0.5\baselineskip}
We examine the sensitivity of our method to the number of keypoints by testing different variants of our framework while varying the number of keypoints extracted by the pose extractor to 50, 25, and 10, respectively.
These variants are trained using the same SMPL~\cite{Loper:2015SMPL} human body shapes and animal shapes from the \deformthings~\cite{Li:2021Deform4D} dataset.
Our per-identity refinement stage (\shortsec~\ref{subsec:refinement}) is omitted to focus exclusively on the impact of keypoint counts on performance.
\tab~\ref{tab:keypoint_analysis} summarizes FID and PMD measured using the \deformthings and SMPL dataset, respectively.
We showcase qualitative results in \fig~\ref{fig:deform4d_keypoint_analysis} and \fig~\ref{fig:smpl_keypoint_analysis}.
As reflected in both quantitative and qualitative results, reducing the number of keypoints does not significantly affect pose transfer accuracy.

\begin{table}[h!]
    \footnotesize
    \setlength{\tabcolsep}{4pt}
    \centering
    \begin{tabularx}{\textwidth}{ccccc|ccccc}
         \toprule
         \multicolumn{5}{c|}{\deformthings~\cite{Li:2021Deform4D}} & \multicolumn{5}{c}{SMPL~\cite{Loper:2015SMPL}} \\
         Method & Ours--10 & Ours-25 & Ours-50 & \textbf{Ours-100} & Method & Ours--10 & Ours-25 & Ours-50 & \textbf{Ours-100}\\
         \midrule
         \makecell{FID \\ ($\times 10^{-2}$)} & 1.25 & 0.87 & 0.83 & \textbf{0.72} & \makecell{PMD \\ ($\times 10^{-3}$)} & 0.20 & 0.17 & 0.17 & \textbf{0.13} \\
         \bottomrule \\
    \end{tabularx}
    \caption{Quantitative results from the variants of our framework trained to extract different number of keypoints. Ours-$N$ denotes a variant of our network trained to extract $N$ keypoints.}
    \label{tab:keypoint_analysis}
\end{table}

\begin{figure}[!h]
    \vspace{-\baselineskip}
    \centering
    \setlength{\tabcolsep}{0em}
    \def\arraystretch{0.0}
    {\footnotesize
    \begin{tabular}{P{0.2\textwidth}P{0.1\textwidth}P{0.175\textwidth}P{0.175\textwidth}P{0.175\textwidth}P{0.175\textwidth}}
        $\overline{\mathcal{M}}^T$ & $\mathcal{M}^S$ & Ours-10 & Ours-25 & Ours-50 & Ours-100 \\
        \multicolumn{6}{c}{\includegraphics[width=\textwidth]{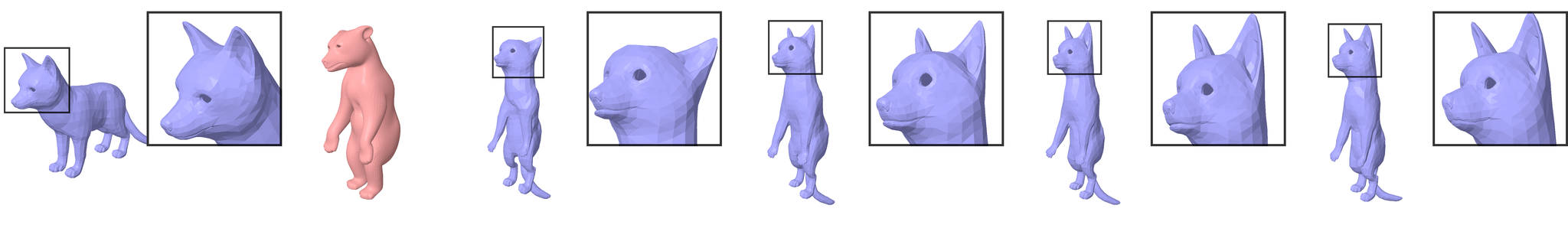}} \\
    \end{tabular}
    }
    \vspace{-\baselineskip}
\caption{Qualitative results of transferring a pose of the source shape $\mathcal{M}^S$ (\src{red}) in the \deformthings~\cite{Li:2021Deform4D} to the target template shape $\overline{\mathcal{M}}^T$  (\tgt{blue}) using variants of our framework (Ours-$N$), trained to extract $N$ keypoints.}
\label{fig:deform4d_keypoint_analysis}
\end{figure}

\begin{figure}[!h]
    \centering
    \setlength{\tabcolsep}{0em}
    \def\arraystretch{0.0}
    {\footnotesize
    \begin{tabular}{P{0.075\textwidth}P{0.125\textwidth}P{0.160\textwidth}P{0.160\textwidth}P{0.160\textwidth}P{0.160\textwidth}P{0.160\textwidth}}
        $\overline{\mathcal{M}}^T$ & $\mathcal{M}^S$ & Ours-10 & Ours-25 & Ours-50 & Ours-100 & $\mathcal{M}^T_{\text{GT}}$ \\
        \multicolumn{7}{c}{\includegraphics[width=\textwidth]{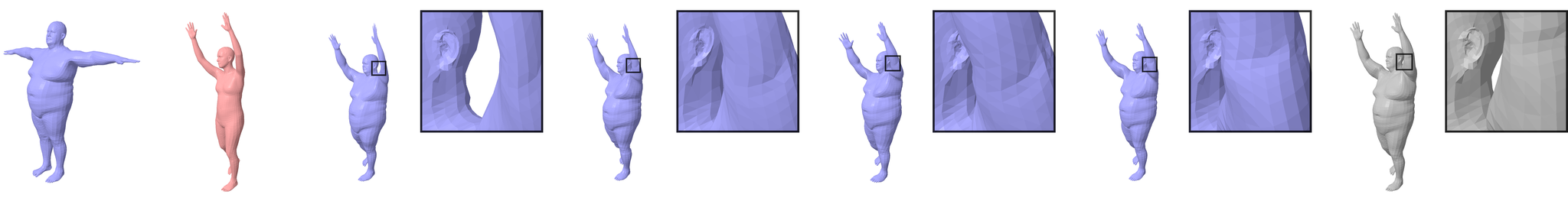}} \\
    \end{tabular}
    }
    \vspace{-\baselineskip}
\caption{Qualitative results of transferring a pose of the default human mesh $\mathcal{M}^S$ (\src{red}) to the target template mesh $\overline{\mathcal{M}}^T$ (\tgt{blue}) using variants of our framework (Ours-$N$), trained to extract $N$ keypoints.}
\label{fig:smpl_keypoint_analysis}
\end{figure}

\vspace{-0.5\baselineskip}
\subsection{Pose Variation Generation Using Diffusion Models}
\label{subsec:uncond_gen}
\begin{figure*}[!h]
{
    \centering
    \setlength{\tabcolsep}{0em}
    \def\arraystretch{0.0}

    \begin{minipage}{0.49\textwidth}
        \centering
        {\footnotesize
        \begin{tabularx}{\linewidth}{YYYY}
            \makecell{$\mathcal{M}^S$ \\ (Vertex)} & \makecell{$\mathcal{M}^S$ \\ (Ours)} & \makecell{$\mathcal{M}^T$ \\ (Vertex)} & \makecell{$\mathcal{M}^T$ \\ (Ours)} \\
            \multicolumn{4}{c}{\includegraphics[width=\textwidth]{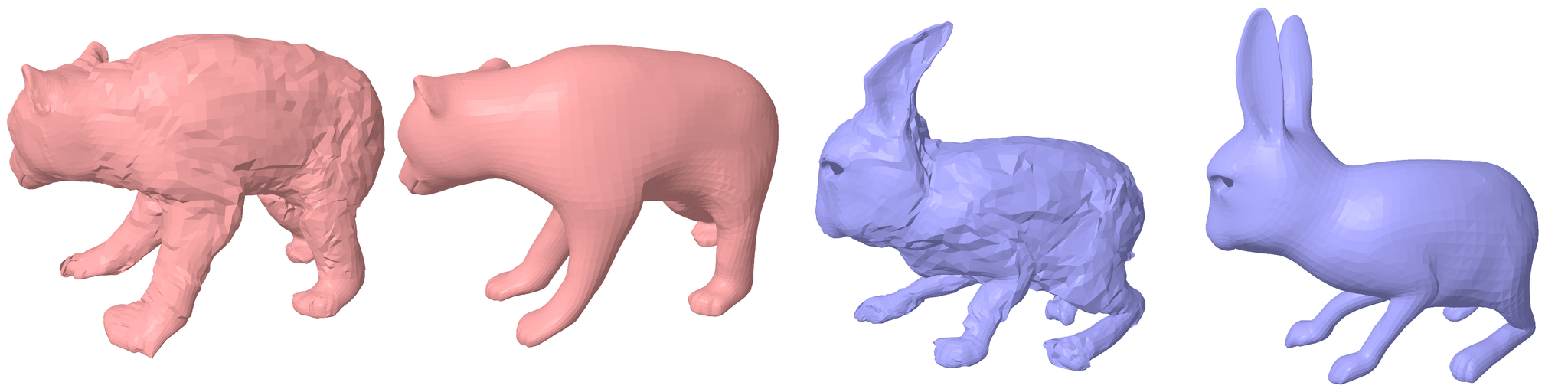}} \\
            \multicolumn{4}{c}{\includegraphics[width=\textwidth]{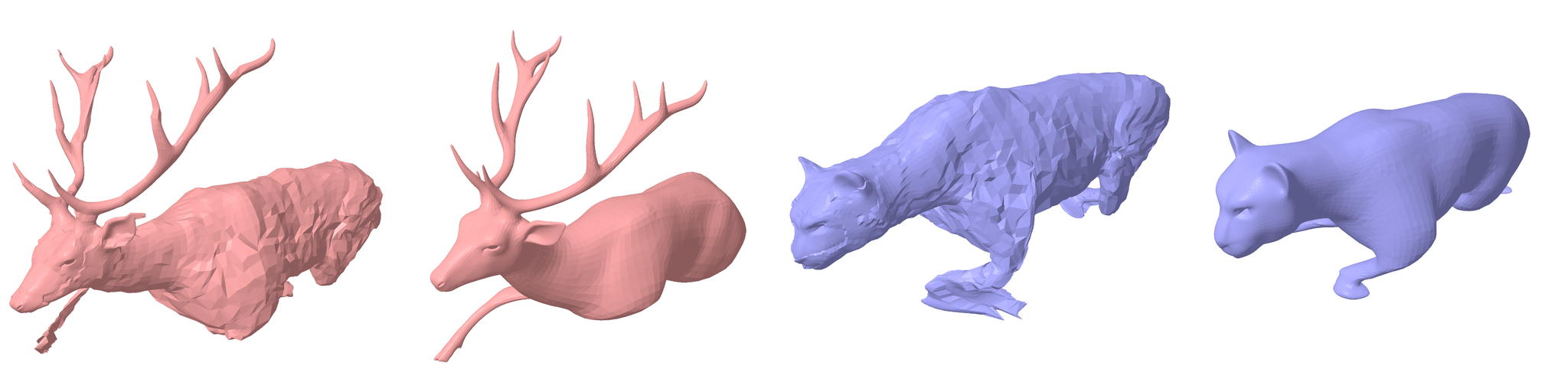}} \\
            \multicolumn{4}{c}{\includegraphics[width=\textwidth]{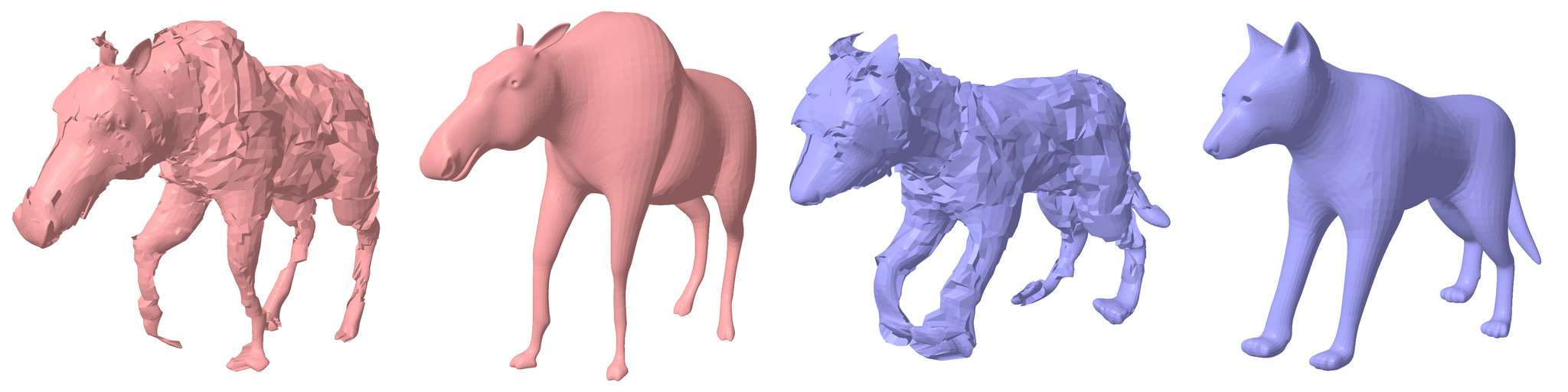}} \\
        \end{tabularx}
        }
    \end{minipage}\hfill\vline\hfill%
    \begin{minipage}{0.49\textwidth}
        \centering
        {\footnotesize
        \begin{tabularx}{\linewidth}{YYYY}
            \makecell{$\mathcal{M}^S$ \\ (Vertex)} & \makecell{$\mathcal{M}^S$ \\ (Ours)} & \makecell{$\mathcal{M}^T$ \\ (Vertex)} & \makecell{$\mathcal{M}^T$ \\ (Ours)} \\
            \multicolumn{4}{c}{\includegraphics[width=\textwidth]{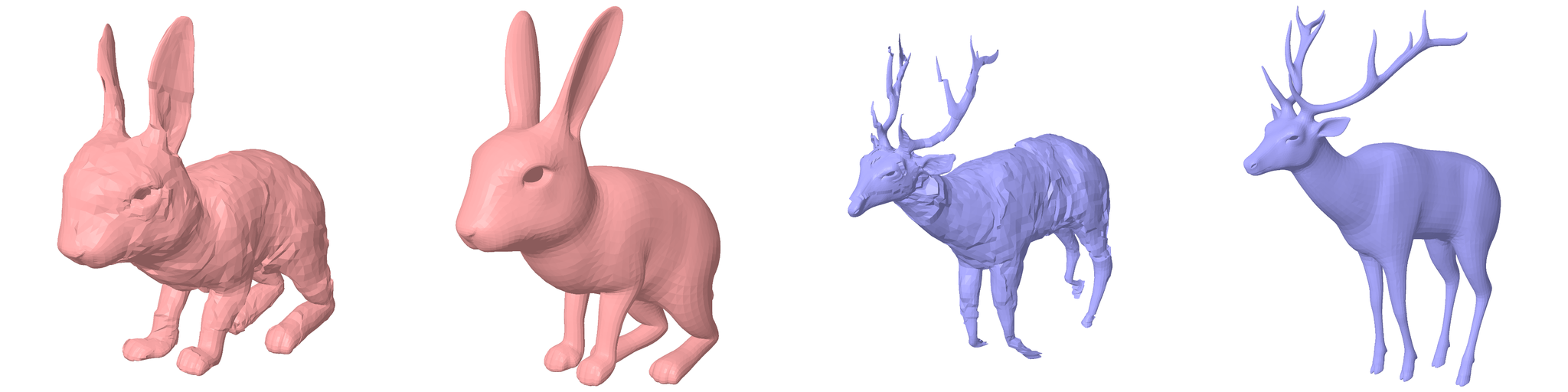}} \\
            \multicolumn{4}{c}{\includegraphics[width=\textwidth]{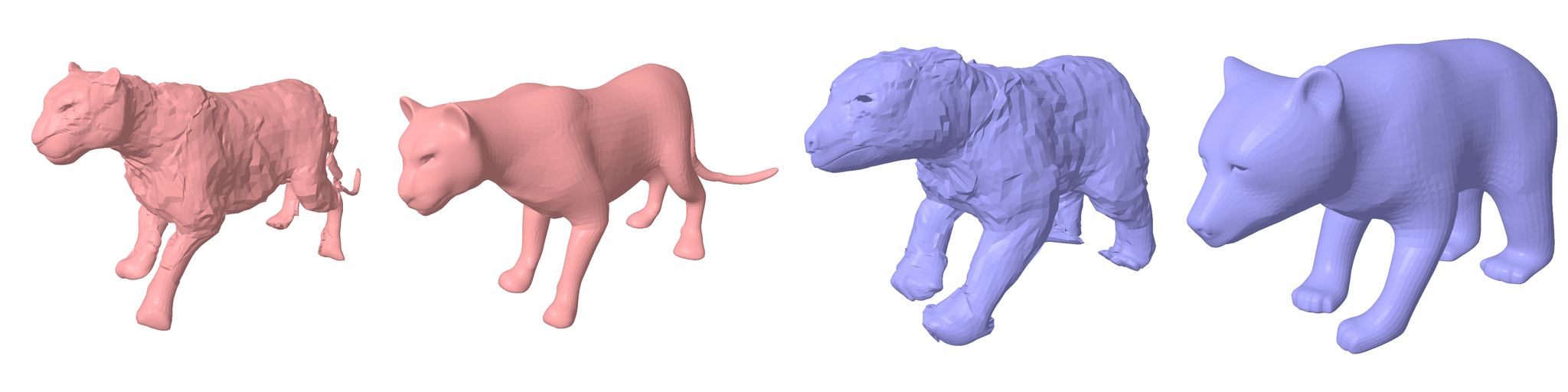}} \\
            \multicolumn{4}{c}{\includegraphics[width=\textwidth]{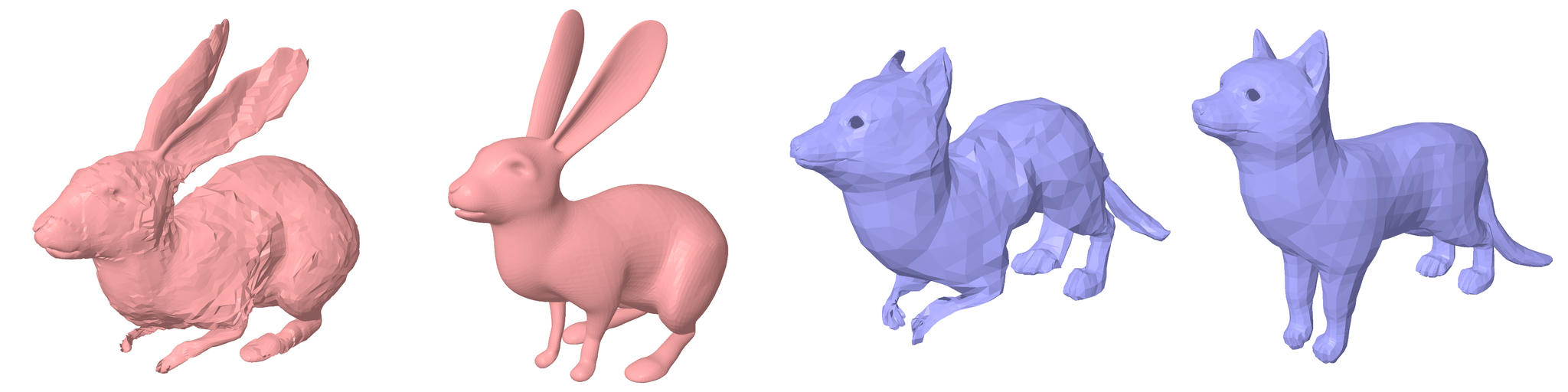}} \\
        \end{tabularx}
        }
    \end{minipage}
}
\vspace{-0.5\baselineskip}
\caption{Pose variation generation results. Each row illustrates the outcomes of applying a generated pose to a source template mesh $\overline{\mathcal{M}}^S$ (\src{red}) and a target template mesh $\overline{\mathcal{M}}^T$ (\tgt{blue}).}
\label{fig:uncond_gen}
\end{figure*}

\vspace{-0.5\baselineskip}
We evaluate the generation capabilities of our diffusion models trained using different pose representations. Since no existing generative model can learn pose representations transferable across various shapes, we focus on analyzing the impact of using Jacobian fields on generation quality. We use shapes obtained by applying 300 generated poses to both $\overline{\mathcal{M}}^S$'s (\src{red}) and various $\overline{\mathcal{M}}^T$'s (\tgt{blue}). The quantitative and qualitative results are summarized in \tab~\ref{tbl:ablation_diffusion} (right) and \fig~\ref{fig:uncond_gen}, respectively. While the poses are generated using the diffusion model, our model still achieves ResNet classification accuracy comparable to the pose transfer experiment (\tab~\ref{tbl:ablation_diffusion}, left).
This tendency is also reflected in the qualitative results shown in \fig~\ref{fig:uncond_gen}.
These results validate that the latent space learned from variations of Jacobian fields is more suitable for generating high-quality shape and pose variations compared to the one based on vertices. More results can be found in the appendix.

\vspace{-0.5\baselineskip}

\vspace{-1.5\baselineskip}
\section{Conclusion}
\label{sec:conclusion}
\vspace{-0.5\baselineskip}
We have presented a method for learning a novel neural representation of the pose of non-rigid 3D shapes, which facilitates: 1) the disentanglement of pose and object identity, 2) the training of a generative model due to its compactness, and 3) the transfer of poses to other objects' meshes. In our experiments, we demonstrated the state-of-the-art performance of our method in pose transfer, as well as its ability to generate diverse shapes by applying the generated poses to different identities.

\vspace{-0.5\baselineskip}
\paragraph{Limitations.}
Our method leverages differential operators to compute the Jacobian field of the given template mesh, requiring additional preprocessing when dealing with meshes that have multiple disconnected components or defects in the triangulation.
Our framework also assumes that a template mesh of the shape is known for pose transfer.
We plan to extend our framework for transferring poses between arbitrary shapes in future work.

\paragraph{Societal Impacts.} Our generative model for poses and the pose transfer technique could potentially be misused for deepfakes. Developing robust guidelines and techniques to prevent such misuse is an important area for future research.

\clearpage
\newpage

\section*{Acknowledgments}
\vspace{-0.5\baselineskip}
We appreciate Minh Hieu Nguyen and Jisung Hwang for their support on creating figures.
S. Yoo acknowledges the support of the Graduate School National Presidential Science Scholarship provided by Korea Student Aid Foundation.
This work was supported by the NRF grant (RS-2023-00209723), IITP grants (RS-2022-II220594, RS-2023-00227592, RS-2024-00399817), and KEIT grant (RS-2024-00423625), all funded by the Korean government (MSIT and MOTIE), as well as grants from the DRB-KAIST SketchTheFuture Research Center, NAVER-Intel Co-Lab, Hyundai NGV, KT, and Samsung Electronics.

\bibliographystyle{plainnat}
\bibliography{ref}

\newif\ifpaper
\papertrue

\renewcommand{\thesection}{A}
\renewcommand{\thetable}{A\arabic{table}}
\renewcommand{\thefigure}{A\arabic{figure}}

\clearpage
\newpage

\section*{Appendix}
\renewcommand{\thesubsection}{\Alph{subsection}}

In the following appendix, we provide implementation details of our method, including dataset processing, network architectures, training details, and hyperparameter selection in \shortsec~\ref{subsec:appendix_impl}.
We also present multi-view renderings of a pose transfer example showcased in \shortsec~\ref{subsec:human_transfer} in \shortsec~\ref{subsec:appendix_qualitative_multiview}, the full list of evaluation metrics reported in \tab~\ref{tbl:ablation_diffusion} in \shortsec~\ref{subsec:appendix_abl}, and additional qualitative results from our pose transfer and generation experiments in \shortsec~\ref{subsec:appendix_additional_qual}.

\vspace{-0.5\baselineskip}
\subsection{Implementation Details}
\label{subsec:appendix_impl}

\vspace{-0.5\baselineskip}
\paragraph{Data.} We use nine animal shapes from the \deformthings~\cite{Li:2021Deform4D} dataset in our experiments, specifically: \textsc{Bear}, \textsc{Bunny}, \textsc{Canine}, \textsc{Deer}, \textsc{Dog}, \textsc{Elk}, \textsc{Fox}, \textsc{Moose}, and \textsc{Puma}. We employ the first frame of the first animation sequence (alphabetically ordered) as the template for each animal, and the last frame of randomly sampled animation sequences for variations. Additionally, we use T-posed humanoid shapes from both the SMPL~\cite{Loper:2015SMPL} and Mixamo~\cite{Mixamo:2020} datasets as template meshes.

\vspace{-0.5\baselineskip}
\paragraph{Networks.} We utilize Point Transformer layers from Zhao et al.\cite{Zhao:2021PointTransformer} and Tang et al.\cite{Tang:2022NSDP} for implementing the pose extractor $g$ and the pose applier $h$. The network architectures for our cascaded diffusion models, as detailed in \shortsec~\ref{subsec:uncond_gen}, are adapted from Koo et al.~\cite{Koo:2023Salad}. These models operate over $T=1000$ timesteps with a linear noise schedule ranging from $\beta_1 = 1 \times 10^{-4}$ to $\beta_T = 5 \times 10^{-2}$. For model training, we employ the ADAM optimizer at a learning rate of $1\times 10^{-3}$ and standard parameters. Our experiments are conducted on RTX 3090 GPUs (24 GB VRAM) and A6000 GPUs (48 GB VRAM).

For per-identity refinement modules, we set $\lambda_{\text{lap}} = 1.0$, $\lambda_{\text{edge}} = 1.0$, and $\lambda_{\text{reg}} = 5 \times 10^{-2}$ during training.

\vspace{-0.5\baselineskip}
\paragraph{Loss Functions.}
We provide the definitions of the loss functions used to supervise the training of our per-identity refinement module.

The Laplacian loss~\cite{Liu:2021DeepMetaHandles} is defined between two sets of vertices with the vertex-wise correspondences:
\begin{align}
    \mathcal{L}_{\text{lap}} \left(\mathbf{V}, \overline{\mathbf{V}}^T \right) &= \mathbf{L}_T \left(\mathbf{V} - \overline{\mathbf{V}}^T \right),
\end{align}
where $\mathbf{V}$ is a set of new vertex coordinates, $\overline{\mathbf{V}}^T$ is a set of vertex coordinates of the target template mesh $\overline{\mathcal{M}}^T$, and $\mathbf{L}_T$ is the cotangent Laplacian of $\overline{\mathcal{M}}^T$.

Likewise, the edge length preservation loss~\cite{Liao:2022SPT} is defined as:
\begin{align}
    \mathcal{L}_{\text{edge}} &= \sum_{\{i, j\} \in \mathcal{E}} \vert \Vert \mathbf{V}_i - \mathbf{V}_j \Vert_2 - \Vert \overline{\mathbf{V}}^T_i - \overline{\mathbf{V}}^T_j \Vert_2 \vert,
\end{align}
where $\mathcal{E}$ is a set of edges comprising the target shapes and $\mathbf{V}_i$ and $\overline{\mathbf{V}}_i^T$ are the $i$-th vertex of a new set of vertex coordinates and that of $\overline{\mathcal{M}}^T$.

\vspace{-0.5\baselineskip}
\subsection{Multi-View Renderings of Qualitative Results}
\label{subsec:appendix_qualitative_multiview}
\begin{figure}[!h]
    \vspace{-0.5\baselineskip}
    \centering
    \setlength{\tabcolsep}{0em}
    \def\arraystretch{0.0}
    {\footnotesize
    \begin{tabular}{P{0.25\textwidth}P{0.25\textwidth}P{0.25\textwidth}P{0.25\textwidth}}
        Ours (View 1) & Ours (View 2) & Ours (View 3) & Ours (View 4) \\
        \multicolumn{4}{c}{\includegraphics[width=\textwidth]{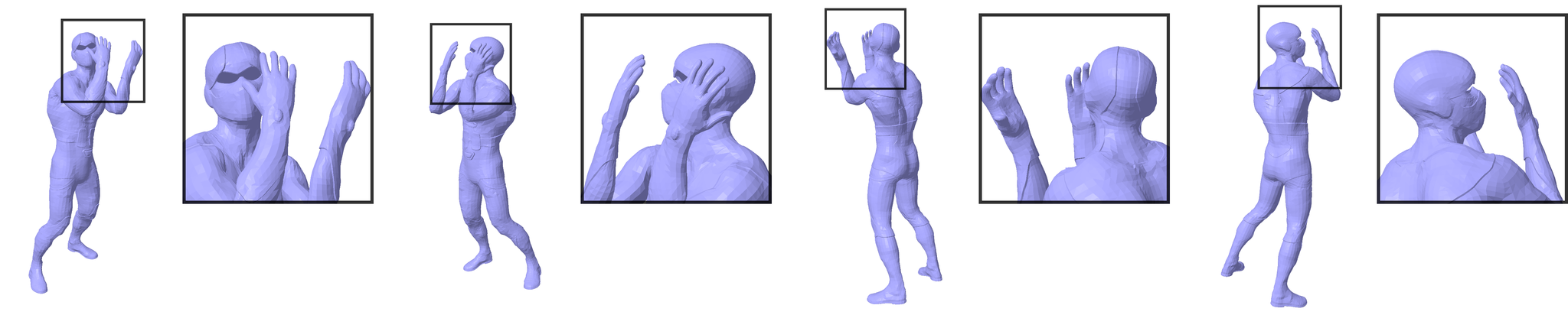}}
    \end{tabular}
    }
    \vspace{-0.5\baselineskip}
\caption{A pose transfer example showcased in \fig~\ref{fig:mixamo_transfer}, rendered from 4 different viewpoints.}
\label{fig:mixamo_consistency}
\end{figure}

While \fig~\ref{fig:mixamo_transfer} showcases only single-view images, our method produces high-fidelity 3D shapes after pose transfer as shown in \fig~\ref{fig:mixamo_consistency}.

\clearpage
\newpage

\subsection{Full List of Quantitative Metrics Reported in \tab~\ref{tbl:ablation_diffusion}}
\label{subsec:appendix_abl}

We report the full lists of evaluation metrics reported in \tab~\ref{tbl:ablation_diffusion} from the following page.

\begin{table}[h!]
    \centering
    \caption{\footnotesize{Full list of FIDs reported in \tab~\ref{tbl:ablation_diffusion} (left)}}
    {
    \scriptsize
    \setlength{\tabcolsep}{0.0em}
    \begin{tabularx}{\textwidth}{Y | Y Y Y Y Y Y Y Y Y}
        \toprule
         & Bear & Bunny & Canie & Deer & Dog & Elk & Fox & Moose & Puma \\
        \midrule
      \multicolumn{10}{c}{Vertices (\shortsec~\ref{subsec:identity_agnostic})} \\
        \midrule
        Bear &  -  &  3.47  &  11.53  &  6.76  &  6.14  &  8.40  &  4.92  &  5.26  &  9.80 \\
    Bunny &  4.43  &  -  &  7.71  &  4.03  &  6.07  &  5.94  &  4.62  &  4.88  &  8.11 \\
    Canie &  2.47  &  3.46  &  -  &  2.95  &  2.83  &  5.43  &  2.62  &  2.39  &  4.24 \\
    Deer &  2.39  &  2.37  &  4.80  &  -  &  3.32  &  3.00  &  3.12  &  1.45  &  6.53 \\
    Dog &  1.82  &  2.74  &  5.49  &  2.09  &  -  &  3.99  &  2.80  &  1.61  &  3.01 \\
    Elk &  2.35  &  1.74  &  9.04  &  3.13  &  2.79  &  -  &  2.58  &  2.83  &  5.02 \\
    Fox &  0.54  &  0.97  &  1.28  &  0.37  &  0.67  &  0.69  &  -  &  0.71  &  1.40 \\
    Moose &  3.38  &  2.12  &  8.31  &  4.25  &  3.76  &  6.13  &  3.63  &  -  &  6.74 \\
    Puma &  1.92  &  3.54  &  6.33  &  2.90  &  2.72  &  5.58  &  2.78  &  2.10  &  - \\
        \midrule
        \multicolumn{10}{c}{Jacobians (\shortsec~\ref{subsec:jacobian_fields})} \\
        \midrule
        Bear &  -  &  6.77  &  1.16  &  0.61  &  0.46  &  0.27  &  0.73  &  1.20  &  0.42 \\
    Bunny &  3.36  &  -  &  6.08  &  2.16  &  4.94  &  3.87  &  2.12  &  1.41  &  4.37 \\
    Canie &  0.45  &  1.40  &  -  &  0.41  &  0.78  &  0.92  &  0.62  &  0.35  &  0.64 \\
    Deer &  0.49  &  1.94  &  2.44  &  -  &  1.25  &  1.68  &  0.34  &  0.27  &  0.98 \\
    Dog &  0.21  &  3.70  &  0.48  &  0.42  &  -  &  0.17  &  0.47  &  0.62  &  0.16 \\
    Elk &  0.44  &  5.23  &  0.99  &  0.63  &  0.46  &  -  &  0.49  &  1.03  &  0.51 \\
    Fox &  0.99  &  0.98  &  1.74  &  0.74  &  1.47  &  1.37  &  -  &  0.53  &  1.27 \\
    Moose &  0.39  &  2.49  &  1.69  &  0.25  &  0.82  &  1.16  &  0.48  &  -  &  0.69 \\
    Puma &  0.39  &  2.80  &  0.46  &  0.46  &  0.36  &  0.35  &  0.24  &  0.47  &  - \\
        \midrule
        \multicolumn{10}{c}{Ours (Jacobians + Refinement (\shortsec~\ref{subsec:refinement}))} \\
        \midrule
        Bear &  -  &  3.83  &  1.10  &  0.60  &  0.56  &  0.30  &  0.58  &  0.92  &  0.41 \\
    Bunny &  3.97  &  -  &  6.22  &  2.71  &  4.75  &  3.39  &  2.89  &  2.16  &  4.72 \\
    Canie &  0.64  &  0.98  &  -  &  0.51  &  0.80  &  0.96  &  0.66  &  0.40  &  0.83 \\
    Deer &  0.68  &  1.35  &  2.18  &  -  &  1.29  &  1.10  &  0.40  &  0.28  &  1.11 \\
    Dog &  0.33  &  2.03  &  0.46  &  0.45  &  -  &  0.22  &  0.25  &  0.52  &  0.20 \\
    Elk &  0.42  &  3.70  &  0.89  &  0.61  &  0.49  &  -  &  0.46  &  0.85  &  0.53 \\
    Fox &  1.61  &  0.60  &  1.73  &  1.34  &  1.46  &  1.04  &  -  &  1.08  &  1.51 \\
    Moose &  0.58  &  1.58  &  1.74  &  0.34  &  0.93  &  0.95  &  0.48  &  -  &  0.79 \\
    Puma &  0.50  &  1.87  &  0.55  &  0.53  &  0.46  &  0.40  &  0.29  &  0.47  &  - \\
        \bottomrule
    \end{tabularx}
    }
\end{table}

\begin{table}[h!]
    \centering
    \caption{\footnotesize{Full list of KIDs reported in \tab~\ref{tbl:ablation_diffusion} (left)}}
    {
    \scriptsize
    \setlength{\tabcolsep}{0.0em}
    \begin{tabularx}{\textwidth}{Y | Y Y Y Y Y Y Y Y Y}
        \toprule
         & Bear & Bunny & Canie & Deer & Dog & Elk & Fox & Moose & Puma \\
        \midrule
      \multicolumn{10}{c}{Vertices (\shortsec~\ref{subsec:identity_agnostic})} \\
        \midrule
        Bear &  -  &  2.56  &  8.87  &  4.95  &  4.46  &  6.15  &  3.65  &  3.61  &  6.95 \\
    Bunny &  2.06  &  -  &  3.87  &  2.69  &  3.20  &  2.76  &  2.42  &  2.50  &  3.79 \\
    Canie &  1.55  &  2.38  &  -  &  1.97  &  1.82  &  3.79  &  1.68  &  1.46  &  2.82 \\
    Deer &  1.05  &  1.01  &  2.07  &  -  &  1.53  &  1.24  &  1.68  &  0.62  &  3.93 \\
    Dog &  1.30  &  1.70  &  3.99  &  1.49  &  -  &  2.92  &  2.08  &  1.11  &  2.00 \\
    Elk &  1.37  &  0.65  &  6.57  &  2.22  &  1.60  &  -  &  1.07  &  1.91  &  3.02 \\
    Fox &  0.26  &  0.25  &  0.39  &  0.16  &  0.25  &  0.27  &  -  &  0.39  &  0.64 \\
    Moose &  1.63  &  0.83  &  5.29  &  2.91  &  1.67  &  3.33  &  1.39  &  -  &  2.93 \\
    Puma &  1.03  &  2.10  &  4.38  &  1.65  &  1.72  &  3.85  &  1.86  &  1.14  &  - \\
        \midrule
        \multicolumn{10}{c}{Jacobians (\shortsec~\ref{subsec:jacobian_fields})} \\
        \midrule
        Bear &  -  &  2.36  &  0.29  &  0.21  &  0.20  &  0.01  &  0.17  &  0.27  &  0.03 \\
    Bunny &  1.18  &  -  &  3.00  &  1.02  &  2.26  &  1.70  &  0.85  &  0.43  &  1.97 \\
    Canie &  0.17  &  0.67  &  -  &  0.10  &  0.25  &  0.25  &  0.14  &  0.11  &  0.19 \\
    Deer &  0.10  &  0.11  &  0.99  &  -  &  0.70  &  0.93  &  0.08  &  0.03  &  0.41 \\
    Dog &  0.08  &  1.71  &  0.11  &  0.15  &  -  &  0.05  &  0.27  &  0.18  &  -0.00 \\
    Elk &  0.02  &  1.97  &  0.09  &  0.15  &  0.05  &  -  &  0.06  &  0.17  &  0.05 \\
    Fox &  0.62  &  0.27  &  0.99  &  0.74  &  1.04  &  0.84  &  -  &  0.32  &  0.79 \\
    Moose &  0.22  &  1.18  &  0.72  &  0.13  &  0.45  &  0.46  &  0.09  &  -  &  0.20 \\
    Puma &  0.15  &  0.77  &  0.08  &  0.29  &  0.15  &  0.15  &  0.03  &  0.22  &  - \\
        \midrule
        \multicolumn{10}{c}{Ours (Jacobians + Refinement (\shortsec~\ref{subsec:refinement}))} \\
        \midrule
        Bear &  -  &  1.26  &  0.29  &  0.23  &  0.20  &  0.01  &  0.08  &  0.27  &  0.07 \\
    Bunny &  1.42  &  -  &  2.92  &  1.02  &  2.10  &  1.34  &  0.85  &  0.80  &  1.97 \\
    Canie &  0.17  &  0.44  &  -  &  0.14  &  0.25  &  0.25  &  0.09  &  0.11  &  0.27 \\
    Deer &  0.19  &  0.11  &  0.94  &  -  &  0.66  &  0.46  &  0.08  &  0.03  &  0.41 \\
    Dog &  0.08  &  0.93  &  0.11  &  0.17  &  -  &  0.05  &  0.08  &  0.18  &  0.03 \\
    Elk &  0.02  &  1.34  &  0.09  &  0.17  &  0.05  &  -  &  0.04  &  0.17  &  0.09 \\
    Fox &  0.92  &  0.27  &  0.90  &  0.74  &  0.90  &  0.44  &  -  &  0.65  &  0.79 \\
    Moose &  0.22  &  0.68  &  0.72  &  0.19  &  0.45  &  0.46  &  0.04  &  -  &  0.29 \\
    Puma &  0.22  &  0.77  &  0.16  &  0.29  &  0.21  &  0.17  &  0.03  &  0.25  &  - \\
        \bottomrule
    \end{tabularx}
    }
\end{table}

\begin{table}[h!]
    \centering
    \caption{\footnotesize{Full list of ResNet classification accuracies reported in \tab~\ref{tbl:ablation_diffusion} (left)}}
    {
    \scriptsize
    \setlength{\tabcolsep}{0.0em}
    \begin{tabularx}{\textwidth}{Y | Y Y Y Y Y Y Y Y Y}
        \toprule
         & Bear & Bunny & Canie & Deer & Dog & Elk & Fox & Moose & Puma \\
        \midrule
      \multicolumn{10}{c}{Vertices (\shortsec~\ref{subsec:identity_agnostic})} \\
        \midrule
        Bear &  -  &  81.33  &  37.58  &  58.00  &  72.75  &  53.17  &  65.83  &  54.00  &  41.33 \\
    Bunny &  57.25  &  -  &  26.83  &  51.25  &  41.33  &  35.58  &  52.83  &  49.75  &  15.33 \\
    Canie &  77.67  &  69.92  &  -  &  74.25  &  78.75  &  63.75  &  77.75  &  76.17  &  57.58 \\
    Deer &  95.08  &  90.42  &  70.33  &  -  &  93.50  &  85.33  &  97.58  &  98.08  &  67.42 \\
    Dog &  67.50  &  45.42  &  17.58  &  51.08  &  -  &  38.00  &  69.17  &  64.00  &  18.58 \\
    Elk &  92.08  &  81.58  &  72.25  &  86.25  &  92.50  &  -  &  84.92  &  90.08  &  81.17 \\
    Fox &  78.92  &  58.08  &  34.25  &  81.58  &  68.75  &  46.42  &  -  &  84.42  &  17.17 \\
    Moose &  44.33  &  22.33  &  9.83  &  74.33  &  40.17  &  33.67  &  25.08  &  -  &  6.42 \\
    Puma &  82.25  &  73.92  &  71.00  &  69.00  &  81.08  &  71.50  &  81.00  &  77.92  &  - \\
        \midrule
        \multicolumn{10}{c}{Jacobians (\shortsec~\ref{subsec:jacobian_fields})} \\
        \midrule
        Bear &  -  &  93.50  &  85.50  &  92.08  &  95.58  &  95.50  &  96.42  &  93.83  &  91.50 \\
    Bunny &  86.67  &  -  &  60.00  &  72.83  &  69.00  &  66.75  &  85.33  &  79.08  &  63.08 \\
    Canie &  77.83  &  80.25  &  -  &  83.92  &  81.33  &  76.17  &  92.83  &  81.17  &  80.08 \\
    Deer &  98.17  &  93.00  &  97.25  &  -  &  100.00  &  97.67  &  99.58  &  99.17  &  99.83 \\
    Dog &  85.50  &  58.58  &  60.58  &  82.92  &  -  &  75.92  &  85.50  &  76.75  &  65.33 \\
    Elk &  86.67  &  73.17  &  86.75  &  81.42  &  89.00  &  -  &  91.17  &  85.67  &  91.00 \\
    Fox &  86.83  &  62.33  &  64.17  &  82.67  &  74.08  &  72.25  &  -  &  84.58  &  76.50 \\
    Moose &  94.08  &  92.92  &  69.33  &  92.83  &  89.42  &  69.17  &  80.83  &  -  &  78.75 \\
    Puma &  90.75  &  78.08  &  87.58  &  87.17  &  89.75  &  88.83  &  88.25  &  88.25  &  - \\
        \midrule
        \multicolumn{10}{c}{Ours (Jacobians + Refinement (\shortsec~\ref{subsec:refinement}))} \\
        \midrule
        Bear &  -  &  91.67  &  92.50  &  93.00  &  96.58  &  95.75  &  97.42  &  94.42  &  94.67 \\
    Bunny &  88.58  &  -  &  72.33  &  74.83  &  80.33  &  74.42  &  89.33  &  81.75  &  73.42 \\
    Canie &  80.00  &  83.92  &  -  &  86.50  &  84.42  &  79.92  &  91.67  &  81.58  &  83.33 \\
    Deer &  98.58  &  94.00  &  99.08  &  -  &  99.92  &  99.50  &  99.75  &  99.25  &  99.67 \\
    Dog &  86.92  &  66.17  &  64.25  &  86.00  &  -  &  80.83  &  86.58  &  77.75  &  74.08 \\
    Elk &  88.25  &  75.42  &  90.17  &  82.92  &  91.17  &  -  &  92.08  &  86.83  &  92.00 \\
    Fox &  87.58  &  81.08  &  84.25  &  89.00  &  91.17  &  87.42  &  -  &  89.92  &  91.42 \\
    Moose &  95.75  &  94.00  &  75.50  &  93.33  &  91.92  &  77.25  &  92.75  &  -  &  86.08 \\
    Puma &  91.25  &  82.58  &  91.58  &  88.00  &  91.50  &  90.33  &  89.83  &  90.33  &  - \\
        \bottomrule
    \end{tabularx}
    }
\end{table}

\begin{table}[h!]
    \centering
    \caption{\footnotesize{Full list of FIDs reported in \tab~\ref{tbl:ablation_diffusion} (right)}}
    {
    \scriptsize
    \setlength{\tabcolsep}{0.0em}
    \begin{tabularx}{\textwidth}{Y | Y Y Y Y Y Y Y Y Y}
        \toprule
         & Bear & Bunny & Canie & Deer & Dog & Elk & Fox & Moose & Puma \\
        \midrule
         \multicolumn{10}{c}{Vertices (\shortsec~\ref{subsec:identity_agnostic})} \\
        \midrule
        Bear &  -  &  6.48  &  30.77  &  12.79  &  12.01  &  20.99  &  10.43  &  24.06  &  12.48 \\
        Bunny &  11.58  &  -  &  36.26  &  13.34  &  15.52  &  24.21  &  12.93  &  25.78  &  15.91 \\
        Canie &  4.03  &  2.58  &  -  &  5.06  &  4.95  &  8.12  &  4.54  &  9.67  &  4.71 \\
        Deer &  6.42  &  4.82  &  23.16  &  -  &  10.69  &  14.91  &  9.67  &  15.48  &  13.75 \\
        Dog &  2.62  &  1.69  &  12.12  &  4.06  &  -  &  7.53  &  3.47  &  9.51  &  3.57 \\
        Elk &  4.58  &  3.33  &  16.49  &  5.82  &  6.27  &  -  &  6.90  &  12.78  &  6.82 \\
        Fox &  4.31  &  2.18  &  20.87  &  7.74  &  7.95  &  13.40  &  -  &  16.59  &  6.30 \\
        Moose &  7.60  &  6.18  &  22.20  &  9.00  &  10.64  &  15.46  &  10.74  &  -  &  11.63 \\
        Puma &  3.46  &  2.89  &  12.97  &  5.18  &  5.22  &  8.75  &  4.57  &  10.18  &  - \\
        \midrule
        \multicolumn{10}{c}{Jacobians (\shortsec~\ref{subsec:jacobian_fields})} \\
        \midrule
        Bear &  -  &  2.33  &  9.66  &  2.12  &  4.62  &  14.47  &  1.82  &  0.80  &  2.74 \\
        Bunny &  7.29  &  -  &  19.93  &  7.19  &  12.65  &  23.89  &  6.74  &  3.02  &  8.76 \\
        Canie &  1.75  &  0.94  &  -  &  2.22  &  4.60  &  10.37  &  2.21  &  1.03  &  3.16 \\
        Deer &  1.75  &  0.99  &  10.24  &  -  &  5.68  &  17.01  &  2.35  &  0.53  &  3.04 \\
        Dog &  1.21  &  1.38  &  5.16  &  1.69  &  -  &  8.80  &  0.91  &  0.80  &  1.51 \\
        Elk &  0.80  &  2.39  &  5.89  &  2.11  &  3.61  &  -  &  1.60  &  0.73  &  1.78 \\
        Fox &  3.17  &  0.65  &  9.70  &  4.07  &  6.86  &  14.63  &  -  &  1.10  &  4.18 \\
        Moose &  1.97  &  1.38  &  10.27  &  3.59  &  6.09  &  16.94  &  3.70  &  -  &  3.36 \\
        Puma &  1.43  &  1.37  &  5.25  &  1.85  &  3.12  &  8.93  &  1.49  &  0.71  &  - \\
        \midrule
        \multicolumn{10}{c}{Ours (Jacobians + Refinement (\shortsec~\ref{subsec:refinement}))} \\
        \midrule
        Bear &  -  &  1.46  &  9.01  &  2.20  &  4.49  &  13.71  &  2.14  &  0.85  &  2.72 \\
        Bunny &  8.03  &  -  &  17.47  &  7.42  &  11.33  &  21.57  &  7.42  &  3.79  &  8.62 \\
        Canie &  2.05  &  1.04  &  -  &  2.23  &  4.37  &  9.95  &  2.33  &  1.20  &  3.21 \\
        Deer &  2.18  &  1.04  &  8.96  &  -  &  5.45  &  14.57  &  2.53  &  0.78  &  3.11 \\
        Dog &  1.58  &  1.03  &  4.81  &  1.83  &  -  &  8.42  &  1.18  &  0.98  &  1.62 \\
        Elk &  0.99  &  1.84  &  5.57  &  2.09  &  3.58  &  -  &  1.65  &  0.79  &  1.79 \\
        Fox &  3.91  &  1.61  &  7.59  &  3.85  &  5.53  &  12.00  &  -  &  2.14  &  3.79 \\
        Moose &  2.36  &  1.25  &  9.65  &  3.72  &  5.98  &  16.01  &  3.46  &  -  &  3.50 \\
        Puma &  1.66  &  1.23  &  5.09  &  1.94  &  3.16  &  8.59  &  1.60  &  0.84  &  - \\
        \bottomrule
    \end{tabularx}
    }
\end{table}

\begin{table}[h!]
    \centering
    \caption{\footnotesize{Full list of KIDs reported in \tab~\ref{tbl:ablation_diffusion} (right)}}
    {
    \scriptsize
    \setlength{\tabcolsep}{0.0em}
    \begin{tabularx}{\textwidth}{Y | Y Y Y Y Y Y Y Y Y}
        \toprule
         & Bear & Bunny & Canie & Deer & Dog & Elk & Fox & Moose & Puma \\
        \midrule
      \multicolumn{10}{c}{Vertices (\shortsec~\ref{subsec:identity_agnostic})} \\
        \midrule
        Bear &  -  &  3.53  &  15.17  &  7.02  &  5.90  &  10.50  &  5.51  &  10.86  &  6.91 \\
    Bunny &  5.15  &  -  &  17.82  &  6.40  &  7.48  &  11.71  &  6.10  &  12.39  &  7.15 \\
    Canie &  1.74  &  1.33  &  -  &  2.21  &  2.05  &  3.52  &  1.99  &  4.47  &  2.00 \\
    Deer &  3.41  &  2.80  &  14.00  &  -  &  6.58  &  8.71  &  5.83  &  8.33  &  8.77 \\
    Dog &  1.31  &  0.83  &  5.61  &  1.88  &  -  &  3.36  &  1.66  &  4.15  &  1.65 \\
    Elk &  2.13  &  1.46  &  7.88  &  2.70  &  2.85  &  -  &  3.04  &  5.55  &  3.25 \\
    Fox &  2.28  &  1.10  &  11.40  &  4.44  &  4.24  &  7.07  &  -  &  8.84  &  3.36 \\
    Moose &  3.49  &  2.77  &  10.45  &  4.22  &  5.14  &  6.91  &  4.92  &  -  &  4.88 \\
    Puma &  1.48  &  1.39  &  5.73  &  2.24  &  2.10  &  3.63  &  2.09  &  4.46  &  - \\
        \midrule
        \multicolumn{10}{c}{Jacobians (\shortsec~\ref{subsec:jacobian_fields})} \\
        \midrule
        Bear &  -  &  0.66  &  5.20  &  1.35  &  2.61  &  7.83  &  0.95  &  0.23  &  1.37 \\
    Bunny &  2.85  &  -  &  10.52  &  3.68  &  6.35  &  12.27  &  2.81  &  0.97  &  3.84 \\
    Canie &  0.65  &  0.39  &  -  &  1.29  &  2.25  &  5.33  &  0.77  &  0.24  &  1.35 \\
    Deer &  0.72  &  0.17  &  6.97  &  -  &  4.06  &  12.34  &  1.33  &  0.15  &  1.45 \\
    Dog &  0.52  &  0.60  &  2.90  &  1.13  &  -  &  4.84  &  0.40  &  0.19  &  0.67 \\
    Elk &  0.20  &  0.82  &  3.32  &  1.38  &  1.97  &  -  &  0.77  &  0.14  &  0.85 \\
    Fox &  1.96  &  0.27  &  6.08  &  2.89  &  4.52  &  8.89  &  -  &  0.64  &  2.48 \\
    Moose &  0.79  &  0.54  &  5.68  &  2.43  &  3.41  &  9.92  &  1.86  &  -  &  1.51 \\
    Puma &  0.78  &  0.53  &  2.99  &  1.29  &  1.77  &  5.13  &  0.74  &  0.38  &  - \\
        \midrule
        \multicolumn{10}{c}{Ours (Jacobians + Refinement (\shortsec~\ref{subsec:refinement}))} \\
        \midrule
        Bear &  -  &  0.35  &  4.87  &  1.39  &  2.57  &  7.38  &  1.16  &  0.28  &  1.38 \\
    Bunny &  3.12  &  -  &  9.03  &  3.84  &  5.52  &  10.76  &  3.11  &  1.30  &  3.71 \\
    Canie &  0.79  &  0.41  &  -  &  1.26  &  2.11  &  5.04  &  0.83  &  0.33  &  1.39 \\
    Deer &  0.93  &  0.35  &  5.93  &  -  &  3.74  &  10.15  &  1.40  &  0.28  &  1.45 \\
    Dog &  0.71  &  0.42  &  2.67  &  1.20  &  -  &  4.54  &  0.55  &  0.28  &  0.72 \\
    Elk &  0.29  &  0.59  &  3.12  &  1.39  &  1.97  &  -  &  0.77  &  0.18  &  0.86 \\
    Fox &  2.23  &  0.84  &  4.52  &  2.46  &  3.47  &  6.96  &  -  &  1.19  &  2.11 \\
    Moose &  0.97  &  0.44  &  5.31  &  2.52  &  3.35  &  9.27  &  1.68  &  -  &  1.61 \\
    Puma &  0.91  &  0.48  &  2.91  &  1.34  &  1.81  &  4.91  &  0.80  &  0.46  &  - \\
        \bottomrule
    \end{tabularx}
    }
\end{table}

\begin{table}[h!]
    \centering
    \caption{\footnotesize{Full list of ResNet classification accuracies reported in \tab~\ref{tbl:ablation_diffusion} (right)}}
    {
    \scriptsize
    \setlength{\tabcolsep}{0.0em}
    \begin{tabularx}{\textwidth}{Y | Y Y Y Y Y Y Y Y Y}
        \toprule
         & Bear & Bunny & Canie & Deer & Dog & Elk & Fox & Moose & Puma \\
        \midrule
      \multicolumn{10}{c}{Vertices (\shortsec~\ref{subsec:identity_agnostic})} \\
        \midrule
            Bear &  -  &  59.25  &  8.17  &  24.08  &  48.75  &  15.50  &  36.00  &  20.75  &  30.17 \\
    Bunny &  38.50  &  -  &  0.83  &  15.00  &  21.25  &  9.75  &  24.33  &  13.83  &  5.17 \\
    Canie &  72.58  &  78.42  &  -  &  56.67  &  71.92  &  39.50  &  61.50  &  35.67  &  66.58 \\
    Deer &  92.42  &  84.83  &  65.58  &  -  &  88.25  &  75.58  &  82.00  &  85.00  &  73.83 \\
    Dog &  66.75  &  45.00  &  6.17  &  28.08  &  -  &  13.92  &  51.17  &  24.25  &  14.75 \\
    dragonOLO &  99.67  &  97.92  &  82.25  &  100.00  &  99.58  &  -  &  99.67  &  94.83  &  96.33 \\
    Elk &  86.83  &  71.33  &  32.50  &  62.92  &  74.50  &  57.50  &  -  &  44.50  &  81.92 \\
    Fox &  64.42  &  62.83  &  10.58  &  47.17  &  54.50  &  22.42  &  66.67  &  -  &  13.25 \\
    Moose &  32.58  &  15.42  &  1.33  &  41.50  &  19.83  &  11.50  &  13.00  &  35.83  &  - \\
        \midrule
        \multicolumn{10}{c}{Jacobians (\shortsec~\ref{subsec:jacobian_fields})} \\
        \midrule
            Bear &  -  &  91.58  &  77.75  &  90.33  &  93.92  &  82.67  &  94.92  &  94.83  &  90.67 \\
    Bunny &  90.92  &  -  &  42.67  &  70.17  &  74.33  &  44.92  &  84.17  &  84.08  &  68.25 \\
    Canie &  85.17  &  87.42  &  -  &  77.83  &  84.67  &  68.08  &  96.00  &  85.75  &  89.58 \\
    Deer &  99.67  &  96.75  &  99.58  &  -  &  99.92  &  99.58  &  100.00  &  99.75  &  100.00 \\
    Dog &  90.92  &  69.75  &  69.00  &  83.08  &  -  &  78.92  &  93.17  &  87.83  &  81.33 \\
    Elk &  94.50  &  80.67  &  94.42  &  78.58  &  92.42  &  -  &  95.00  &  91.92  &  92.50 \\
    Fox &  89.17  &  76.33  &  59.92  &  79.25  &  78.67  &  39.08  &  -  &  92.17  &  89.58 \\
    Moose &  97.00  &  91.58  &  57.92  &  84.50  &  86.83  &  45.17  &  68.67  &  -  &  85.33 \\
    Puma &  93.67  &  84.42  &  93.42  &  84.75  &  92.42  &  95.25  &  89.83  &  90.75  &  - \\
        \midrule
        \multicolumn{10}{c}{Ours (Jacobians + Refinement (\shortsec~\ref{subsec:refinement}))} \\
        \midrule
            Bear &  -  &  90.83  &  84.75  &  90.58  &  96.08  &  86.75  &  96.42  &  94.75  &  95.92 \\
    Bunny &  91.08  &  -  &  56.00  &  72.33  &  81.67  &  53.17  &  88.17  &  86.00  &  77.42 \\
    Canie &  87.75  &  89.08  &  -  &  79.33  &  85.92  &  69.08  &  95.25  &  86.50  &  90.75 \\
    Deer &  99.83  &  98.00  &  99.92  &  -  &  100.00  &  100.00  &  100.00  &  99.67  &  99.83 \\
    Dog &  91.83  &  73.50  &  76.92  &  84.50  &  -  &  85.67  &  93.58  &  89.17  &  87.08 \\
    Elk &  95.00  &  81.92  &  95.00  &  81.75  &  92.50  &  -  &  95.00  &  93.17  &  93.75 \\
    Fox &  88.17  &  85.58  &  78.58  &  82.92  &  91.25  &  61.33  &  -  &  90.50  &  94.75 \\
    Moose &  97.08  &  92.00  &  67.08  &  84.92  &  91.08  &  55.83  &  86.92  &  -  &  89.75 \\
    Puma &  93.75  &  87.08  &  94.50  &  86.75  &  92.08  &  95.50  &  91.17  &  92.75  &  - \\
        \bottomrule
    \end{tabularx}
    }
\end{table}

\clearpage
\newpage
\subsection{Additional Qualitative Results}
\label{subsec:appendix_additional_qual}

We showcase additional qualitative results from the pose transfer and pose generation experiments below.

\begin{figure*}[!h]
{
    \centering
    \setlength{\tabcolsep}{0em}
    \def\arraystretch{0.0}

    \begin{minipage}{0.49\textwidth}
        \centering
        \begin{tabularx}{\linewidth}{YYYYY}
            \makecell{$\overline{\mathcal{M}}^T$} & \makecell{$\mathcal{M}^S$} & \makecell{NJF~\cite{Aigerman:2022NJF}} & \makecell{ZPT~\cite{Wang:2023ZPT}} & \makecell{Ours} \\
            \multicolumn{5}{c}{\includegraphics[width=\linewidth]{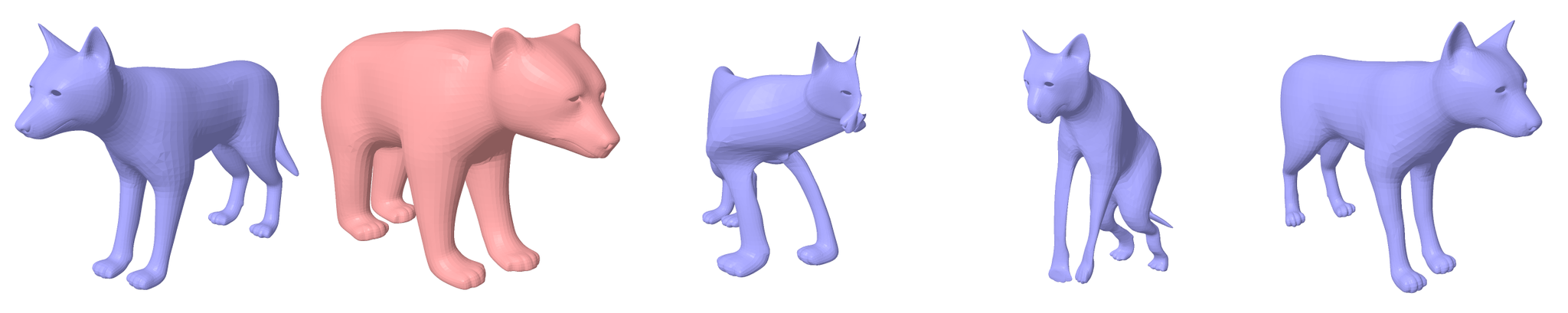}} \\
            \multicolumn{5}{c}{\includegraphics[width=\linewidth]{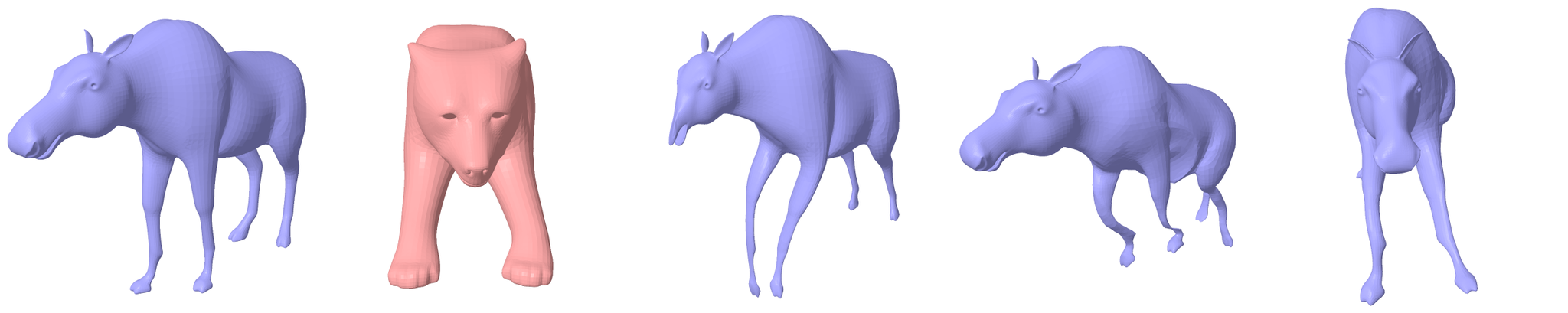}} \\
            \multicolumn{5}{c}{\includegraphics[width=\linewidth]{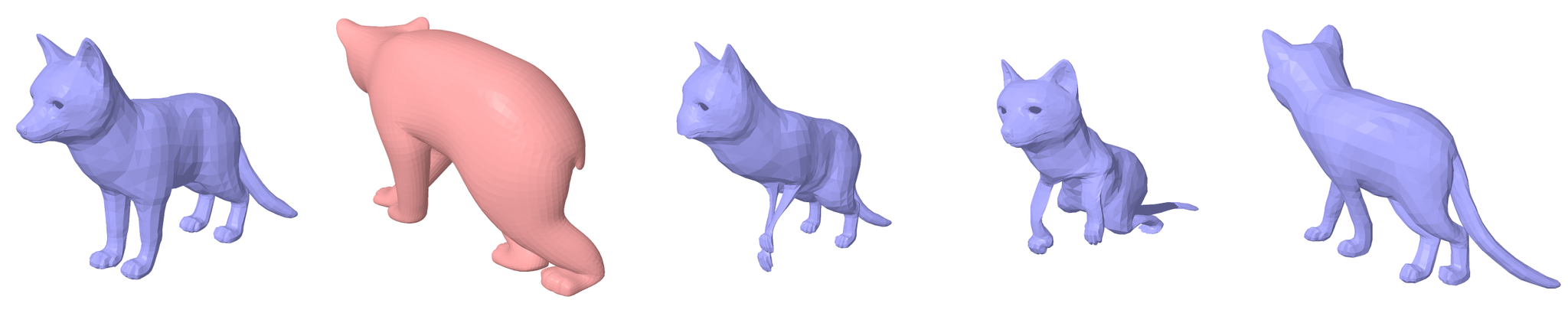}} \\
            \multicolumn{5}{c}{\includegraphics[width=\linewidth]{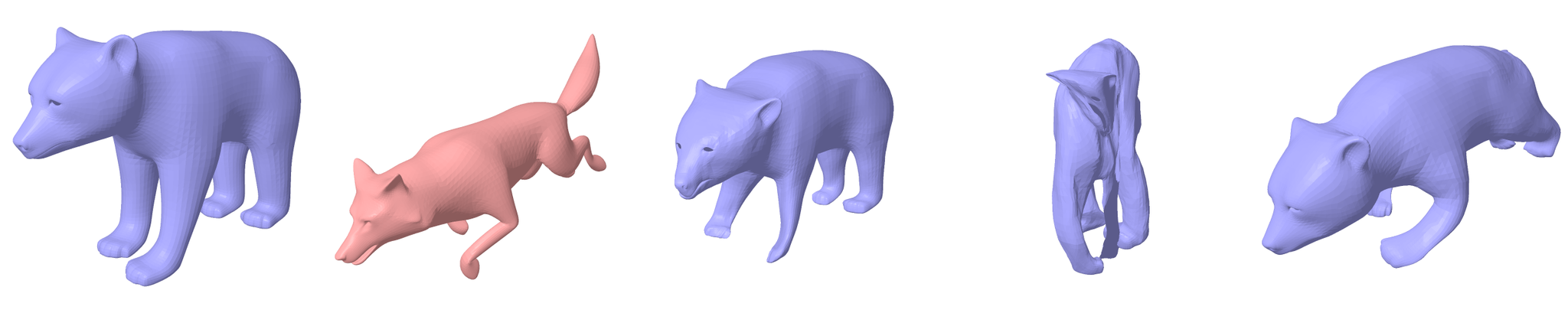}} \\
            \multicolumn{5}{c}{\includegraphics[width=\linewidth]{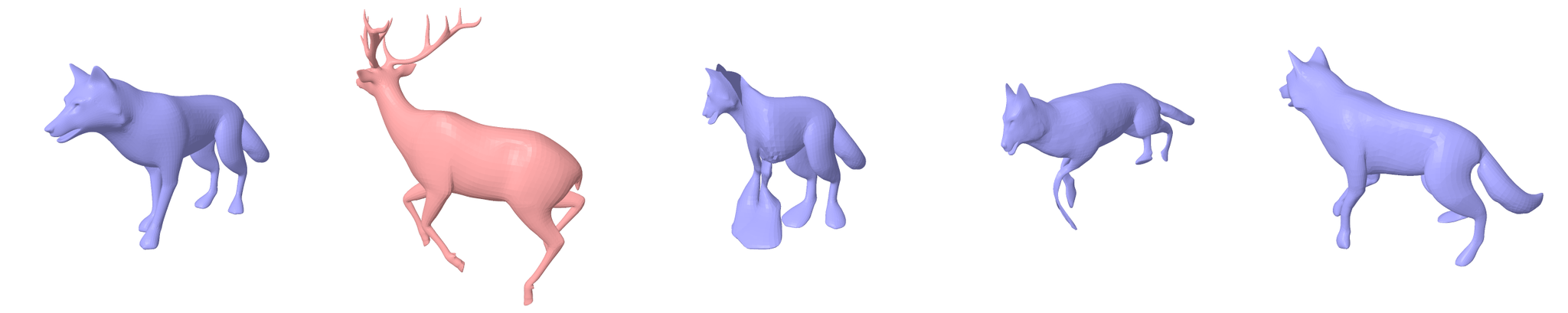}} \\
            \multicolumn{5}{c}{\includegraphics[width=\linewidth]{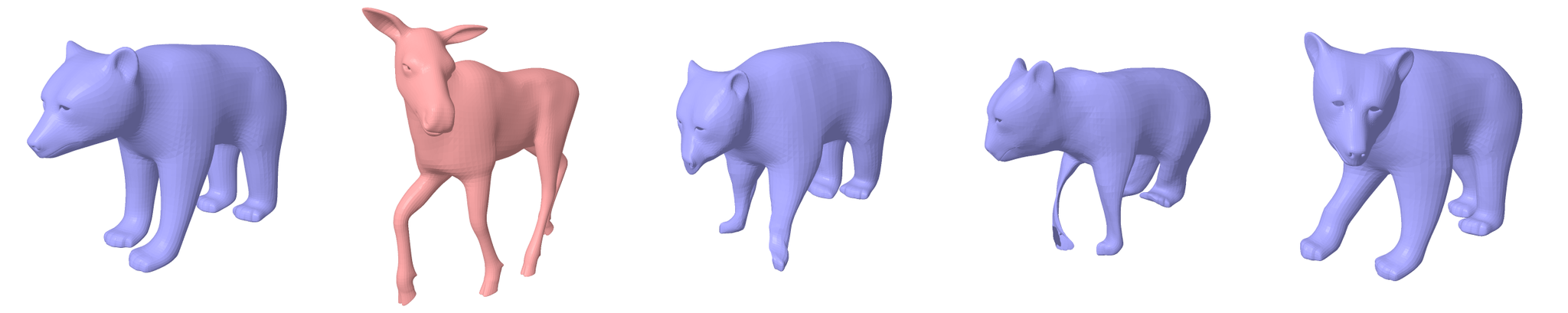}} \\
            \multicolumn{5}{c}{\includegraphics[width=\linewidth]{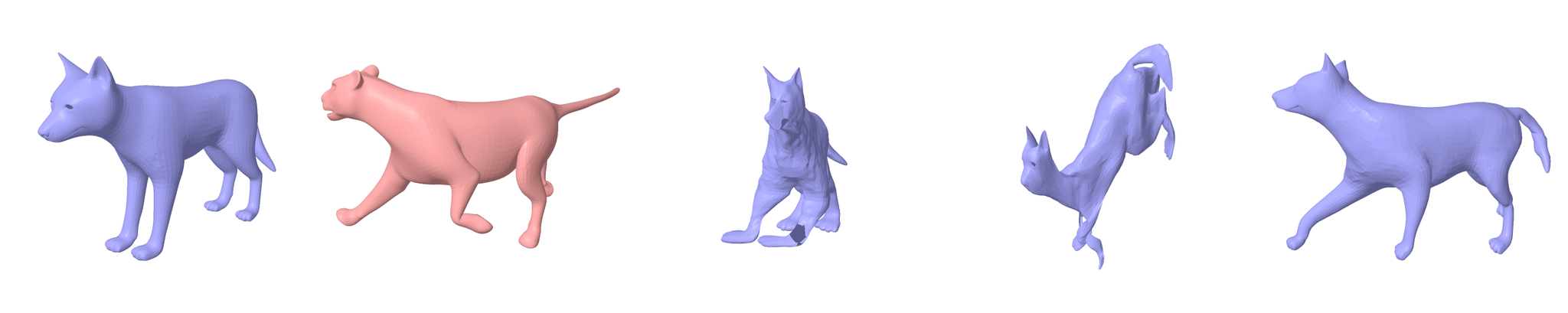}} \\
            \multicolumn{5}{c}{\includegraphics[width=\linewidth]{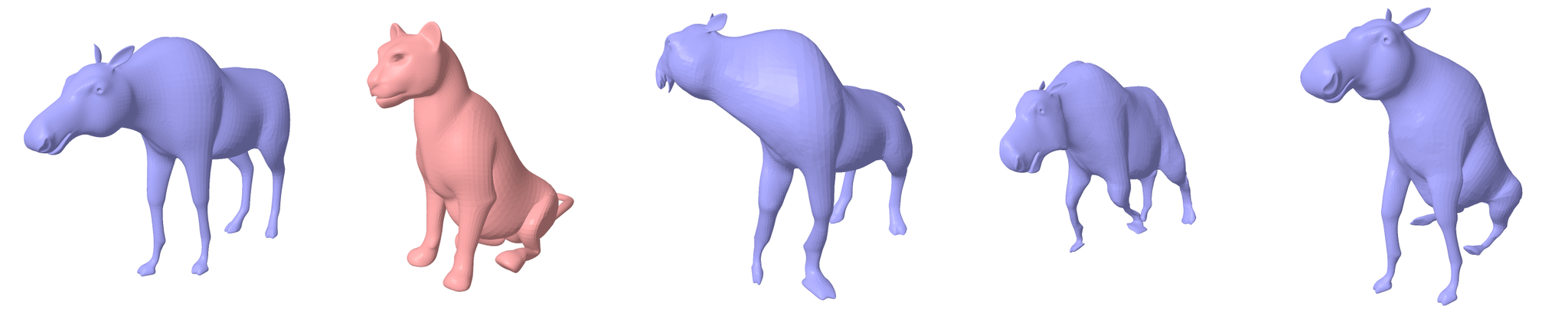}} \\
        \end{tabularx}
    \end{minipage}\hfill\vline\hfill%
    \begin{minipage}{0.49\textwidth}
        \centering
        \begin{tabularx}{\linewidth}{YYYYY}
            \makecell{$\overline{\mathcal{M}}^T$} & \makecell{$\mathcal{M}^S$} & \makecell{NJF~\cite{Aigerman:2022NJF}} & \makecell{ZPT~\cite{Wang:2023ZPT}} & \makecell{Ours} \\
            \multicolumn{5}{c}{\includegraphics[width=\linewidth]{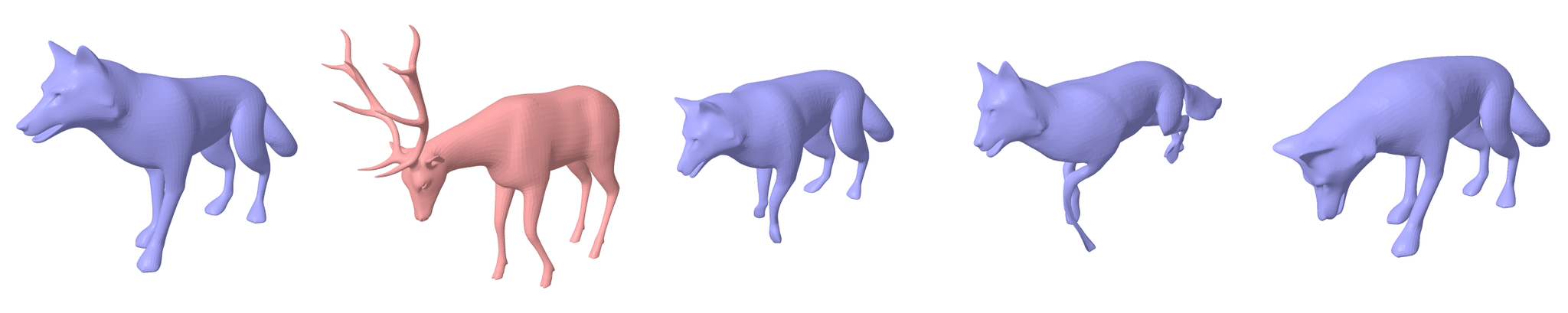}} \\
            \multicolumn{5}{c}{\includegraphics[width=\linewidth]{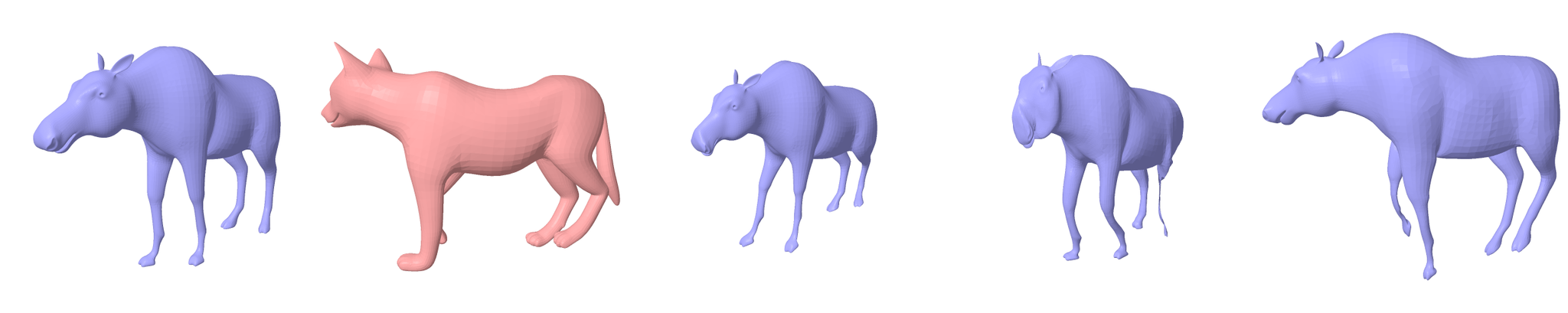}} \\
            \multicolumn{5}{c}{\includegraphics[width=\linewidth]{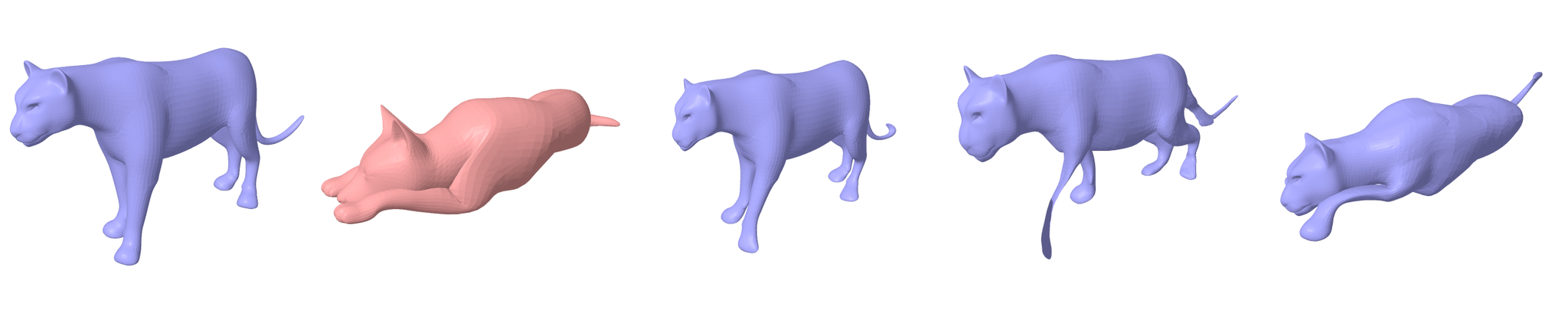}} \\
            \multicolumn{5}{c}{\includegraphics[width=\linewidth]{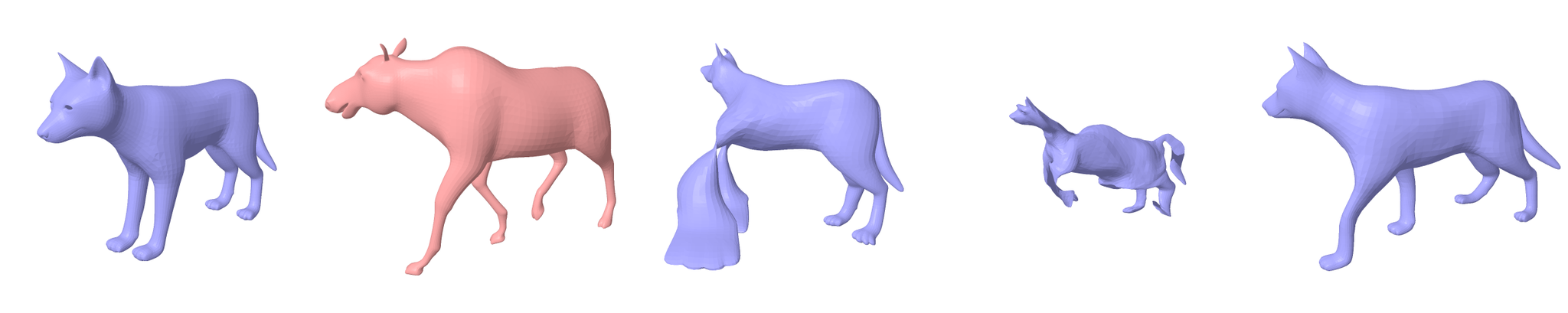}} \\
            \multicolumn{5}{c}{\includegraphics[width=\linewidth]{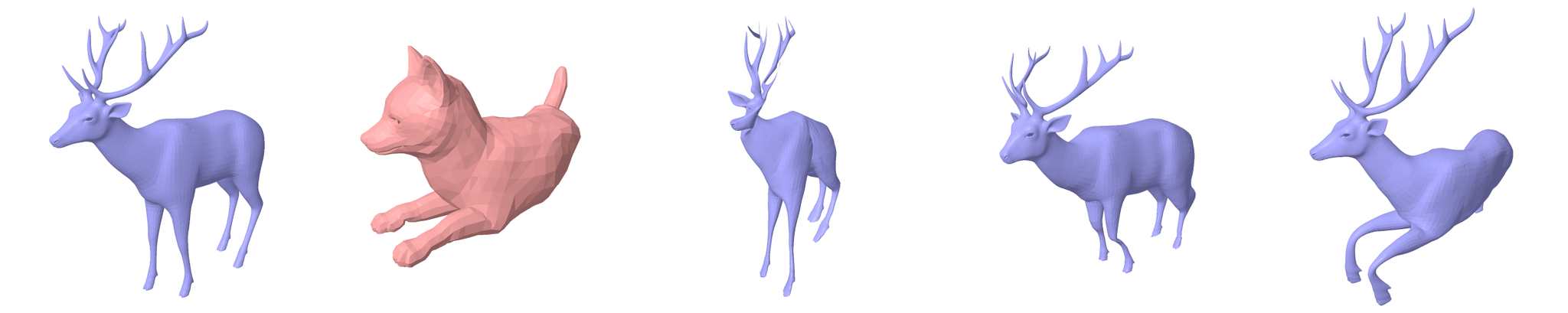}} \\
            \multicolumn{5}{c}{\includegraphics[width=\linewidth]{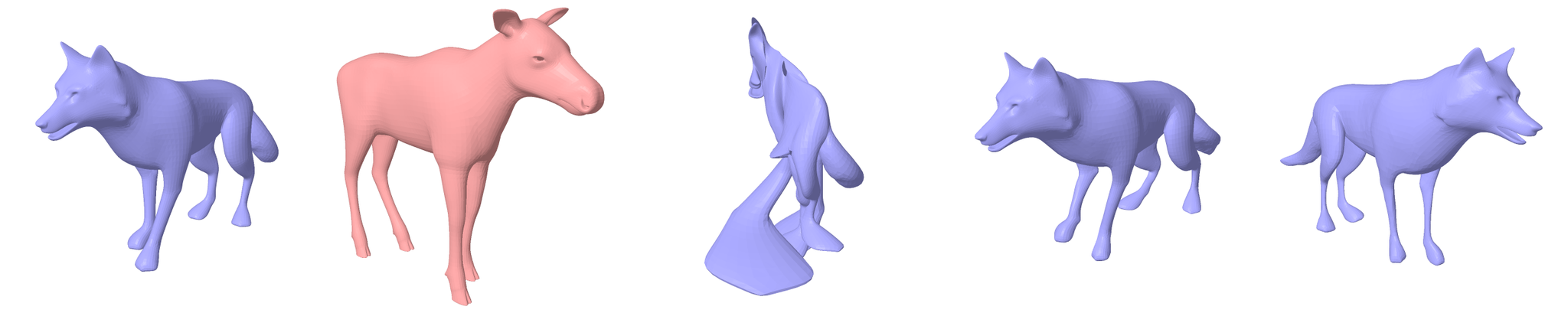}} \\
            \multicolumn{5}{c}{\includegraphics[width=\linewidth]{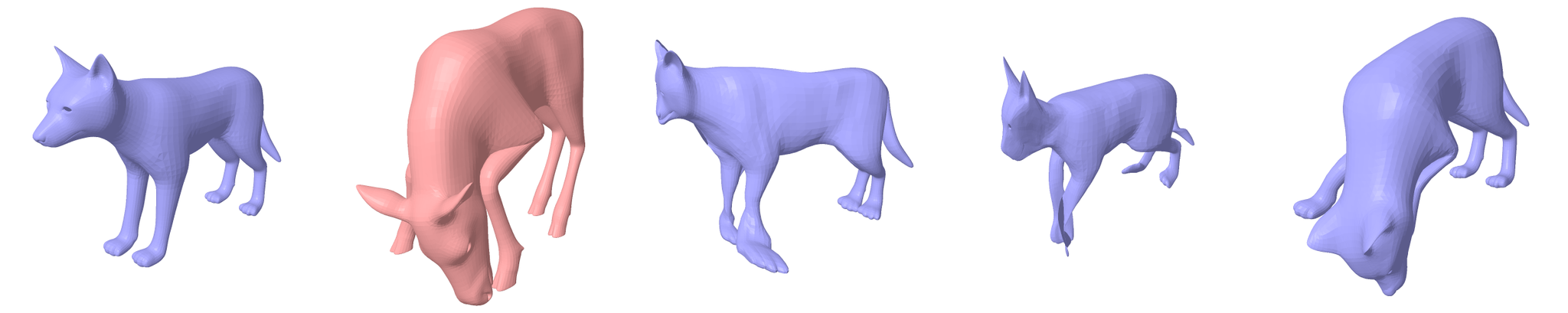}} \\
            \multicolumn{5}{c}{\includegraphics[width=\linewidth]{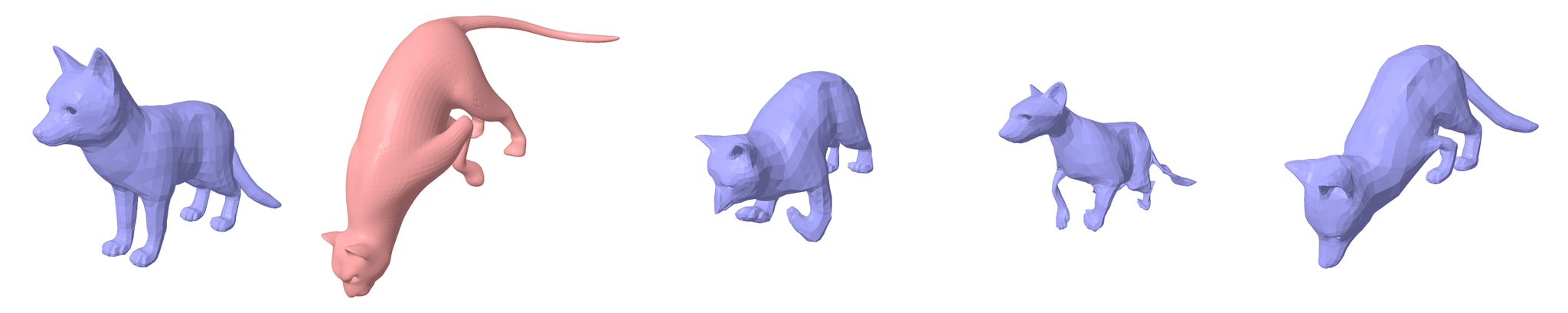}} \\
        \end{tabularx}
    \end{minipage}
}
\caption{Qualitative results of pose transfer across DeformingThings4D animals~\cite{Li:2021Deform4D}.}
\label{fig:deform4d_transfer_supp}
\end{figure*}

\begin{figure*}[!h]
    \centering
    \setlength{\tabcolsep}{0em}
    \def\arraystretch{0.0}
    \begin{tabular}{P{0.1\textwidth}P{0.180\textwidth}P{0.180\textwidth}P{0.180\textwidth}P{0.180\textwidth}P{0.180\textwidth}}
        $\overline{\mathcal{M}}^T$ & $\mathcal{M}^S$ & NJF~\cite{Aigerman:2022NJF} & SPT~\cite{Liao:2022SPT} & ZPT~\cite{Wang:2023ZPT} & Ours \\
        \multicolumn{6}{c}{\includegraphics[width=\textwidth]{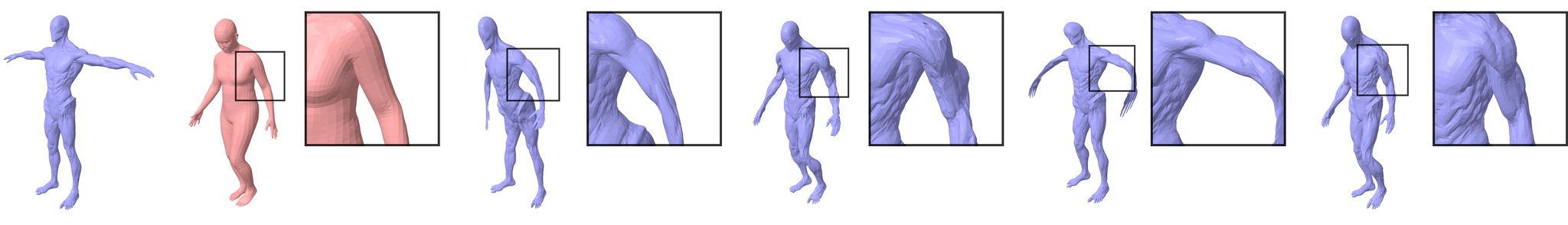}} \\
        \multicolumn{6}{c}{\includegraphics[width=\textwidth]{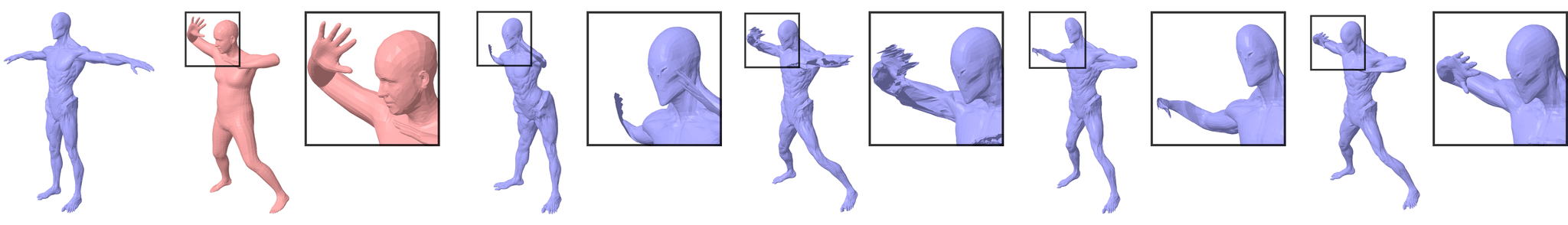}} \\
        \multicolumn{6}{c}{\includegraphics[width=\textwidth]{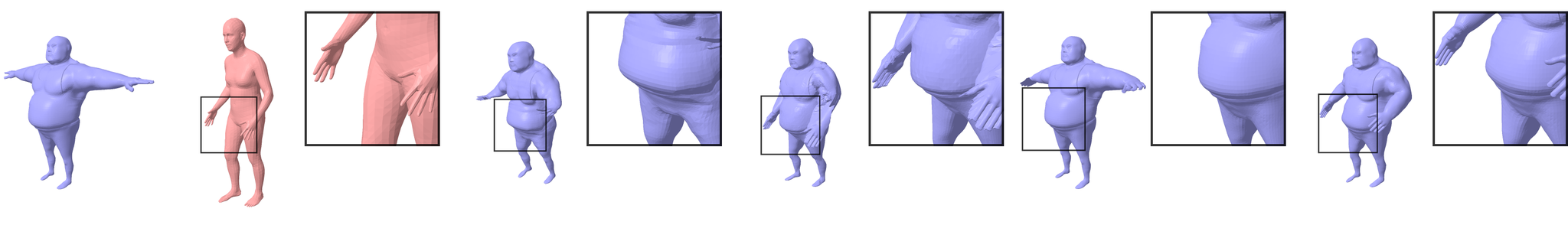}} \\
        \multicolumn{6}{c}{\includegraphics[width=\textwidth]{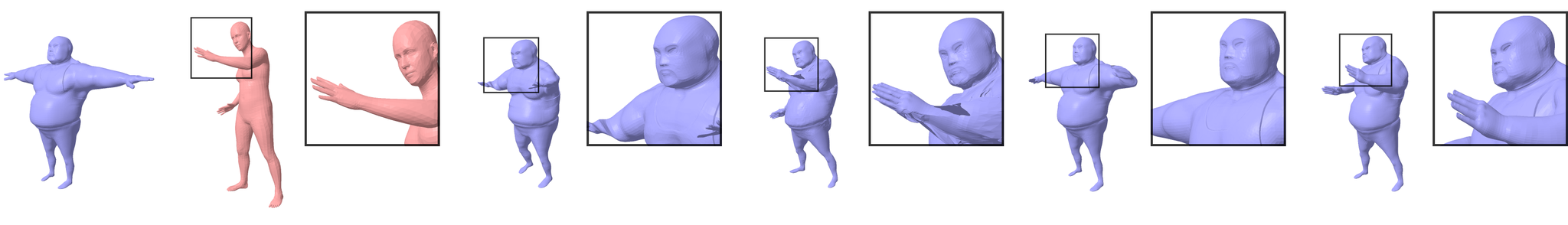}} \\
        \multicolumn{6}{c}{\includegraphics[width=\textwidth]{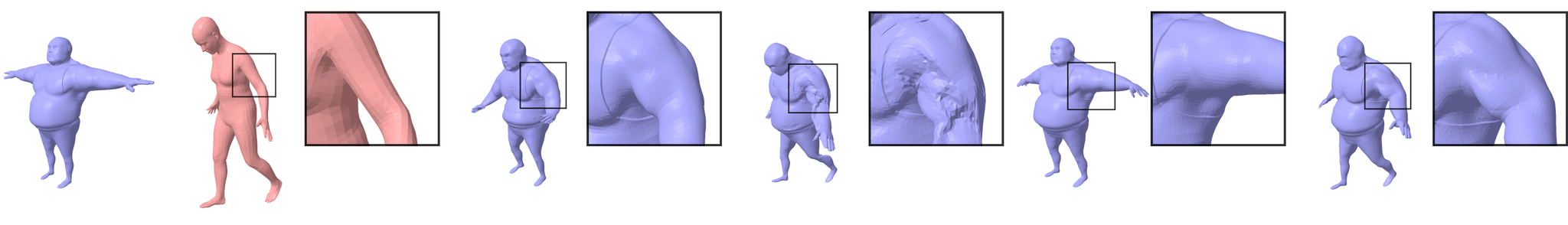}} \\
        \multicolumn{6}{c}{\includegraphics[width=\textwidth]{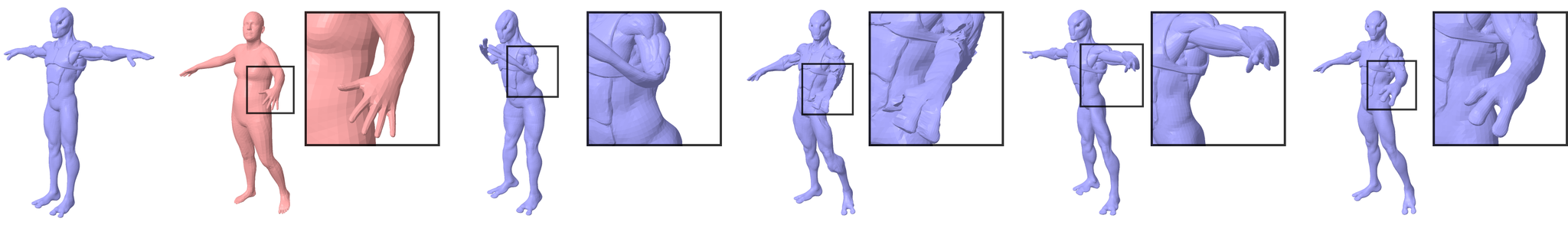}} \\
        \multicolumn{6}{c}{\includegraphics[width=\textwidth]{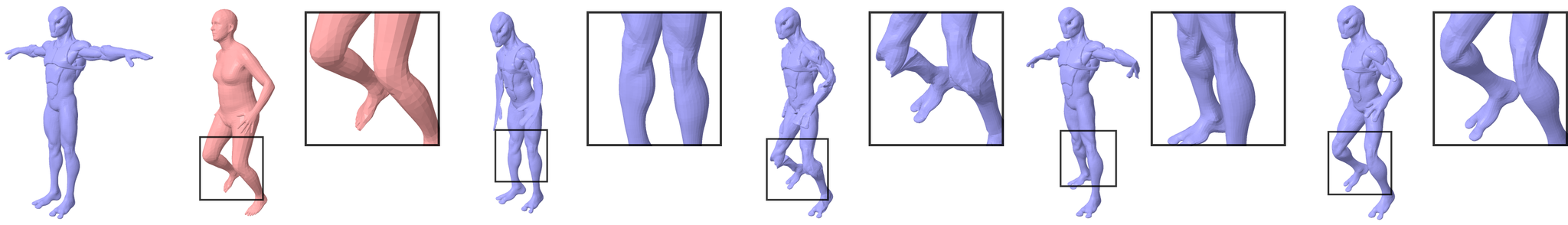}} \\
        \multicolumn{6}{c}{\includegraphics[width=\textwidth]{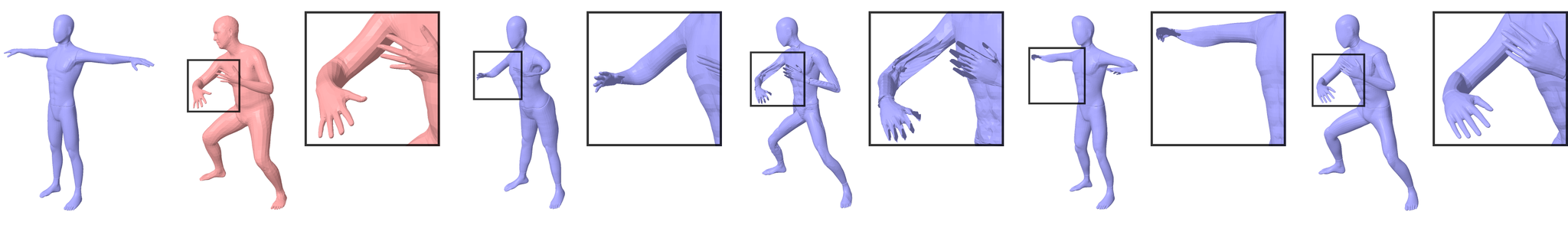}} \\
        \multicolumn{6}{c}{\includegraphics[width=\textwidth]{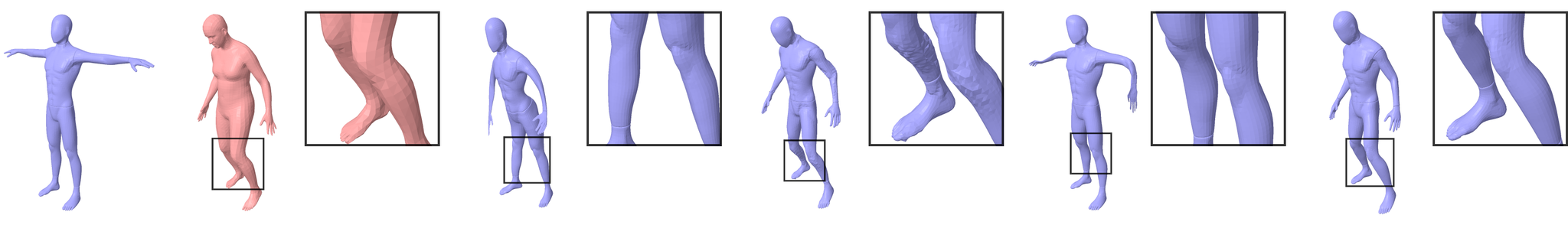}} \\
        \multicolumn{6}{c}{\includegraphics[width=\textwidth]{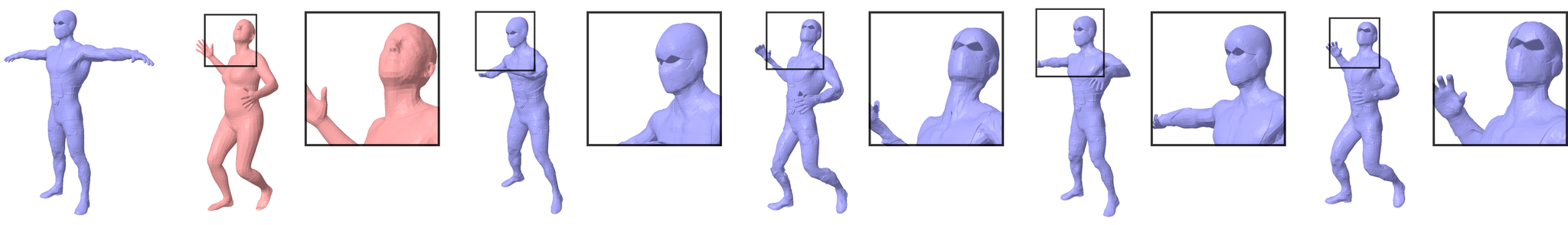}} \\
    \end{tabular}
\caption{Qualitative results of pose transfer from a SMPL~\cite{Loper:2015SMPL} mesh to Mixamo characters~\cite{Mixamo:2020}. Best viewed when zoomed in.}
\label{fig:mixamo_transfer_supp}
\vspace{-1\baselineskip}
\end{figure*}

\begin{figure*}[!h]
    \centering
    \setlength{\tabcolsep}{0em}
    \def\arraystretch{0.0}
    \begin{tabular}{P{0.092\textwidth}P{0.092\textwidth}P{0.163\textwidth}P{0.163\textwidth}P{0.163\textwidth}P{0.163\textwidth}P{0.164\textwidth}}
        $\overline{\mathcal{M}}^T$ & $\mathcal{M}^S$ & NJF~\cite{Aigerman:2022NJF} & SPT~\cite{Liao:2022SPT} & ZPT~\cite{Wang:2023ZPT} & Ours & $\mathcal{M}^T_{\text{GT}}$\\
        \multicolumn{7}{c}{\includegraphics[width=\textwidth]{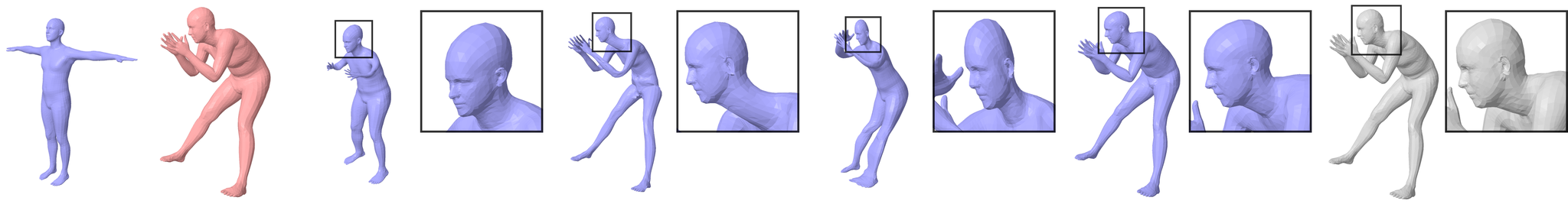}} \\
        \multicolumn{7}{c}{\includegraphics[width=\textwidth]{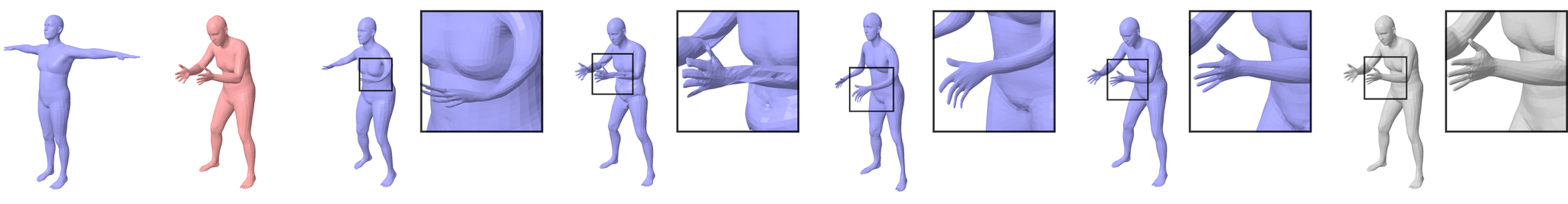}} \\
        \multicolumn{7}{c}{\includegraphics[width=\textwidth]{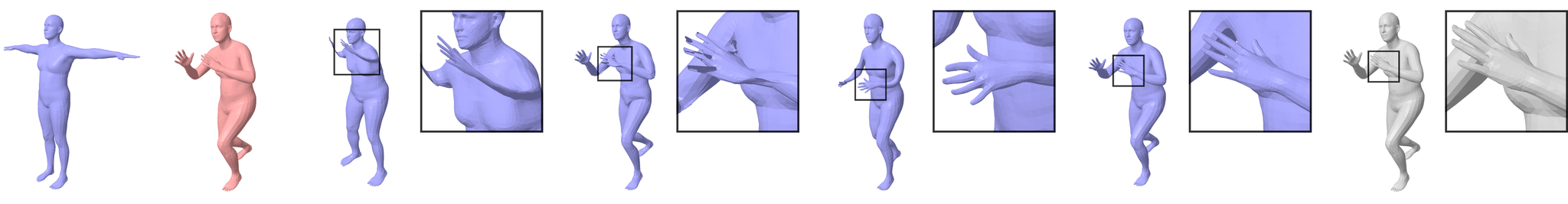}} \\
        \multicolumn{7}{c}{\includegraphics[width=\textwidth]{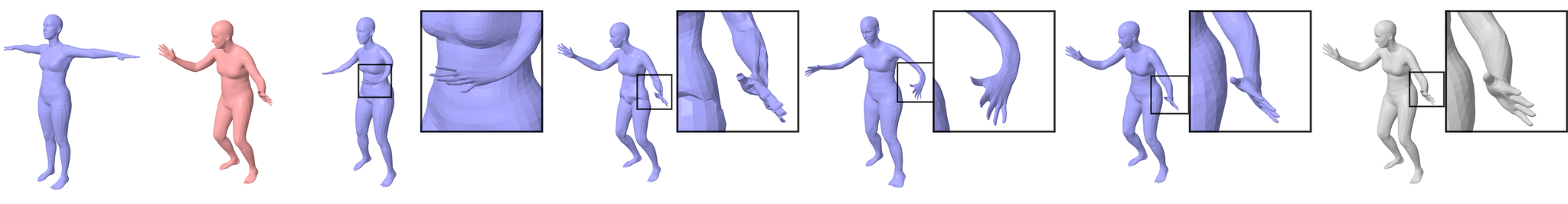}} \\
        \multicolumn{7}{c}{\includegraphics[width=\textwidth]{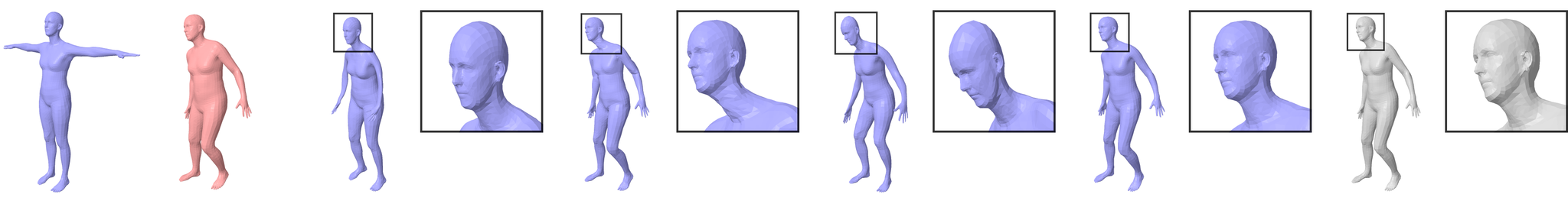}} \\
        \multicolumn{7}{c}{\includegraphics[width=\textwidth]{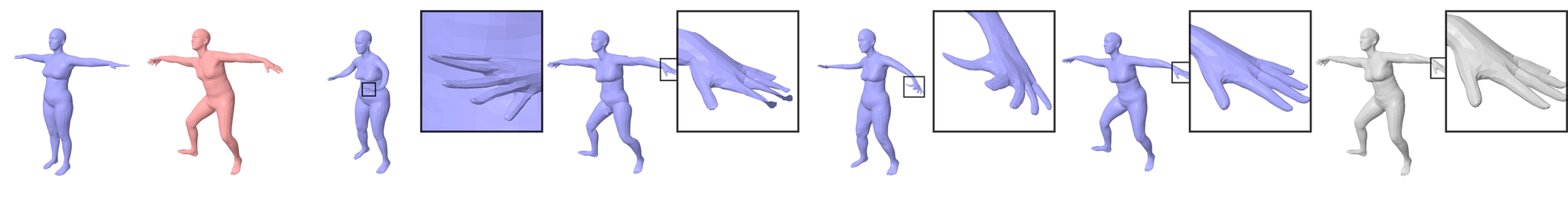}} \\
        \multicolumn{7}{c}{\includegraphics[width=\textwidth]{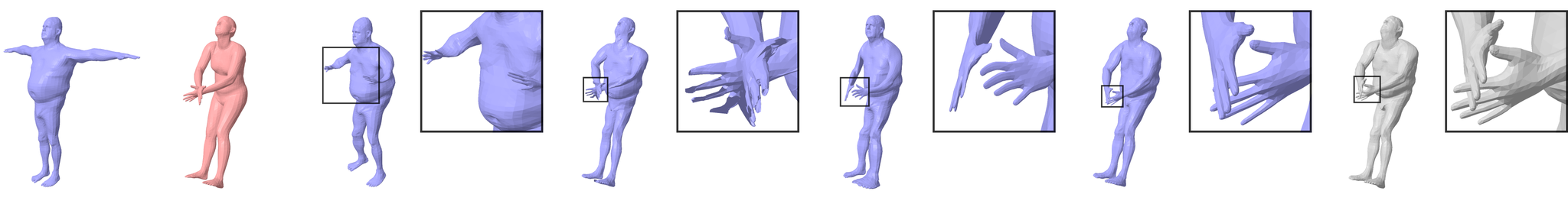}} \\
        \multicolumn{7}{c}{\includegraphics[width=\textwidth]{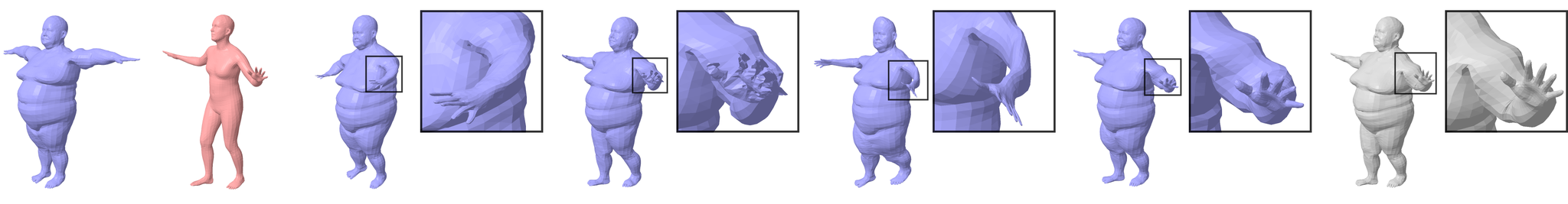}} \\
        \multicolumn{7}{c}{\includegraphics[width=\textwidth]{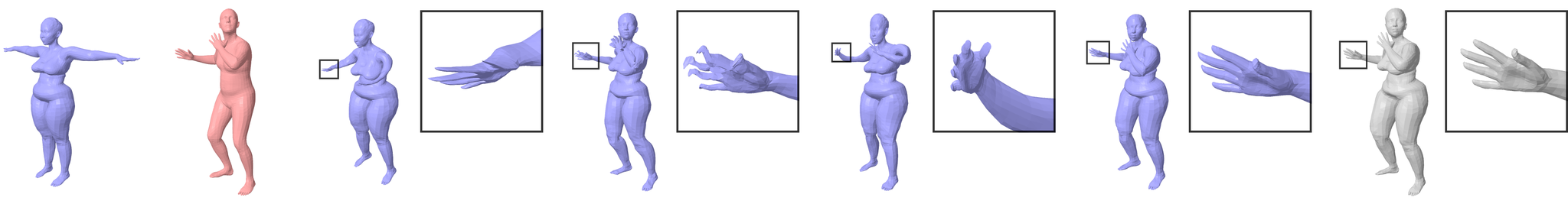}} \\
        \multicolumn{7}{c}{\includegraphics[width=\textwidth]{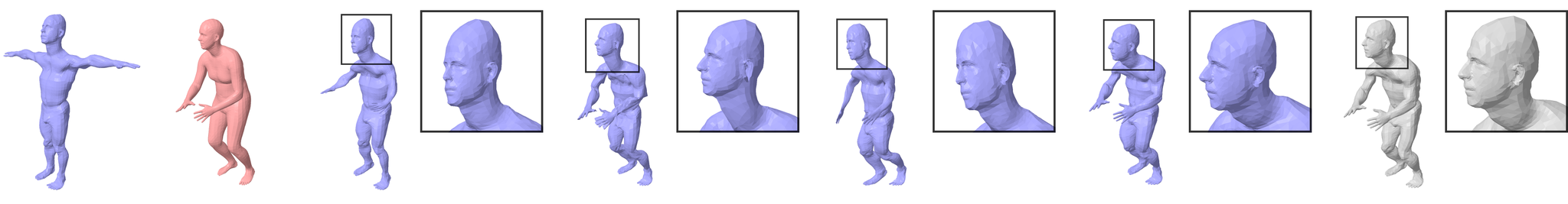}} \\
        \multicolumn{7}{c}{\includegraphics[width=\textwidth]{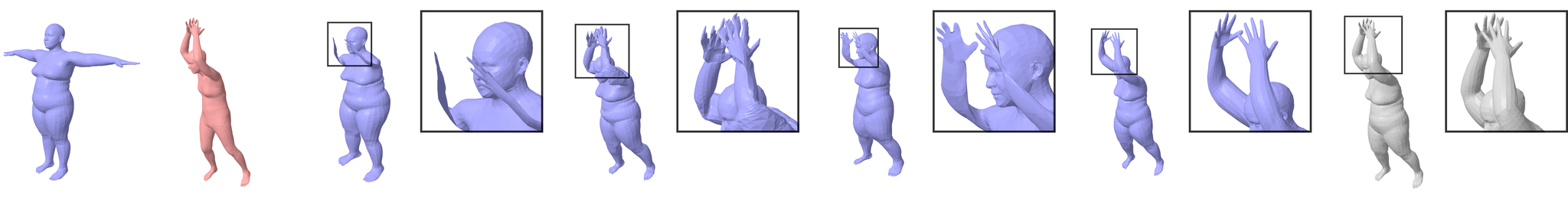}} \\
    \end{tabular}
\caption{Qualitative results of pose transfer across different SMPL~\cite{Loper:2015SMPL} human body shapes. Best viewed when zoomed-in.}
\label{fig:smpl_transfer_error_supp}
\vspace{-1\baselineskip}
\end{figure*}

\vspace{-0.5\baselineskip}
\begin{figure*}[!h]
{
    \centering
    \setlength{\tabcolsep}{0em}
    \def\arraystretch{0.0}
    \begin{minipage}{0.49\textwidth}
        \centering
        \begin{tabularx}{\linewidth}{YYYY}
            \makecell{$\mathcal{M}^S$ \\ (Vertex)} & \makecell{$\mathcal{M}^S$ \\ (Ours)} & \makecell{$\mathcal{M}^T$ \\ (Vertex)} & \makecell{$\mathcal{M}^T$ \\ (Ours)} \\
            \multicolumn{4}{c}{\includegraphics[width=\textwidth]{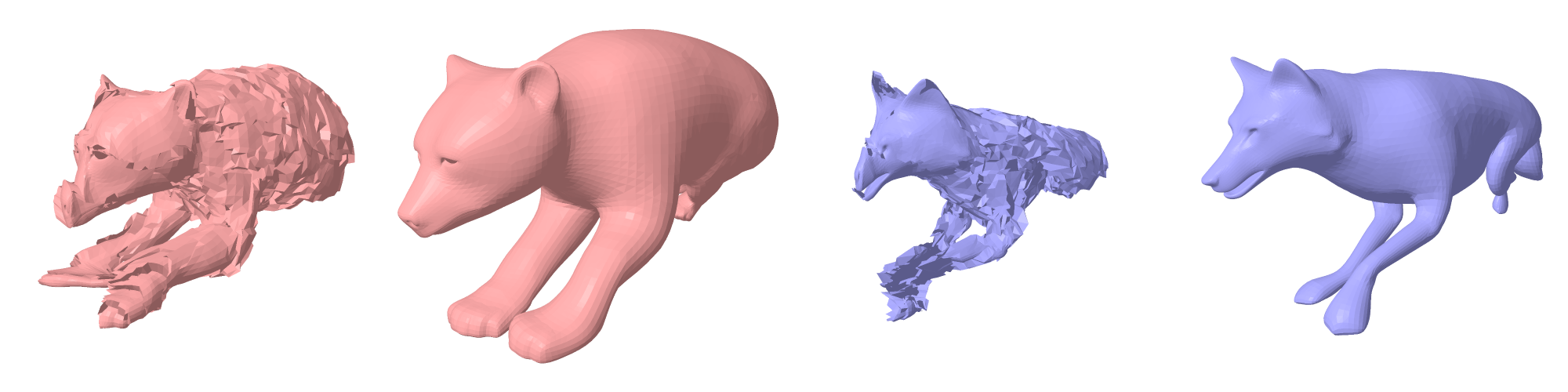}} \\
            \multicolumn{4}{c}{\includegraphics[width=\textwidth]{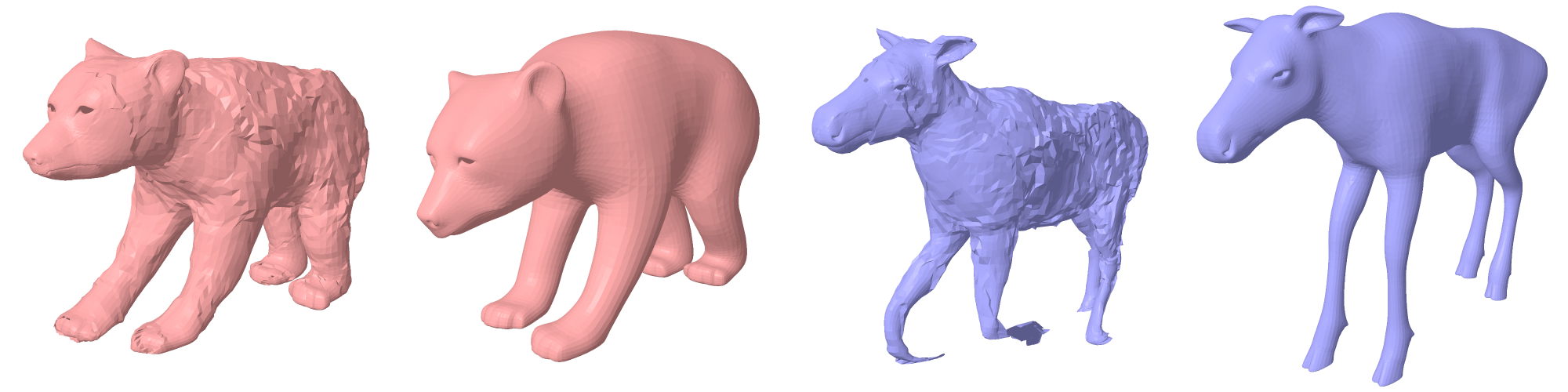}} \\
            \multicolumn{4}{c}{\includegraphics[width=\textwidth]{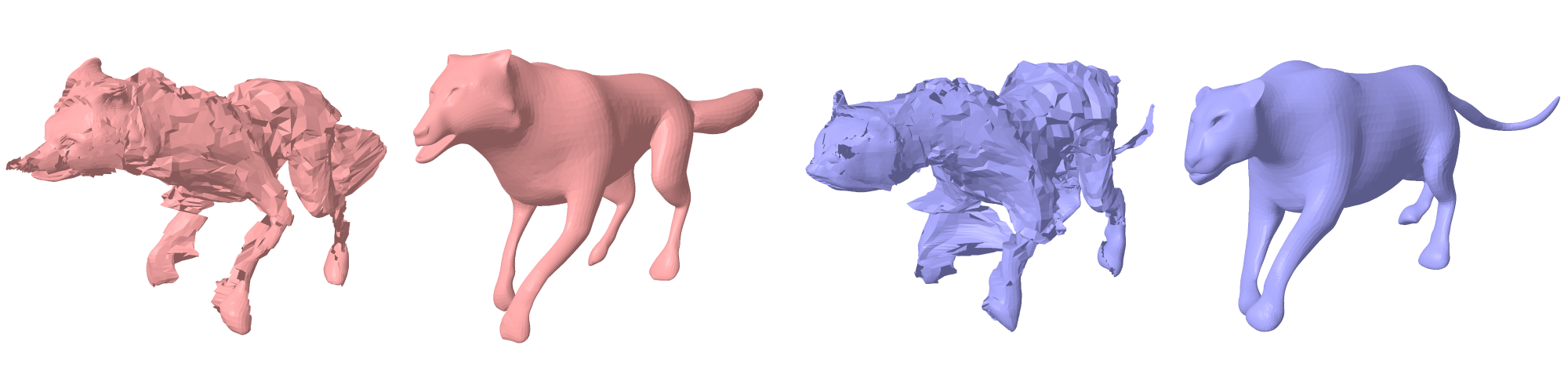}} \\
            \multicolumn{4}{c}{\includegraphics[width=\textwidth]{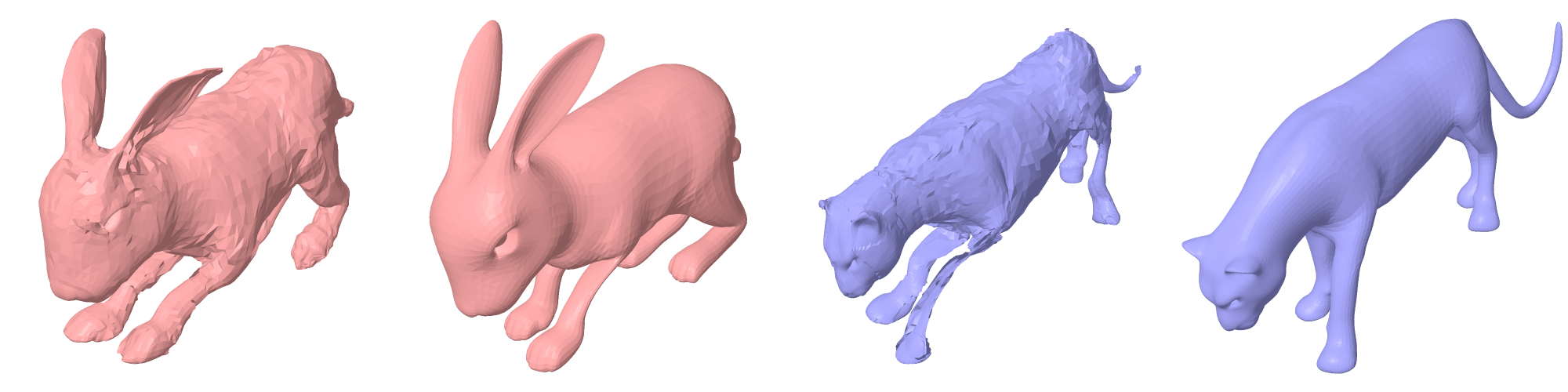}} \\
            \multicolumn{4}{c}{\includegraphics[width=\textwidth]{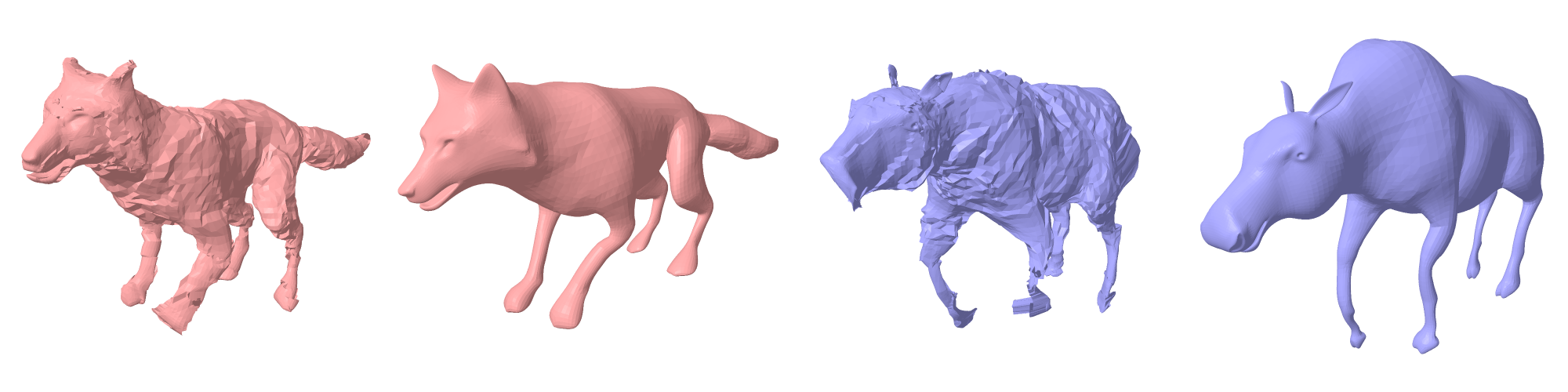}} \\
            \multicolumn{4}{c}{\includegraphics[width=\textwidth]{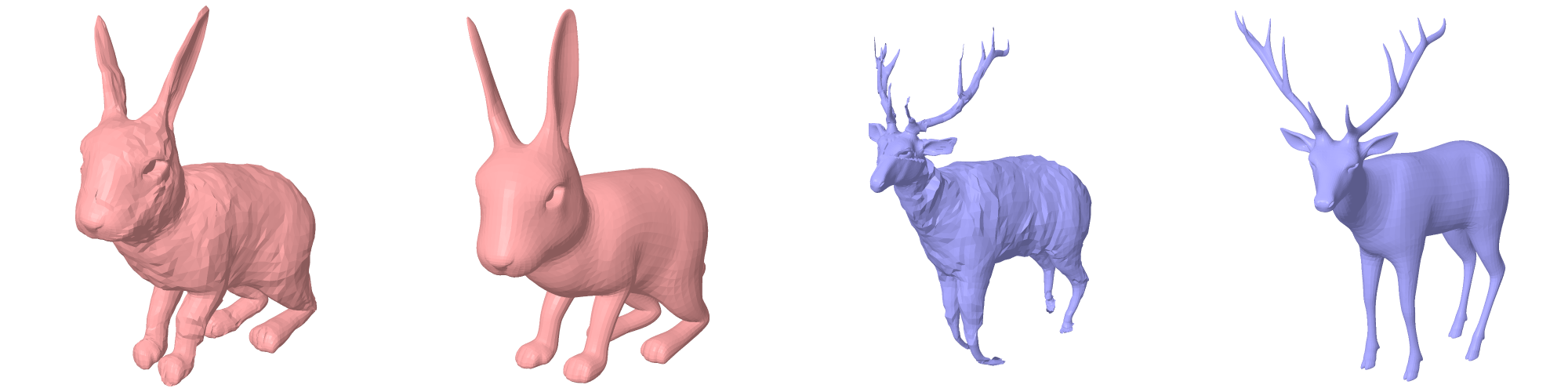}} \\
            \multicolumn{4}{c}{\includegraphics[width=\textwidth]{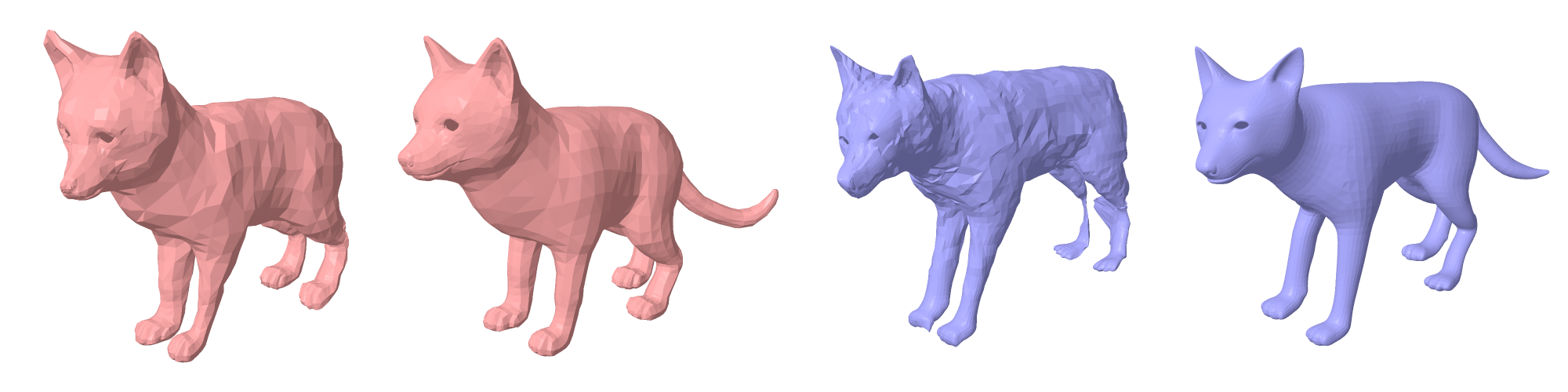}} \\
            \multicolumn{4}{c}{\includegraphics[width=\textwidth]{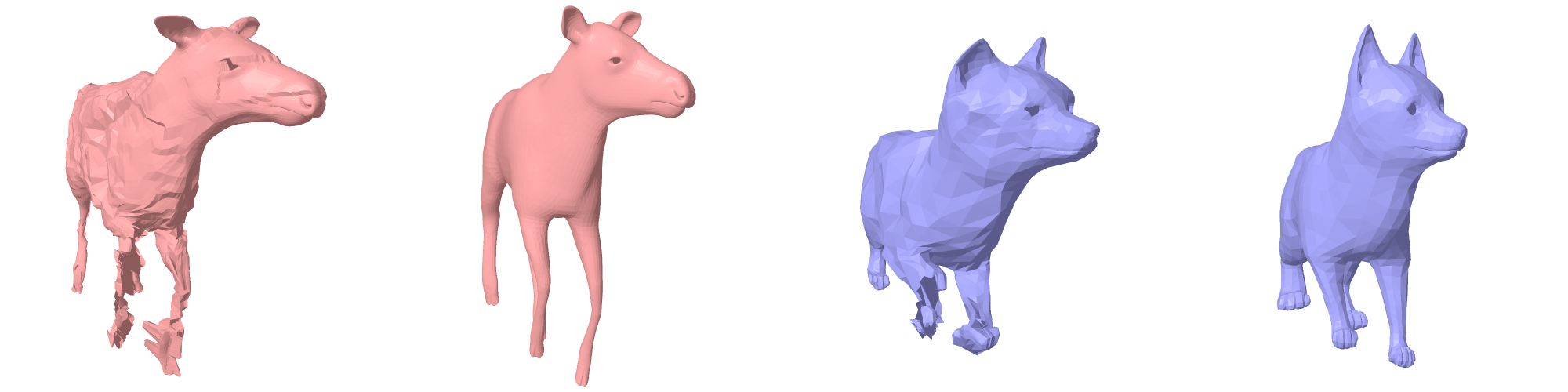}} \\
            \multicolumn{4}{c}{\includegraphics[width=\textwidth]{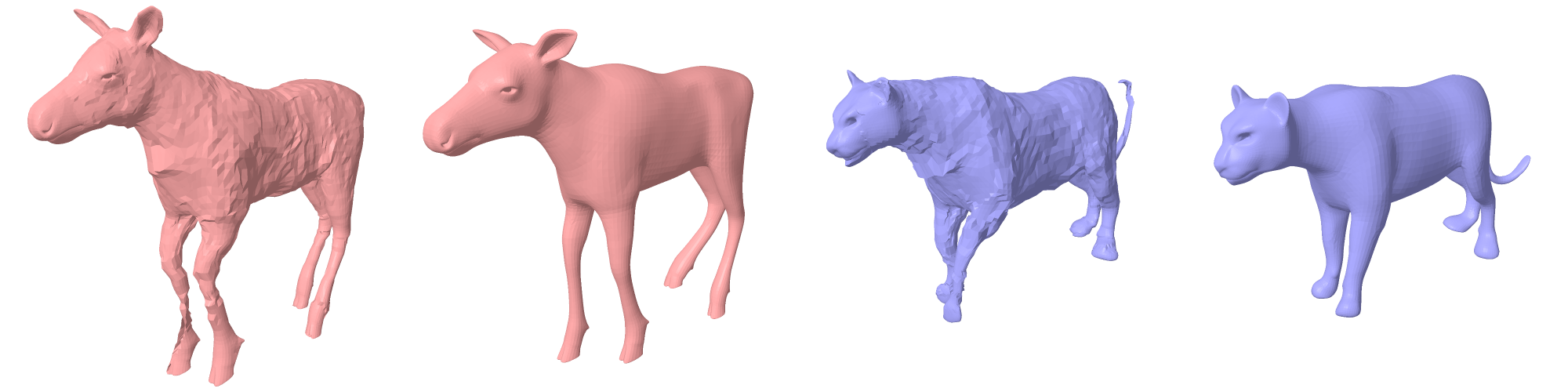}} \\
        \end{tabularx}
    \end{minipage}\hfill\vline\hfill%
    \begin{minipage}{0.49\textwidth}
        \centering
        \begin{tabularx}{\linewidth}{YYYY}
            \makecell{$\mathcal{M}^S$ \\ (Vertex)} & \makecell{$\mathcal{M}^S$ \\ (Ours)} & \makecell{$\mathcal{M}^T$ \\ (Vertex)} & \makecell{$\mathcal{M}^T$ \\ (Ours)} \\
            \multicolumn{4}{c}{\includegraphics[width=\textwidth]{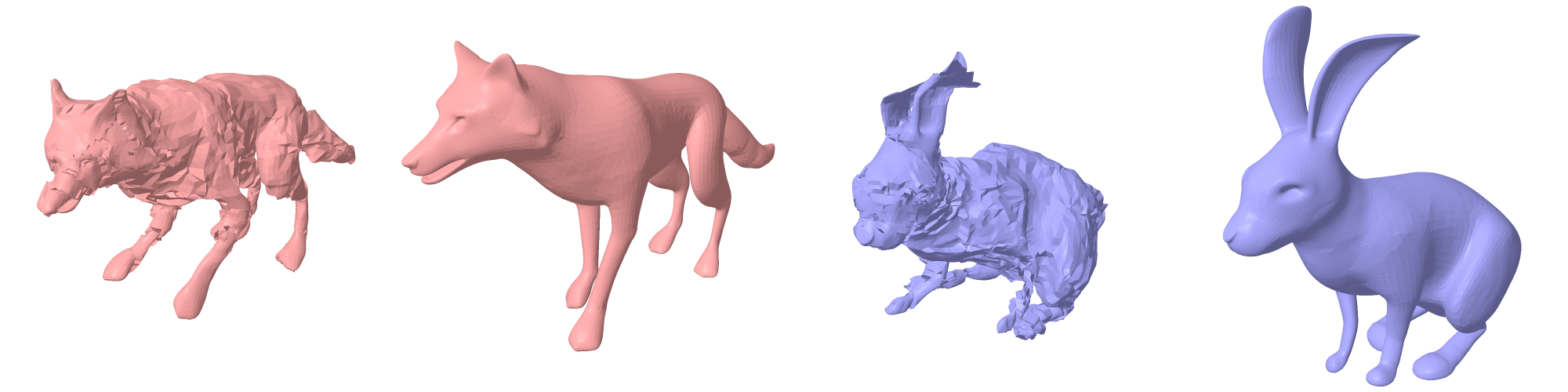}} \\
            \multicolumn{4}{c}{\includegraphics[width=\textwidth]{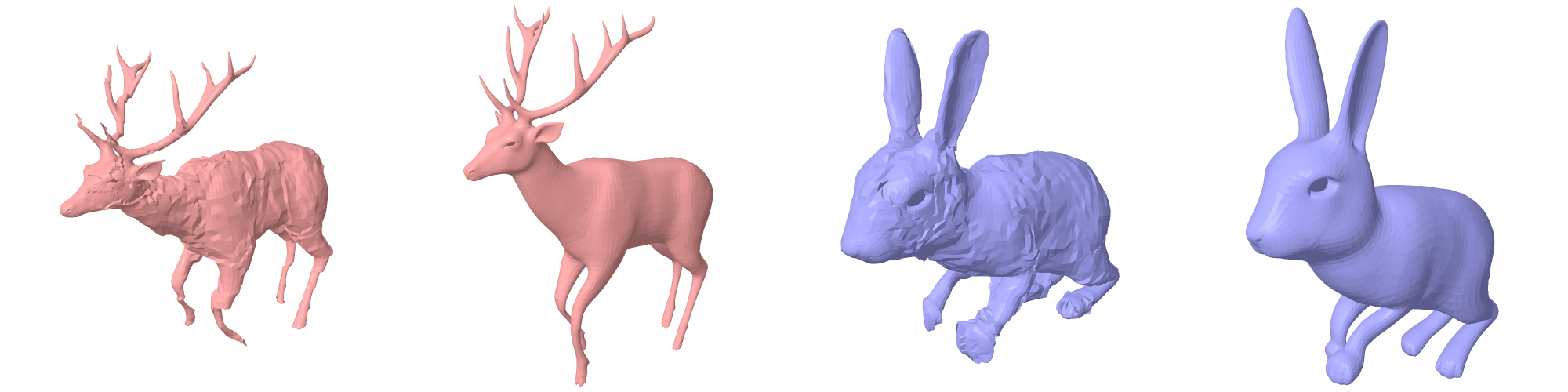}} \\
            \multicolumn{4}{c}{\includegraphics[width=\textwidth]{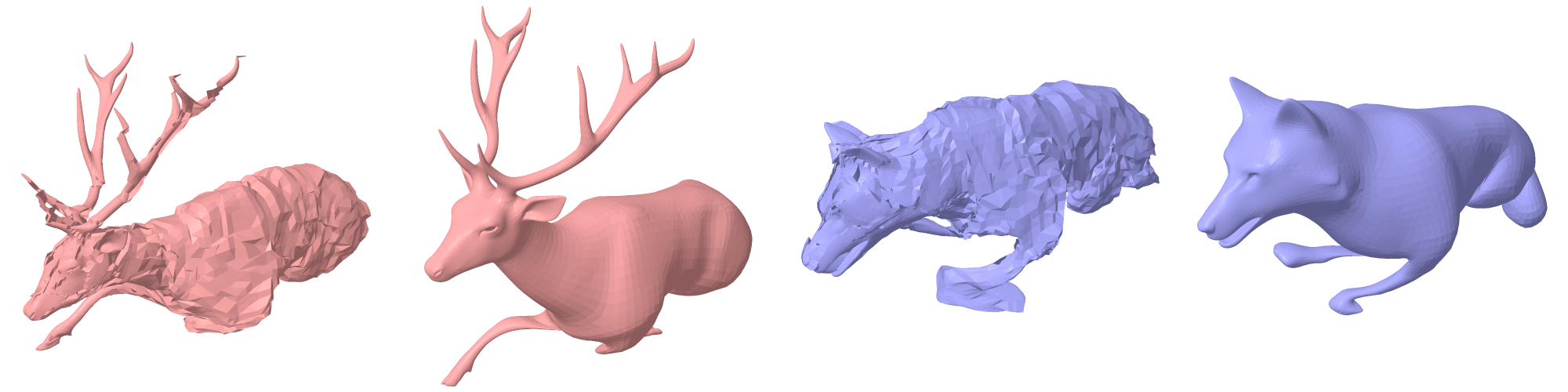}} \\
            \multicolumn{4}{c}{\includegraphics[width=\textwidth]{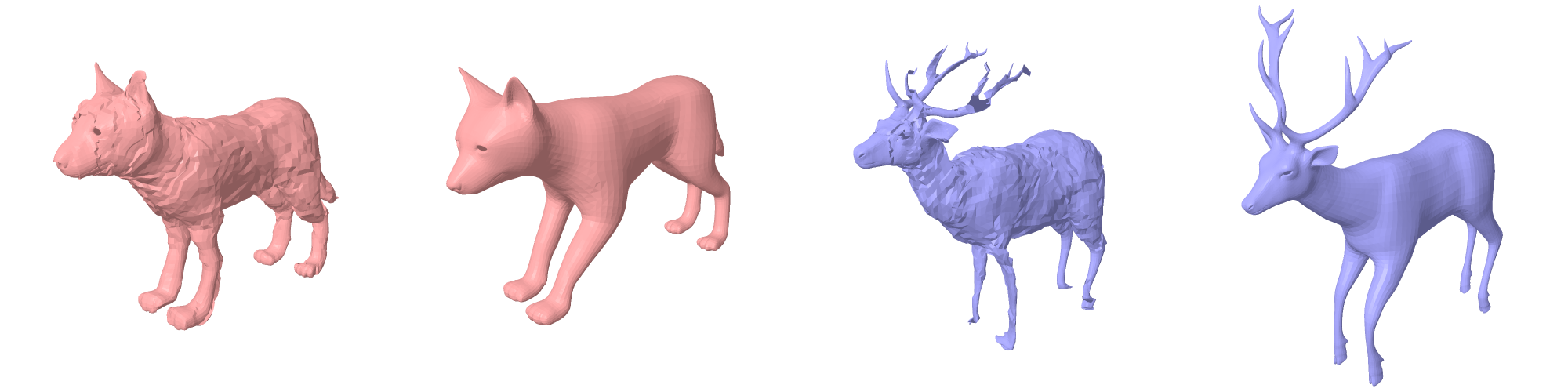}} \\
            \multicolumn{4}{c}{\includegraphics[width=\textwidth]{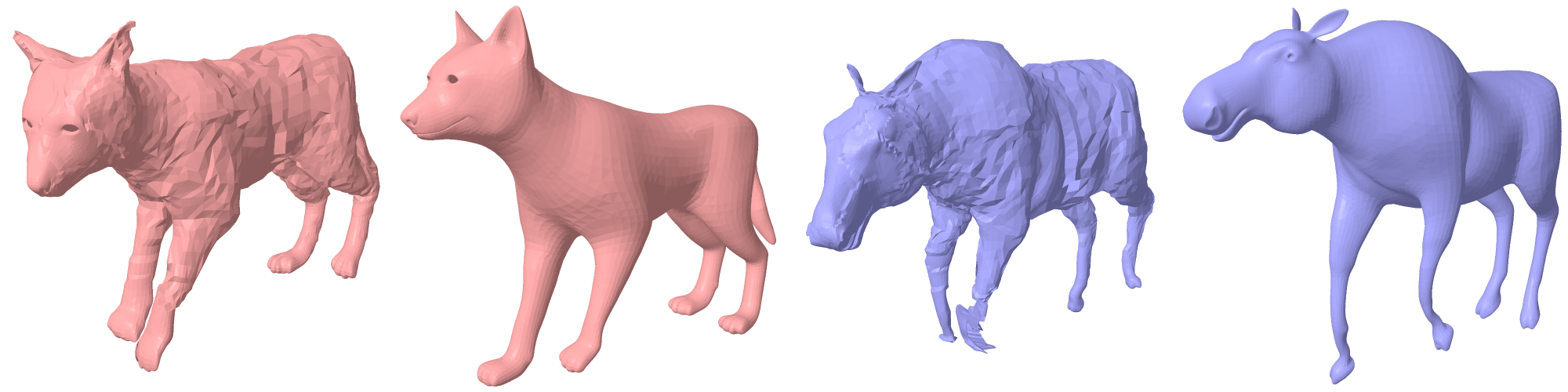}} \\
            \multicolumn{4}{c}{\includegraphics[width=\textwidth]{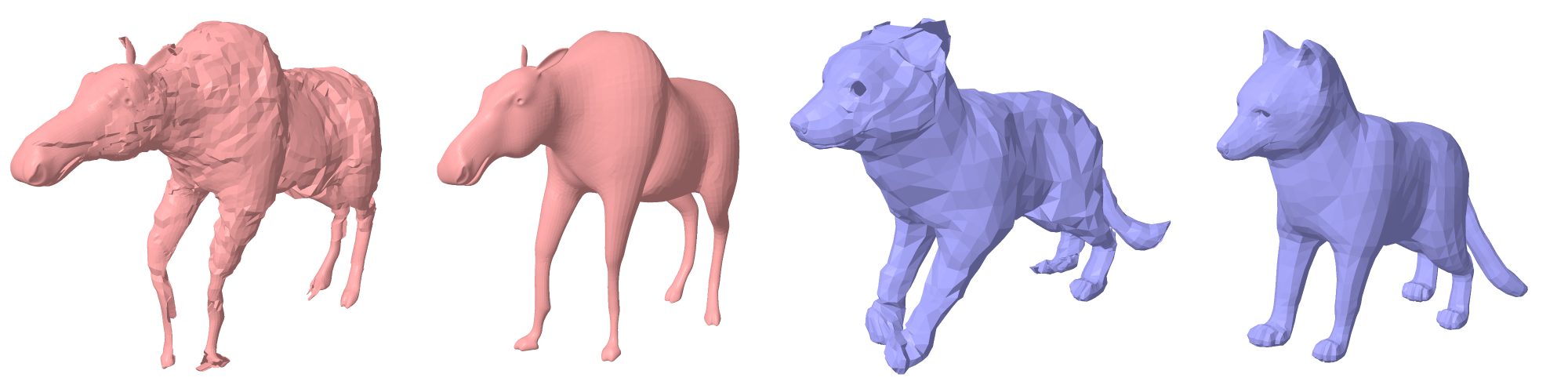}} \\
            \multicolumn{4}{c}{\includegraphics[width=\textwidth]{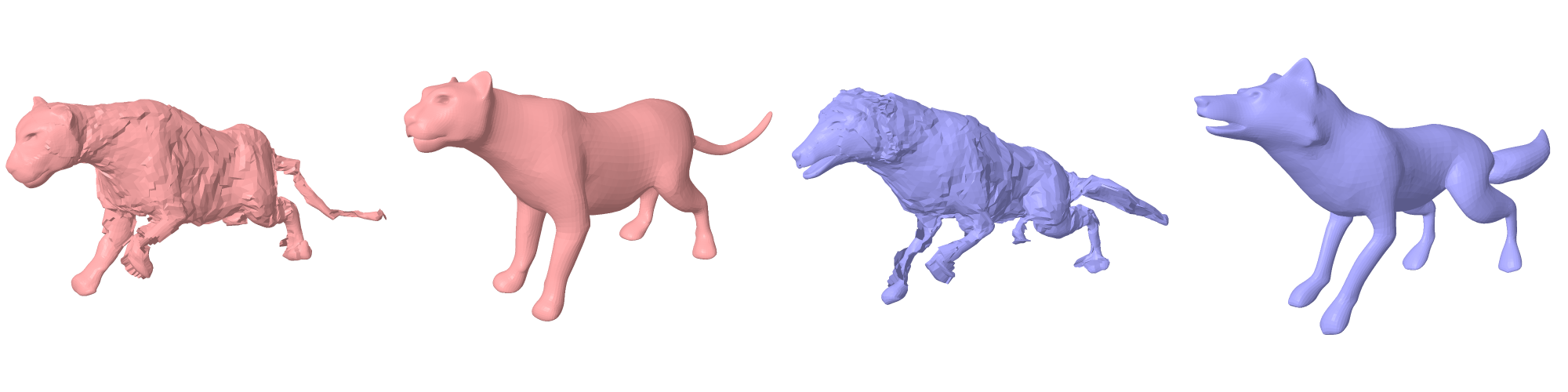}} \\
            \multicolumn{4}{c}{\includegraphics[width=\textwidth]{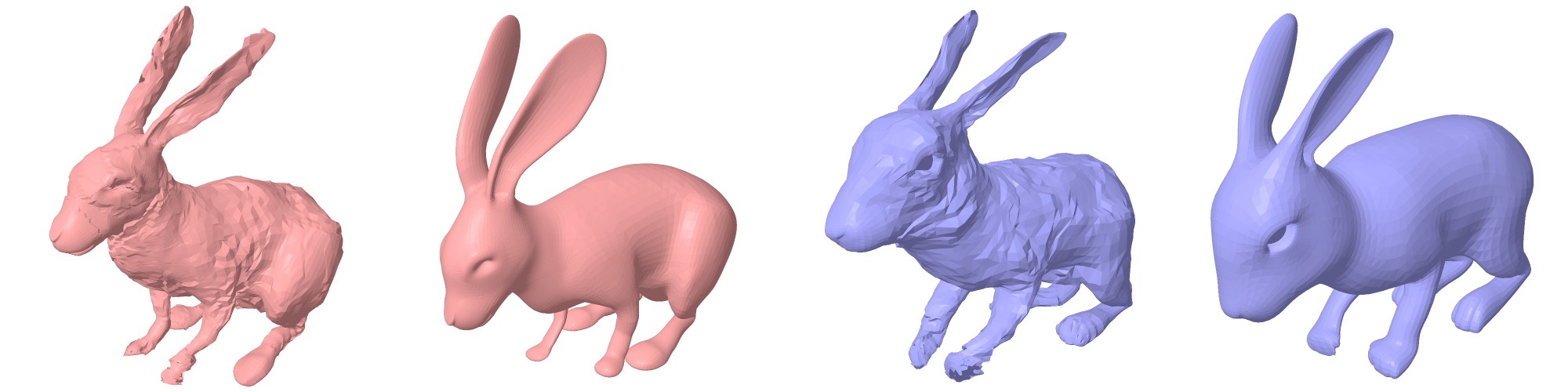}} \\
            \multicolumn{4}{c}{\includegraphics[width=\textwidth]{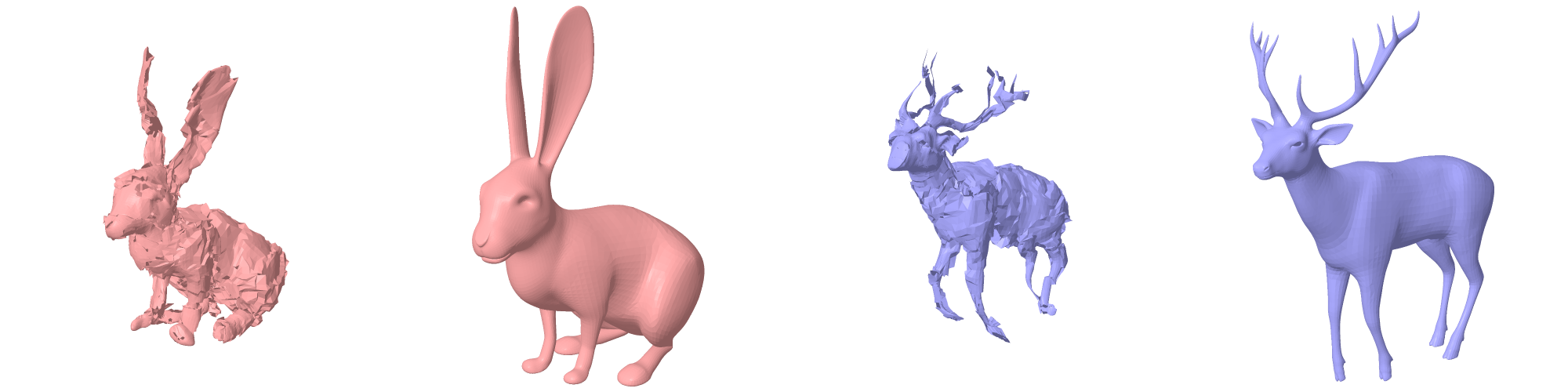}} \\
        \end{tabularx}
    \end{minipage}
}
\caption{Unconditional generation results. Each row illustrates the outcome of directly applying the generated poses to the source shape $\mathcal{M}^S$ and then transferring them to various target shapes $\mathcal{M}^T$.}
\label{fig:uncond_gen_supp}
\end{figure*}

\clearpage
\newpage
\newpage
\section*{NeurIPS Paper Checklist}

\begin{enumerate}

\item {\bf Claims}
    \item[] Question: Do the main claims made in the abstract and introduction accurately reflect the paper's contributions and scope?
    \item[] Answer: \answerYes{} %
    \item[] Justification: We fully discuss the claims made in the abstract and introduction throughout the paper.
    \item[] Guidelines:
    \begin{itemize}
        \item The answer NA means that the abstract and introduction do not include the claims made in the paper.
        \item The abstract and/or introduction should clearly state the claims made, including the contributions made in the paper and important assumptions and limitations. A No or NA answer to this question will not be perceived well by the reviewers. 
        \item The claims made should match theoretical and experimental results, and reflect how much the results can be expected to generalize to other settings. 
        \item It is fine to include aspirational goals as motivation as long as it is clear that these goals are not attained by the paper. 
    \end{itemize}

\item {\bf Limitations}
    \item[] Question: Does the paper discuss the limitations of the work performed by the authors?
    \item[] Answer: \answerYes{} %
    \item[] Justification: We discuss the limitations of our work in the paper.
    \item[] Guidelines:
    \begin{itemize}
        \item The answer NA means that the paper has no limitation while the answer No means that the paper has limitations, but those are not discussed in the paper. 
        \item The authors are encouraged to create a separate "Limitations" section in their paper.
        \item The paper should point out any strong assumptions and how robust the results are to violations of these assumptions (e.g., independence assumptions, noiseless settings, model well-specification, asymptotic approximations only holding locally). The authors should reflect on how these assumptions might be violated in practice and what the implications would be.
        \item The authors should reflect on the scope of the claims made, e.g., if the approach was only tested on a few datasets or with a few runs. In general, empirical results often depend on implicit assumptions, which should be articulated.
        \item The authors should reflect on the factors that influence the performance of the approach. For example, a facial recognition algorithm may perform poorly when image resolution is low or images are taken in low lighting. Or a speech-to-text system might not be used reliably to provide closed captions for online lectures because it fails to handle technical jargon.
        \item The authors should discuss the computational efficiency of the proposed algorithms and how they scale with dataset size.
        \item If applicable, the authors should discuss possible limitations of their approach to address problems of privacy and fairness.
        \item While the authors might fear that complete honesty about limitations might be used by reviewers as grounds for rejection, a worse outcome might be that reviewers discover limitations that aren't acknowledged in the paper. The authors should use their best judgment and recognize that individual actions in favor of transparency play an important role in developing norms that preserve the integrity of the community. Reviewers will be specifically instructed to not penalize honesty concerning limitations.
    \end{itemize}

\item {\bf Theory Assumptions and Proofs}
    \item[] Question: For each theoretical result, does the paper provide the full set of assumptions and a complete (and correct) proof?
    \item[] Answer: \answerNA{} %
    \item[] Justification: All results reported in the paper are obtained through experiments.
    \item[] Guidelines:
    \begin{itemize}
        \item The answer NA means that the paper does not include theoretical results. 
        \item All the theorems, formulas, and proofs in the paper should be numbered and cross-referenced.
        \item All assumptions should be clearly stated or referenced in the statement of any theorems.
        \item The proofs can either appear in the main paper or the supplemental material, but if they appear in the supplemental material, the authors are encouraged to provide a short proof sketch to provide intuition. 
        \item Inversely, any informal proof provided in the core of the paper should be complemented by formal proofs provided in appendix or supplemental material.
        \item Theorems and Lemmas that the proof relies upon should be properly referenced. 
    \end{itemize}

    \item {\bf Experimental Result Reproducibility}
    \item[] Question: Does the paper fully disclose all the information needed to reproduce the main experimental results of the paper to the extent that it affects the main claims and/or conclusions of the paper (regardless of whether the code and data are provided or not)?
    \item[] Answer: \answerYes{} %
    \item[] Justification: We provide detailed descriptions for the implementation of our framework and the data collection procedure.
    \item[] Guidelines:
    \begin{itemize}
        \item The answer NA means that the paper does not include experiments.
        \item If the paper includes experiments, a No answer to this question will not be perceived well by the reviewers: Making the paper reproducible is important, regardless of whether the code and data are provided or not.
        \item If the contribution is a dataset and/or model, the authors should describe the steps taken to make their results reproducible or verifiable. 
        \item Depending on the contribution, reproducibility can be accomplished in various ways. For example, if the contribution is a novel architecture, describing the architecture fully might suffice, or if the contribution is a specific model and empirical evaluation, it may be necessary to either make it possible for others to replicate the model with the same dataset, or provide access to the model. In general. releasing code and data is often one good way to accomplish this, but reproducibility can also be provided via detailed instructions for how to replicate the results, access to a hosted model (e.g., in the case of a large language model), releasing of a model checkpoint, or other means that are appropriate to the research performed.
        \item While NeurIPS does not require releasing code, the conference does require all submissions to provide some reasonable avenue for reproducibility, which may depend on the nature of the contribution. For example
        \begin{enumerate}
            \item If the contribution is primarily a new algorithm, the paper should make it clear how to reproduce that algorithm.
            \item If the contribution is primarily a new model architecture, the paper should describe the architecture clearly and fully.
            \item If the contribution is a new model (e.g., a large language model), then there should either be a way to access this model for reproducing the results or a way to reproduce the model (e.g., with an open-source dataset or instructions for how to construct the dataset).
            \item We recognize that reproducibility may be tricky in some cases, in which case authors are welcome to describe the particular way they provide for reproducibility. In the case of closed-source models, it may be that access to the model is limited in some way (e.g., to registered users), but it should be possible for other researchers to have some path to reproducing or verifying the results.
        \end{enumerate}
    \end{itemize}

\item {\bf Open access to data and code}
    \item[] Question: Does the paper provide open access to the data and code, with sufficient instructions to faithfully reproduce the main experimental results, as described in supplemental material?
    \item[] Answer: \answerNo{} %
    \item[] Justification: We will publicly release the code upon acceptance.
    \item[] Guidelines:
    \begin{itemize}
        \item The answer NA means that paper does not include experiments requiring code.
        \item Please see the NeurIPS code and data submission guidelines (\url{https://nips.cc/public/guides/CodeSubmissionPolicy}) for more details.
        \item While we encourage the release of code and data, we understand that this might not be possible, so “No” is an acceptable answer. Papers cannot be rejected simply for not including code, unless this is central to the contribution (e.g., for a new open-source benchmark).
        \item The instructions should contain the exact command and environment needed to run to reproduce the results. See the NeurIPS code and data submission guidelines (\url{https://nips.cc/public/guides/CodeSubmissionPolicy}) for more details.
        \item The authors should provide instructions on data access and preparation, including how to access the raw data, preprocessed data, intermediate data, and generated data, etc.
        \item The authors should provide scripts to reproduce all experimental results for the new proposed method and baselines. If only a subset of experiments are reproducible, they should state which ones are omitted from the script and why.
        \item At submission time, to preserve anonymity, the authors should release anonymized versions (if applicable).
        \item Providing as much information as possible in supplemental material (appended to the paper) is recommended, but including URLs to data and code is permitted.
    \end{itemize}

\item {\bf Experimental Setting/Details}
    \item[] Question: Does the paper specify all the training and test details (e.g., data splits, hyperparameters, how they were chosen, type of optimizer, etc.) necessary to understand the results?
    \item[] Answer: \answerYes{} %
    \item[] Justification: Yes, we provide detailed descriptions of the data collection and split procedure. Our texts also include important hyperparameters, such as the learning rates used for model training.
    \item[] Guidelines:
    \begin{itemize}
        \item The answer NA means that the paper does not include experiments.
        \item The experimental setting should be presented in the core of the paper to a level of detail that is necessary to appreciate the results and make sense of them.
        \item The full details can be provided either with the code, in appendix, or as supplemental material.
    \end{itemize}

\item {\bf Experiment Statistical Significance}
    \item[] Question: Does the paper report error bars suitably and correctly defined or other appropriate information about the statistical significance of the experiments?
    \item[] Answer: \answerNo{} %
    \item[] Justification: Due to our limited computational resources, we were unable to report error bars.
    \item[] Guidelines:
    \begin{itemize}
        \item The answer NA means that the paper does not include experiments.
        \item The authors should answer "Yes" if the results are accompanied by error bars, confidence intervals, or statistical significance tests, at least for the experiments that support the main claims of the paper.
        \item The factors of variability that the error bars are capturing should be clearly stated (for example, train/test split, initialization, random drawing of some parameter, or overall run with given experimental conditions).
        \item The method for calculating the error bars should be explained (closed form formula, call to a library function, bootstrap, etc.)
        \item The assumptions made should be given (e.g., Normally distributed errors).
        \item It should be clear whether the error bar is the standard deviation or the standard error of the mean.
        \item It is OK to report 1-sigma error bars, but one should state it. The authors should preferably report a 2-sigma error bar than state that they have a 96\% CI, if the hypothesis of Normality of errors is not verified.
        \item For asymmetric distributions, the authors should be careful not to show in tables or figures symmetric error bars that would yield results that are out of range (e.g. negative error rates).
        \item If error bars are reported in tables or plots, The authors should explain in the text how they were calculated and reference the corresponding figures or tables in the text.
    \end{itemize}

\item {\bf Experiments Compute Resources}
    \item[] Question: For each experiment, does the paper provide sufficient information on the computer resources (type of compute workers, memory, time of execution) needed to reproduce the experiments?
    \item[] Answer: \answerYes{} %
    \item[] Justification: We indicates the GPU resources used for our experiments in our paper.
    \item[] Guidelines:
    \begin{itemize}
        \item The answer NA means that the paper does not include experiments.
        \item The paper should indicate the type of compute workers CPU or GPU, internal cluster, or cloud provider, including relevant memory and storage.
        \item The paper should provide the amount of compute required for each of the individual experimental runs as well as estimate the total compute. 
        \item The paper should disclose whether the full research project required more compute than the experiments reported in the paper (e.g., preliminary or failed experiments that didn't make it into the paper). 
    \end{itemize}
    
\item {\bf Code Of Ethics}
    \item[] Question: Does the research conducted in the paper conform, in every respect, with the NeurIPS Code of Ethics \url{https://neurips.cc/public/EthicsGuidelines}?
    \item[] Answer: \answerYes{} %
    \item[] Justification: This work conforms with the NeurIPS Code of Ethics.
    \item[] Guidelines:
    \begin{itemize}
        \item The answer NA means that the authors have not reviewed the NeurIPS Code of Ethics.
        \item If the authors answer No, they should explain the special circumstances that require a deviation from the Code of Ethics.
        \item The authors should make sure to preserve anonymity (e.g., if there is a special consideration due to laws or regulations in their jurisdiction).
    \end{itemize}

\item {\bf Broader Impacts}
    \item[] Question: Does the paper discuss both potential positive societal impacts and negative societal impacts of the work performed?
    \item[] Answer: \answerYes{} %
    \item[] Justification: Yes, we are aware of potential impact on our work and discussed it in the paper.
    \item[] Guidelines:
    \begin{itemize}
        \item The answer NA means that there is no societal impact of the work performed.
        \item If the authors answer NA or No, they should explain why their work has no societal impact or why the paper does not address societal impact.
        \item Examples of negative societal impacts include potential malicious or unintended uses (e.g., disinformation, generating fake profiles, surveillance), fairness considerations (e.g., deployment of technologies that could make decisions that unfairly impact specific groups), privacy considerations, and security considerations.
        \item The conference expects that many papers will be foundational research and not tied to particular applications, let alone deployments. However, if there is a direct path to any negative applications, the authors should point it out. For example, it is legitimate to point out that an improvement in the quality of generative models could be used to generate deepfakes for disinformation. On the other hand, it is not needed to point out that a generic algorithm for optimizing neural networks could enable people to train models that generate Deepfakes faster.
        \item The authors should consider possible harms that could arise when the technology is being used as intended and functioning correctly, harms that could arise when the technology is being used as intended but gives incorrect results, and harms following from (intentional or unintentional) misuse of the technology.
        \item If there are negative societal impacts, the authors could also discuss possible mitigation strategies (e.g., gated release of models, providing defenses in addition to attacks, mechanisms for monitoring misuse, mechanisms to monitor how a system learns from feedback over time, improving the efficiency and accessibility of ML).
    \end{itemize}
    
\item {\bf Safeguards}
    \item[] Question: Does the paper describe safeguards that have been put in place for responsible release of data or models that have a high risk for misuse (e.g., pretrained language models, image generators, or scraped datasets)?
    \item[] Answer: \answerNo{} %
    \item[] Justification: We do not describe safeguards in our main paper.
    \item[] Guidelines:
    \begin{itemize}
        \item The answer NA means that the paper poses no such risks.
        \item Released models that have a high risk for misuse or dual-use should be released with necessary safeguards to allow for controlled use of the model, for example by requiring that users adhere to usage guidelines or restrictions to access the model or implementing safety filters. 
        \item Datasets that have been scraped from the Internet could pose safety risks. The authors should describe how they avoided releasing unsafe images.
        \item We recognize that providing effective safeguards is challenging, and many papers do not require this, but we encourage authors to take this into account and make a best faith effort.
    \end{itemize}

\item {\bf Licenses for existing assets}
    \item[] Question: Are the creators or original owners of assets (e.g., code, data, models), used in the paper, properly credited and are the license and terms of use explicitly mentioned and properly respected?
    \item[] Answer: \answerYes{} %
    \item[] Justification: We properly cite all resources.
    \item[] Guidelines:
    \begin{itemize}
        \item The answer NA means that the paper does not use existing assets.
        \item The authors should cite the original paper that produced the code package or dataset.
        \item The authors should state which version of the asset is used and, if possible, include a URL.
        \item The name of the license (e.g., CC-BY 4.0) should be included for each asset.
        \item For scraped data from a particular source (e.g., website), the copyright and terms of service of that source should be provided.
        \item If assets are released, the license, copyright information, and terms of use in the package should be provided. For popular datasets, \url{paperswithcode.com/datasets} has curated licenses for some datasets. Their licensing guide can help determine the license of a dataset.
        \item For existing datasets that are re-packaged, both the original license and the license of the derived asset (if it has changed) should be provided.
        \item If this information is not available online, the authors are encouraged to reach out to the asset's creators.
    \end{itemize}

\item {\bf New Assets}
    \item[] Question: Are new assets introduced in the paper well documented and is the documentation provided alongside the assets?
    \item[] Answer: \answerNA{} %
    \item[] Justification: This paper does not present any new assets.
    \item[] Guidelines:
    \begin{itemize}
        \item The answer NA means that the paper does not release new assets.
        \item Researchers should communicate the details of the dataset/code/model as part of their submissions via structured templates. This includes details about training, license, limitations, etc. 
        \item The paper should discuss whether and how consent was obtained from people whose asset is used.
        \item At submission time, remember to anonymize your assets (if applicable). You can either create an anonymized URL or include an anonymized zip file.
    \end{itemize}

\item {\bf Crowdsourcing and Research with Human Subjects}
    \item[] Question: For crowdsourcing experiments and research with human subjects, does the paper include the full text of instructions given to participants and screenshots, if applicable, as well as details about compensation (if any)? 
    \item[] Answer: \answerNA{} %
    \item[] Justification: We did not conduct any form of crowdsourcing or research with human subjects.
    \item[] Guidelines:
    \begin{itemize}
        \item The answer NA means that the paper does not involve crowdsourcing nor research with human subjects.
        \item Including this information in the supplemental material is fine, but if the main contribution of the paper involves human subjects, then as much detail as possible should be included in the main paper. 
        \item According to the NeurIPS Code of Ethics, workers involved in data collection, curation, or other labor should be paid at least the minimum wage in the country of the data collector. 
    \end{itemize}

\item {\bf Institutional Review Board (IRB) Approvals or Equivalent for Research with Human Subjects}
    \item[] Question: Does the paper describe potential risks incurred by study participants, whether such risks were disclosed to the subjects, and whether Institutional Review Board (IRB) approvals (or an equivalent approval/review based on the requirements of your country or institution) were obtained?
    \item[] Answer: \answerNA{} %
    \item[] Justification: Our research does not include experiments that require IRB approvals and involving human subjects.
    \item[] Guidelines:
    \begin{itemize}
        \item The answer NA means that the paper does not involve crowdsourcing nor research with human subjects.
        \item Depending on the country in which research is conducted, IRB approval (or equivalent) may be required for any human subjects research. If you obtained IRB approval, you should clearly state this in the paper. 
        \item We recognize that the procedures for this may vary significantly between institutions and locations, and we expect authors to adhere to the NeurIPS Code of Ethics and the guidelines for their institution. 
        \item For initial submissions, do not include any information that would break anonymity (if applicable), such as the institution conducting the review.
    \end{itemize}

\end{enumerate}

\end{document}